\documentclass[10pt]{article} 

\usepackage[accepted]{rlj} 

\usepackage{amssymb}            
\usepackage{mathtools}          
\usepackage{mathrsfs}           
\usepackage{graphicx}           
\usepackage{subcaption}         
\usepackage[space]{grffile}     
\usepackage{url}                
\usepackage{lipsum}             

\usepackage[linesnumbered,ruled,vlined]{algorithm2e}

\usepackage{booktabs}

\newenvironment{widealgorithm}{
\begin{minipage}{\linewidth}
\begin{algorithm}[H]}
{\end{algorithm}
\end{minipage}}

\DeclareMathOperator*{\argmax}{arg\,max}

\usepackage{natbib}

\usepackage{amsthm}
\newtheorem{lemma}{Lemma}

\usepackage{bbm}

\title{Group Fairness in Multi-Task \\ Reinforcement Learning}

\setrunningtitle{Group Fairness in Multi-Task Reinforcement Learning}

\author{Kefan Song\textsuperscript{1}, Runnan Jiang\textsuperscript{2}, Rohan Chandra\textsuperscript{1}, Shangtong Zhang\textsuperscript{1}}

\emails{ks8vf@virginia.edu, rj2666@columbia.edu,\\ \ \{rohanchandra,shangtong\}@virginia.edu}

\affiliations{
$^{1}$\textbf{University of Virginia}\\
$^{2}$\textbf{Columbia University}\\
}

\contribution{
    This paper introduces a new multi-task group fairness reinforcement learning framework that enforces fairness constraints across multiple tasks simultaneously.
    }
    {
    Prior work only addresses group fairness in single-task settings \citep{satija2023group}.
    }
\contribution{
    We prove that our algorithm achieves zero fairness violations with high probability and sublinear regret in the finite-horizon setting.
    }
    {
    None
    }
\contribution{
    We extend our approach to the infinite-horizon setting and empirically demonstrate its effectiveness in MuJoCo environments.
    }
    {
    None
    }

\keywords{Multi-Task Reinforcement Learning, Group Fairness, Constrained Markov Decision Process, Algorithmic Fairness, Demographic Parity.} 

\summary{This paper addresses a critical societal consideration in the application of Reinforcement Learning (RL): ensuring equitable outcomes across different demographic groups in multi-task settings. While previous work has explored fairness in single-task RL, many real-world applications are multi-task in nature and require policies to maintain fairness across all tasks. We introduce a novel formulation of multi-task group fairness in RL and propose a constrained optimization algorithm that explicitly enforces fairness constraints across multiple tasks simultaneously. We have shown that our proposed algorithm does not violate fairness constraints with high probability and with sublinear regret in the finite-horizon episodic setting. Through experiments in RiverSwim and MuJoCo environments, we demonstrate that our approach better ensures group fairness across multiple tasks compared to previous methods that lack explicit multi-task fairness constraints in both the finite-horizon setting and the infinite-horizon setting. Our results show that the proposed algorithm achieves smaller fairness gaps while maintaining comparable returns across different demographic groups and tasks, suggesting its potential for addressing fairness concerns in real-world multi-task RL applications.
}

\begin{document}

\maketitle  

\begin{abstract}
This paper addresses a critical societal consideration in the application of Reinforcement Learning (RL): ensuring equitable outcomes across different demographic groups in multi-task settings. While previous work has explored fairness in single-task RL, many real-world applications are multi-task in nature and require policies to maintain fairness across all tasks. We introduce a novel formulation of multi-task group fairness in RL and propose a constrained optimization algorithm that explicitly enforces fairness constraints across multiple tasks simultaneously. We have shown that our proposed algorithm does not violate fairness constraints with high probability and with sublinear regret in the finite-horizon episodic setting. Through experiments in RiverSwim and MuJoCo environments, we demonstrate that our approach better ensures group fairness across multiple tasks compared to previous methods that lack explicit multi-task fairness constraints in both the finite-horizon setting and the infinite-horizon setting. Our results show that the proposed algorithm achieves smaller fairness gaps while maintaining comparable returns across different demographic groups and tasks, suggesting its potential for addressing fairness concerns in real-world multi-task RL applications.
\end{abstract}

\section{Introduction}

Learning-based algorithms have been applied more to real-world high-stakes social problems, such as bank loans, medical interventions, and school admissions \citep{barocas-hardt-narayanan, fairness_survey, feng2024fairmachinelearninghealthcare}. They are also applied in less high-stakes scenarios, by making video recommendations, suggesting products to buy, and question answering \citep{ youtube_recommendations, product_recommender_sys, devlin-etal-2019-bert}. In both cases, the deployment of learning-based algorithms will make automated decisions that have a direct impact on our society. Therefore, one critical issue is to ensure the algorithm has low social biases and delivers fair outcomes for people from all demographic groups \citep{dwork2011fairnessawareness}. However, since these social problems are long-term in nature, an unconstrained algorithm may create a feedback loop over time and enlarge the discrepancy between people from different social groups \citep{yin2023longterm}.

Reinforcement Learning has demonstrated a superior performance in many of these tasks \citep{rl_recommender, reinforce_recommender}, which are sequential-decision making problems in nature. When the fairness requirement is accounted for in the RL algorithm, it has the promise of addressing the long-term fairness issue, thus has an advantage over fair machine learning algorithms \citep{gajane2022surveyfairreinforcementlearning}.

In this paper, we study the problem of group fairness reinforcement learning in the multi-task setting. We focus on demographic parity, a particular definition of group fairness. It requires the algorithm to deliver similar outcome for people from different social groups, categorized by their sensitive information such as gender, education or social-economic status. Many real world applications are multi-task in nature \citep{zhang2021surveymultitasklearning} and it is critical to ensure group fairness is achieved for all tasks. To our best knowledge, this is the first paper that account for the algorithm fairness problem in multi-task reinforcement learning problem. To further motivate our problem, we discuss two application scenarios in the following. 

\textbf{Scenario 1: RL-based Recommender Systems.}
Consider an example of multi-task recommender systems, where an RL policy recommends contents catered to the preference of the users and aims to achieve multiple tasks such as a high click-through rate and a high long-term user engagement. The users’ sensitive information, such as age, social economic status and gender, is taken by the policy as input features to make recommendations.  

When there is no group fairness consideration during the algorithm development, it is more likely for the algorithm to maximize the user engagement of the majority social groups, and create a feedback loop by increasing the size of the majority social groups. Algorithms that achieve fairness on a single task such as click-through rate may not ensure fairness on other tasks such as long-term engagement, potentially driving minority groups to leave the platform. 

With a more balanced social group ratio in the system’s user pool, more content creators need to include the minority group as their targeted audience and thus create content that is more inclusive and engaging for the minority group, potentially breaking the feedback loop and fostering a healthy, sustainably growing community.


\textbf{Scenario 2: Fine-tuning LLMs with RL.}
When fine-tuning Large Language Models with Reinforcement Learning from Human Feedback (RLHF) over multiple tasks such as common sense reasoning, question answering, and explanation generation, the training data is a collection of human prompts as inputs for the LLM, which can also be regarded as the states for the RL policy. Since the prompts are collected without considering people's diverse social groups, an inherent imbalance exists with respect to specific demographics in the dataset. Consequently, the LLM fine-tuned with RL may disproportionally improve the quality of responses to prompts from majority groups on tasks without fairness guarantees. A feedback loop exists if the deployed updated LLM results in more active users from the majority groups, who may generate new data for further LLM fine-tuning.

\section{Related Works}

\textbf{Algorithm Fairness in Multi-Task Learning.}
Recent work has explored various approaches to ensure fairness in multi-task learning settings. \citet{fair_mtl} incorporates multi-marginal Wasserstein barycenters to achieve demographic parity in multi-task regression and classification problems. \citet{adaptive_fair_mtl} developed a more flexible approach using teacher-student networks to dynamically balance fairness and accuracy objectives. \citet{Wang_2021} have characterized the fundamental trade-off between fairness and accuracy using Pareto-front analysis, and proposed novel architectures where task-specific fairness losses are backpropagated to head layers while overall fairness objectives influence shared layers.

\textbf{Algorithm Fairness in Single-Task Reinforcement Learning.}
In the single-task reinforcement learning domain, several approaches have emerged to address fairness concerns. \citet{yu2022} introduced advantage regularization techniques for fair credit lending across demographic groups. \citet{chi2022returnparity} contributed by defining return parity as a fairness metric in Markov Decision Processes (MDPs) and developing algorithms to reduce long-term reward disparities through state visitation distribution alignment. Recent work by \citet{yin2023longterm} and \citet{satija2023group} has expanded our understanding of long-term fairness implications in reinforcement learning, leveraging safe RL techniques to maintain fairness constraints throughout the learning process.

\textbf{Fair Resource Allocation through Reinforcement Learning.}
While our work focuses on algorithmic fairness as defined by \citet{barocas-hardt-narayanan}, which addresses societal biases affecting different demographic groups, it's important to distinguish this from resource allocation fairness. In the resource allocation domain, \citet{yao2021reinforcedworkloaddistributionfairness} applied RL to achieve fair workload distribution in data centers through reward shaping. Similarly, \citet{mt-drl} explored multi-task RL for fair network traffic allocation. However, these approaches differ fundamentally from our work as they neither provide fairness guarantees nor account for group-specific MDP transitions, making them unsuitable for addressing algorithmic fairness challenges in reinforcement learning.

Our work is differentiated from these research directions by developing a multi-task reinforcement algorithm that ensures algorithm fairness throughout the learning process while accounting for group-specific dynamics. 

\section{Multi-Task Group Fairness in Finite-horizon MDP}
\subsection{Preliminaries}
A multi-task finite-horizon Markov Decision Process (MDP) is defined as a tuple $\mathcal{M} = (\mathcal{S}, \mathcal{A}, H, P, \{r_m\}_{m=1}^M, \mu)$, where $\mathcal{S}$ is the state space, $H$ is the number of steps in each episode, and $P_h(\cdot|s, a) \in \Delta_{\mathcal{S}}, \forall s \in \mathcal{S}, \forall a \in \mathcal{A}, \forall h \in H$, where $\Delta_{\mathcal{S}}$ is the $|\mathcal{S}|$-dimensional probability simplex. The tasks within MDP are characterized by distinct reward functions $\{r_m\}_{m=1}^M$, where $r_m : \mathcal{S} \times \mathcal{A} \leftarrow [0, 1]$ specifies the reward function for each task $m$, and $M$ is the total number of tasks. The algorithm samples a total of $K$ episodes from the environment. We assume the initial state distribution $\mu$ is known to the agent and reward functions are deterministic. 

\subsection{Group Fairness}
For our definition of fairness, we adopt the demographic parity notion, also commonly known as group fairness. It requires the outcomes experienced by individuals to be independent of their particular social group membership, where each social group is denoted as $z \in \mathcal{Z}$.  

In the long-term group fairness problem, we ensure that the expected return is equal across all groups. We assume all groups share the same state and action spaces, discount factor, and reward functions, but each group has a different initial state distribution $\mu_z$ and a different transition function $P_z$. The return of policy $\pi$ under transition $P_z$ and initial state distribution $\mu_z$ is denoted as $J(\pi_z; \mu_z, P_z, r)$, and the long-term group fairness for a single task $r$ is defined as:  
\begin{equation}
    J(\pi_i; \mu_i, P_i, r) = J(\pi_j; \mu_j, P_j, r), \quad \quad \quad \quad \forall i \leq j; (i, j) \in \mathcal{Z}^2 
\end{equation}

In practice, we relax this constraint by introducing a positive slack variable $\epsilon > 0$ and ensure the difference in return is within this tolerance:
\begin{equation}
    |J(\pi_i; \mu_i, P_i, r) - J(\pi_j; \mu_j, P_j, r)| < \epsilon,\quad \quad \forall i \leq j; (i, j) \in \mathcal{Z}^2.
\end{equation}

The fairness threshold for the acceptable performance difference between any two groups is denoted as $\epsilon: \epsilon \in (0, H]$.

\subsection{Algorithm for Zero-Constraint Violation for Multi-task Setting}


The multi-task group fairness RL problem is formulated as finding a list of optimal policies $\pi^*$ that obey the group fairness constraint across all tasks $m \in [M]$
\begin{equation}
\begin{aligned}
    \pi^* &= \underset{\pi}{\arg\max} \, \sum_{m=1}^M \sum_{z \in \mathcal{Z}} J(\pi_z; \mu_z, P_z, r_m), \\
    \text{s.t.} \quad \max_m(|J(\pi_i; \mu_i, P_i, r_m)& - J(\pi_j; \mu_j, P_j, r_m)|) \leq \epsilon, \quad \forall i \geq j; \, (i, j) \in \mathbb{Z}^2, \forall m \in [M]. 
\end{aligned}
\label{eq:finite-horizon problem formulation}
\end{equation}

In practice, our algorithm iteratively solves the following optimization problem at each episode k
\begin{equation}
\begin{aligned}
\pi^k \in \argmax_{\pi \in \mathbf{\Pi}^k} \sum_{m=1}^M \sum_{z \in \mathcal{Z}} J(\pi_z; \mu_z, \hat{P}_z, {r^{\mathrm{opt}}}_m), 
\end{aligned}
\label{eq:finite-horizon practical problem}
\end{equation}
where $\mathbf{\Pi}^k$ is a conservatively estimated set of policies ensuring fairness, $\hat{P}_z$ is the estimated transition for group $z$, and $r^{\mathrm{opt}}$ is an exploration-augmented reward. We will detail these components below. The full Algorithm can be found in Appendix \ref{sec:appendix_tabular}.

%
\textbf{Conservative Policy Set Construction.}
One key objective of our work is to ensure that our algorithm does not violate the group fairness constraint in \eqref{eq:finite-horizon problem formulation} during training, where the fairness gap is calculated by the absolute difference between the returns of two groups. We seek to construct a set of policies that obey the group fairness constraint, and then find the policy with maximum return within the set. However, the true transition $P_z$ is unknown to our algorithm and we can only estimate the fairness gap by sampling from the true environment to evaluate the returns of different groups. A poor estimation of the fairness gap may result in selecting a policy whose true fairness gap violates the fairness constraint by a large margin. 

To address this issue, we aim to construct a conservative set of policies that will achieve zero-fairness-constraint violation with high probability. Following the techniques from \cite{satija2023group}, we design an optimistic estimation of the fairness gap and then select policies whose optimistic fairness gap is less than or equal to the fairness threshold $\epsilon$ to construct the conservative set of policies. 

Designing the optimistic fairness gap requires an optimistic reward $\bar{r}^k_{m,h}$ and a pessimistic reward $\underbar{r}^k_{m,h}$ defined as:
\begin{align}
    \bar{r}^k_{m, h}(s, a) \doteq r^k_{m,h}(s, a) + |\mathcal{S}|H \beta_{m,h}^k(s, a), \label{eq:optimistic reward} \\
    \underbar{r}^k_{m, h}(s, a) \doteq r^k_{m,h}(s, a) - |\mathcal{S}|H \beta_{m,h}^k(s, a), \label{eq:pessimistic reward}
\end{align}
where $\beta_{m,h}^k(s, a)$ is the confidence radius to account for the uncertainties from the transition probabilities.

Taking a model-based policy evaluation approach, the return of the policy is evaluated using an estimated transition $\hat{P}^k_z$. The optimistic and pessimistic reward estimates then allow us to calculate the difference between an optimistic return from one group and a pessimistic return from the other group, which gives us the optimistic fairness gap. When selecting policies that obey the fairness threshold for every task m, a set of safe policies can be constructed as the following:
\begin{align}
    \label{eq:selecting_fair_policies}
    \mathbf{\Pi}_F^k \coloneqq \left\{ \pi : 
    \begin{array}{l}
        J\big(\pi_i; \mu_i, \hat{P}_i^k, \bar{r}_m^{k}\big) - J\big(\pi_j; \mu_j, \hat{P}_j^k, \underbar{r}_m^k\big) \leq \epsilon, \quad \forall i \geq j; \, (i, j) \in \mathbb{Z}^2, \, \forall m \in [M]. \\[1ex]
        J\big(\pi_j; \mu_j, \hat{P}_j^k, \bar{r}_m^{k}\big) - J\big(\pi_i; \mu_i, \hat{P}_i^k, \underbar{r}_m^k\big) \leq \epsilon, \quad \forall i \geq j; \, (i, j) \in \mathbb{Z}^2, \, \forall m \in [M].
    \end{array}
    \right\}
\end{align}
When the transitions are poorly estimated, it is possible that no policy obeys the constraint. To ensure the problem in Equation \eqref{eq:finite-horizon practical problem} is always feasible, we assume there exists an initial strictly fair policy $\pi_0$ that our algorithm can use to safely sample data from the environment. 

\textbf{Assumption 1.1 (Initial strictly fair policy) }
\label{assumption:initial_fair_policy} The algorithm has access to a policy \( \pi \) that satisfies the fairness constraints in Equation~\eqref{eq:finite-horizon problem formulation}. We also assume $\left| J(\pi^0; \mu_i, P_i, r_m) - J(\pi^0; \mu_j, P_j, r_m) \right|$ $\leq \epsilon^0 < \epsilon, \forall (i, j) \in \mathbb{Z}^2, \forall m \in [M]$ and the value of \(\epsilon^0\) is known to the algorithm.

In case the above policy set is empty, we can simply use the strictly fair policy $\pi_0$ that will not violate the fairness constraint in the true MDP to sample more data for a better estimated transitions $\hat{P}_z$. Executing $\pi_0$ under the condition in the following is sufficient to guarantee that $\mathbf{\Pi}_F^k$ is non-empty in the otherwise condition. 

We construct the conservative set of policies $\mathbf{\Pi}^k$ as follows:
\begin{align}
    \label{eq:conservative set of policies}
    \mathbf{\Pi}^k = 
    \begin{cases} 
    \{\pi^0\}, 
    \begin{cases}
        \text{if } J\big(\pi^0_i; \mu_i, \hat{P}_i^k, \bar{r}_m^{k}\big) - J\big(\pi^0_j; \mu_j, \hat{P}_j^k, \underbar{r}_m^k\big) > \frac{\epsilon + \epsilon^0}{2}, \\[1ex]
        \text{or } J\big(\pi^0_j; \mu_j, \hat{P}_j^k, \bar{r}_m^{k}\big) - J\big(\pi^0_i; \mu_i, \hat{P}_i^k, \underbar{r}_m^k\big) > \frac{\epsilon + \epsilon^0}{2}, \\[1ex]
        \quad\quad\quad \forall i \geq j,\; (i, j) \in \mathbb{Z}^2,\; \exists m \in [M].
    \end{cases} \\[6ex]
    \mathbf{\Pi}^k_F, 
    \enspace \text{otherwise}.
    \end{cases}
\end{align}

\textbf{Exploration Bonus Design.} Besides zero fairness violation, we also care about achieving sub-linear regret. Under the principle of optimism under the face of uncertainty, we set another exploration bonus for the estimated reward function $\hat{r}_{m, h}^k(s, a)$ of each task $m$ and timestep $h$ to achieve efficient exploration
\begin{equation}
\begin{aligned}
    \textstyle {r^{\mathrm{opt}}}_{m, h}^k(s, a) = \hat{r}_{m, h}^k(s, a) + \alpha \beta_{m,h}^k(s, a),
\end{aligned}
\end{equation}
where $\alpha=|\mathcal{S}|H + \frac{4|\mathcal{S}|H}{\epsilon - \epsilon^0}2H$. 



\subsection{Theoretical Guarantees} We now present a result stating that policies chosen from $\mathbf{\Pi}^k$ do not violate the fairness guarantees for any of the subgroups throughout the learning duration with high probability.

\textbf{Theorem 1.1 (Fairness violation) }
\label{thm:fairness_violation}
Given an input confidence parameter $\delta \in (0, 1)$ and an initial fair policy $\pi^0$, the construction of $\mathbf{\Pi}^k$ ensures that there are no fairness violations at any episode in the learning procedure in the true environment with high probability $(1 - \delta)$, i.e., for any $\pi \in \mathbf{\Pi}^k$,
\[
\begin{gathered}
\Pr\Big( \Big| J(\pi_i; \mu_i, P_i, r_m) - J(\pi_j; \mu_j, P_j, r_m) \Big| \leq \epsilon \Big) \geq 1 - \delta, \\[1ex]
\forall m \in [M], \forall k \in [K], \forall i \geq j; \, (i, j) \in \mathcal{Z}^2.
\end{gathered}
\]







With the fairness guarantee established by Theorem \hyperref[thm:fairness_violation]{1.1}, it is equally critical to ensure that the learning algorithm is efficient with respect to exploration. In particular, we need to bound the cumulative regret to demonstrate that the process is not only fair but also effective over time. To address this, we now present a result that provides a sublinear regret guarantee.

\textbf{Theorem 1.2 (Regret Bound) }
\label{thm:regret_bound} For any $\delta \in (0, 1)$, with probability $1 - \delta$, for any task $m$, executing $\pi^k$ from Equation (\ref{eq:finite-horizon practical problem}) at every episode $k \in [K]$ incurs in a regret of at most
\[
\text{Reg}(K; r_m) = \tilde{\mathcal{O}}\left(\frac{|\mathcal{Z}|H^3}{(\epsilon - \epsilon^0)}\sqrt{|\mathcal{S}|^3|\mathcal{A}|K} + \frac{|\mathcal{Z}|^2MH^5|\mathcal{S}|^3|\mathcal{A}|}{\min\{(\epsilon - \epsilon^0),(\epsilon - \epsilon^0)^2\}}\right),
\]
where $\tilde{O}(\cdot)$ hides polylogarithmic terms. Proofs of both theorems are provided in Appendix \ref{high_prob_good_event}.

\subsection{Experimental Results}
\textbf{Extended RiverSwim Multi-Task Environment.}
We propose a multi-task extension of the classic RiverSwim environment, modified for two social groups \citep{satija2023group}, to evaluate fairness violations across correlated tasks. The environment comprises two social groups, $z \in \mathcal{Z}$, each with distinct transition dynamics $P_z$, and a two-task setting that preserves these dynamics across tasks.

The base RiverSwim environment consists of 7 states, where the agent aims to swim from the leftmost state to the rightmost state. For Task 1, the objective is to reach the rightmost state to receive a reward of 1. For Task 2, we modify the reward structure by assigning a reward of 1 when the agent swims rightward passing state 3. Although the reward localization differs between tasks, both require the agent to start from the leftmost state and navigate rightward, reflecting a high correlation similar to real-world scenarios (e.g., recommender systems optimizing both long-term user engagement and click-through rate).

In this setting, an RL algorithm that guarantees fairness in only one task may not extend these guarantees to another, potentially leading to policies that exhibit biased behavior across tasks, and our experimental setup particularly examines this challenge.

\textbf{Baseline.}
In this experiment, we treat the group fairness reinforcement learning algorithm (GFRL) developed for tabular setting as the baseline \citep{satija2023group}. We train the algorithm on the first task and evaluate its fairness violation in the first task and the second task. 

\begin{figure}[htbp]
    \centering
    \begin{subfigure}{0.32\textwidth}
        \centering
        \includegraphics[width=\textwidth]{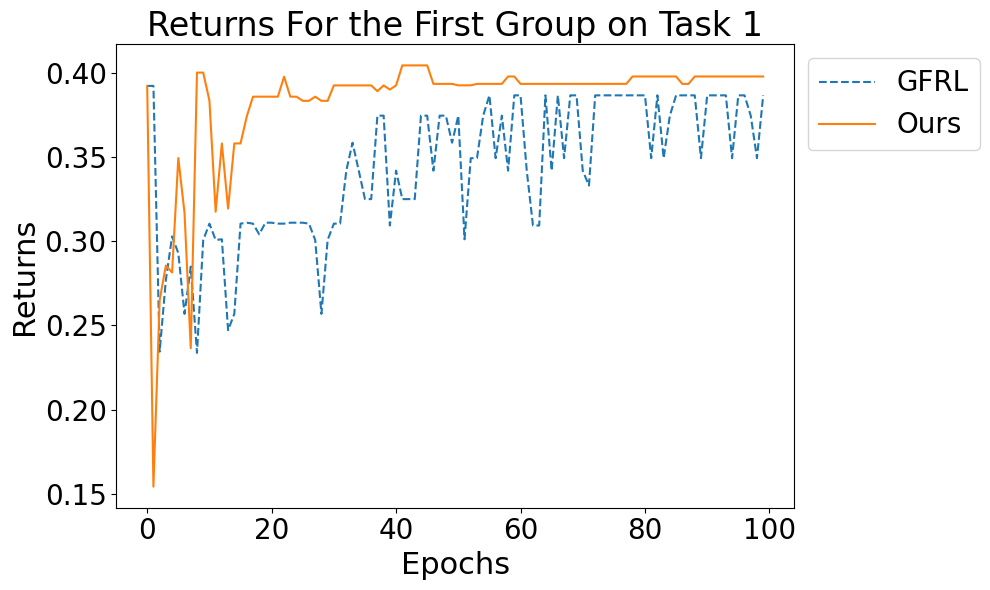}
    \end{subfigure}
    \begin{subfigure}{0.32\textwidth}
        \centering
        \includegraphics[width=\textwidth]{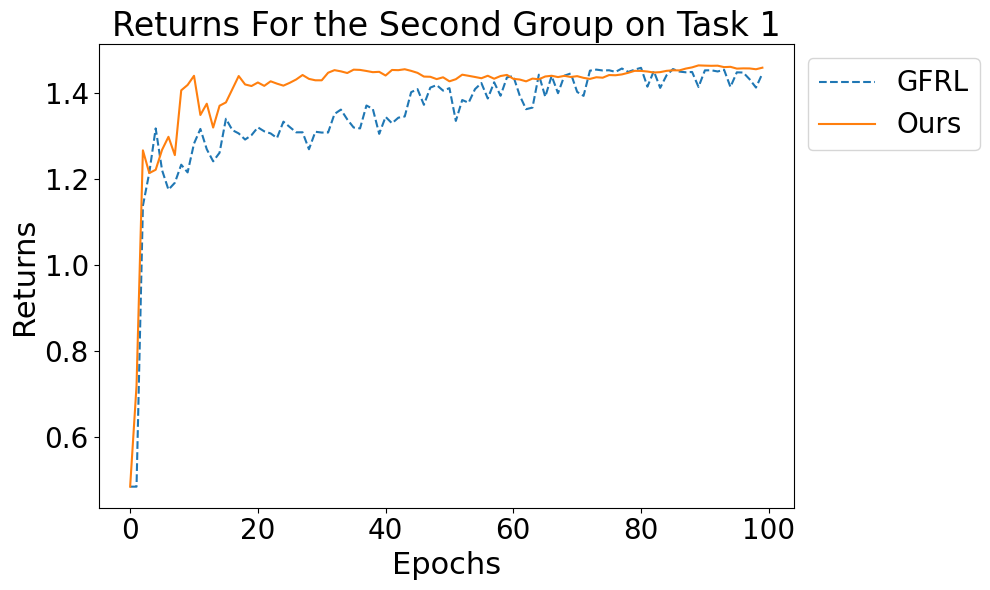}
    \end{subfigure}
    \begin{subfigure}{0.32\textwidth}
        \centering
        \includegraphics[width=\textwidth]{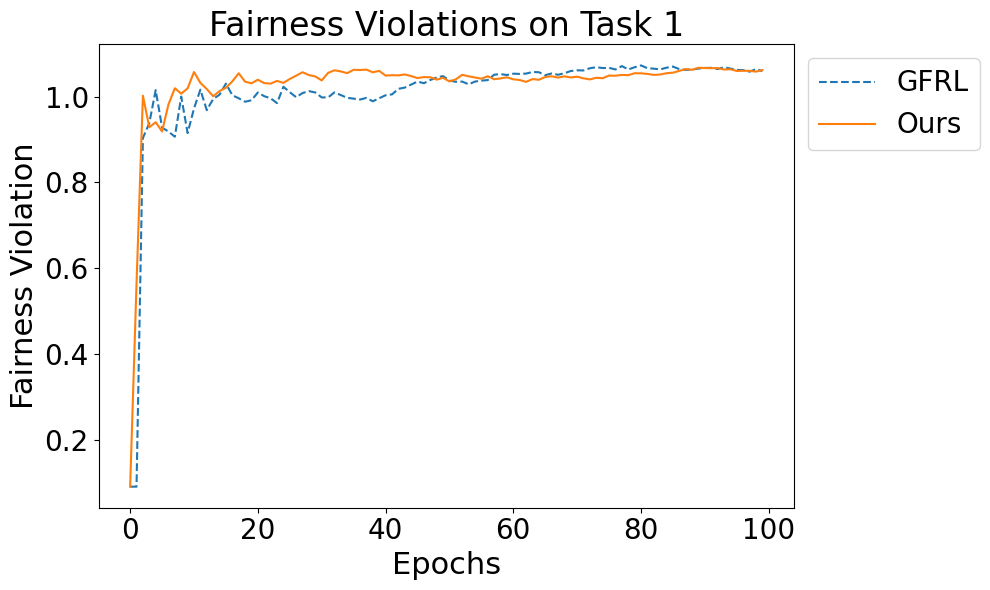}
    \end{subfigure}
    
    \begin{subfigure}{0.32\textwidth}
        \centering
        \includegraphics[width=\textwidth]{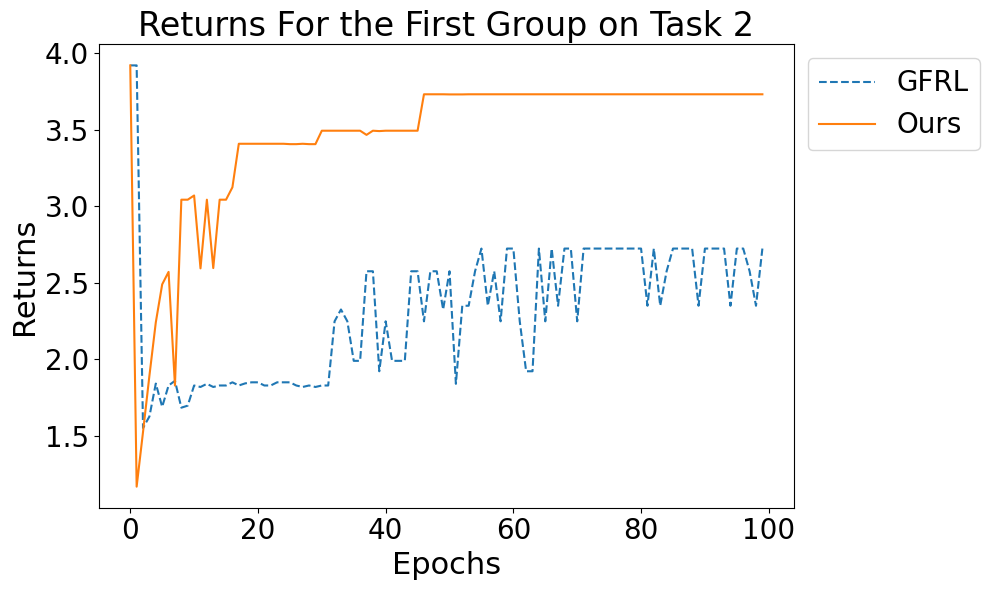}
    \end{subfigure}
    \begin{subfigure}{0.32\textwidth}
        \centering
        \includegraphics[width=\textwidth]{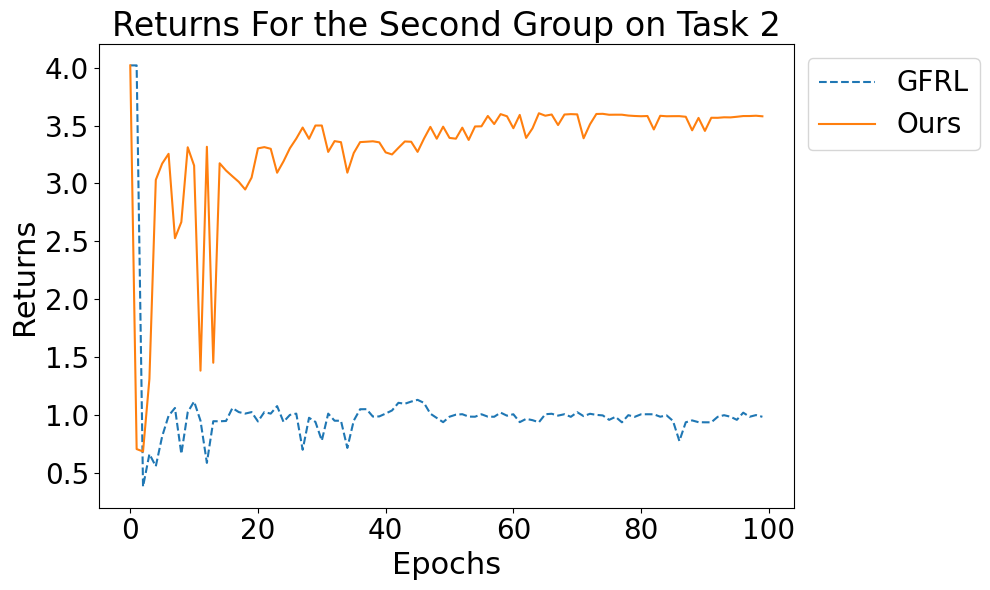}
    \end{subfigure}
    \begin{subfigure}{0.32\textwidth}
        \centering
        \includegraphics[width=\textwidth]{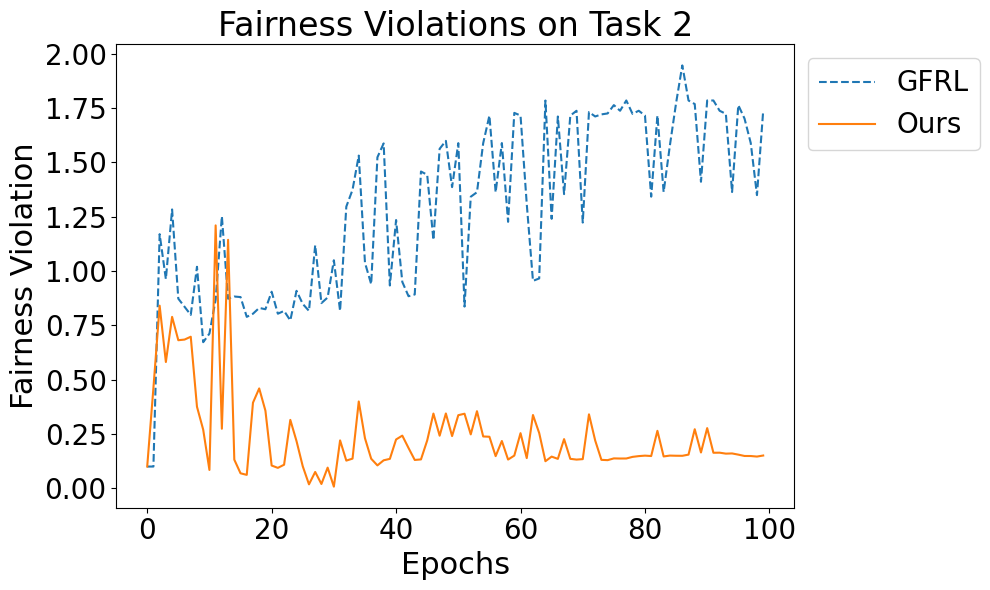}
    \end{subfigure}

    \caption{Results for both tasks: The first row shows results for Task 1, and the second row shows results for Task 2. Columns represent subgroup returns and fairness gaps.}
    \label{fig:combined}
\end{figure}

\textbf{Results.}
From Figure \ref{fig:combined}, we have shown that our algorithm achieves a high return on task 1 without significant fairness violations on both task 1 and task 2, whereas the baseline algorithm achieves similar fairness violation on the first task, but significantly violates fairness on the second task. This illustrates that even for similar tasks, an algorithm that ensures group fairness for one task does not guarantee achieving group fairness in the other task.

\section{Multi-Task Group Fairness in Infinite-Horizon MDP}
In the previous sections, we focused on a tabular, finite-horizon MDP setting for multi-task group fairness. We now extend our framework to an infinite-horizon discounted MDP, which more closely models many real-world scenarios. This setting involves continuous or high-dimensional state spaces, and requires ensuring group fairness over long-term behavior. Below, we describe the formal definition of the infinite-horizon MDP, outline the constrained problem formulation, and propose a methodology to achieve multi-task group fairness in this setting.


\subsection{Preliminaries} We formulate the long-term fairness problem as an infinite-horizon discounted Markov Decision Process (MDP), defined by the tuple $\langle \mathcal{S}, \mathcal{A}, \gamma, \mu, r, P \rangle$, where $\mathcal{S}$ is the state space,  $\mathcal{A}$ is the action state, $\mu: \mathcal{S} \rightarrow [0,1] $ is the initial state distribution, $\gamma \in [0, 1)$ is the discount factor, $r: \mathcal{S} \times \mathcal{A} \rightarrow [0, 1]$ is the reward function, and $P: \mathcal{S} \times \mathcal{A} \times \mathcal{S} \rightarrow [0,1]$ is the transition function. 
In this setting, a stationary policy $\pi$ is defined as $\pi: \mathcal{S} \times \mathcal{A} \rightarrow [0, 1]$. The infinite-horizon discounted return of policy $\pi$ and reward $r$ is defined as $J(\pi; \mu, P, r) \stackrel{\cdot}{=} \mathrm{E}_{\tau \sim p_\pi(\tau)}[\sum_{t=1}^\infty \gamma^t r(s_t, a_t)]$. The value function is defined as $v_\pi(s;\mu, P, r) = \mathrm{E}_{\tau \sim p_\pi(\tau)}[\sum_{t=1}^\infty \gamma^t r(s_t, a_t)| s_t = s]$, and the state-action value function is defined as $q_\pi(s, a; \mu, P, r) = \mathrm{E}_{\tau \sim p_\pi(\tau)}[\sum_{t=1}^\infty \gamma^t r(s_t, a_t) | s_t = s, a_t = a]$. The advantage function is then defined by $A_\pi(s, a;\mu, P, r) = q_\pi(s, a;\mu, P, r) - v_\pi(s, a;\mu, P, r)$.

In this paper, we consider the multi-task Reinforcement Learning problem, where a collection of tasks share the same state and action spaces, discount factor, and transition function, but have different reward functions $r \in \{r_m\}_{m=1}^N$.
\subsection{Constrained Markov Decision Process}
The focus of the Constrained Markov Decision Process (CMDP) is to find a policy that maximizes return, only from the set of policies that obey the constraints. The constraints in CMDP are specified by a set of constraint reward functions $\{C_n\}_{n=1}^N$, where $C_n: \mathcal{S} \times \mathcal{A} \rightarrow \mathbb{R}$, and a set of corresponding scalar constraint tolerances $\{\theta_n\}_{n=1}^N$. The set of policies that obey the constraints is denoted by:
\begin{align}
    \Pi_C \mathrel{\dot{=}} \{\pi \in \Pi : \forall n, J(\pi; \mu, P, C_n) \leq \theta_n  \},
\end{align}
and to find an optimal policy in a CMDP is to solve the following optimization problem:
\begin{align}
    \pi^* = \argmax_{\pi\in \Pi_C} J(\pi).
\end{align}

\subsection{Algorithm for Multi-Task Group Fairness in the Infinite-Horizon Setting}
We are now ready to formulate the Group Fairness in Multi-Task Reinforcement Learning problem. 
We aim to ensure the long-term outcome experienced by different social groups to be equal during the training process, so we are not restricted to using a single policy for all social groups. Let $\pi$ denotes a list of policies $\pi$. A social-group specific policy $\pi_z$ is used to solve for each social group's specific transition $P_z$, and our goal is to find a list of optimal policies $\pi^*$ that obey the relaxed group fairness constraint across all tasks $r_m \in \{r_m\}_{m=1}^M$
\begin{equation}
\begin{aligned}
    \pi^* &= \textstyle \argmax_{\pi} \; \textstyle \sum_{i}\sum_{m=1}^M J(\pi_i; \mu_i, P_i, r_m) \\
    &\text{s.t.} \enspace \max_{r_m} \; |J(\pi_i; \mu_i, P_i, r_m) - J(\pi_j; \mu_j, P_j, r_m)| \leq \epsilon, \quad \quad \forall i \leq j; \, (i,j) \in \mathcal{Z}^2, \, m \in [M].
    \label{drl_problem_formulation}
\end{aligned}
\end{equation}

To practically tackle this problem, we first frame it as a CMDP problem and then use the constrained policy optimization algorithm to solve it. 

In practice, instead of finding the list of all policies at the same time, we update each group's policy $\pi_i$ at a time in a block coordinate descent way, which may not give us the optimal solution of the original problem. Under this setting, the objective function can be simplified to only include the return of group $i$. Since the policies of other social groups are not updated, the returns of reward function $r_n$ for other social groups remain constant, denoted as $\bar{J}_j(r_n) \mathrel{\dot{=}} J(\pi_j; \mu_j, P_j, r_n)$, which can be excluded from the objective function. Note that ensuring the maximum difference in return to be less than $\epsilon$ is equivalent to ensuring all differences in return to be less than $\epsilon$, so the constraint in \eqref{drl_problem_formulation} can be written into $N$ number of inequalities. Therefore, the objective and constraints can be rewritten as the following:
\begin{align}
    \pi_i^* &= \argmax_{\pi_i} \sum_m J(\pi_i; \mu_i, P_i, r_m) \nonumber\\
    &\text{s.t.} \enspace |J(\pi_i; \mu_i, P_i, r_m) - \bar{J}_j| \leq \epsilon, \quad \quad \forall i \leq j; \, (i,j) \in \mathcal{Z}^2, \, r_m \in \{r_m\}_{m=1}^M.
    \label{original problem}
\end{align}

To formulate our problem into a CMDP problem, let the constraint reward function be $C_n(s, a) = r_m(s, a)$ for the first M inequalities where $n \in \{1, 2, ..., M\}$, and the corresponding constraint tolerance $\theta_n = \epsilon + \bar{J}_j(r_m)$. For the second M inequalities, we define the constraint reward function as $C_n(s, a) = -r_m(s, a)$ and set the constraint tolerance to $\theta_n = \epsilon - \bar{J}_j(r_m)$, where $n \in \{M+1, M+2, ..., 2M\}$.

Then, finding the optimal policy for a specific social group $i$ is to solve the following CMDP problem
\begin{equation}
    \pi_i^* = \argmax_{\pi \in \Pi_C} J(\pi; \mu_i, P_i, \sum_m r_m), \label{block descent}
\end{equation}
where 
\begin{equation}
    \Pi_C \mathrel{\dot{=}} \{\pi_i \in \Pi: \forall n, i\leq j; (i, j) \in \mathcal{Z}^2, J(\pi_i, P_j, C_n) \leq \theta_n\} .
\end{equation}

\subsection{Constrained Policy Optimization Methodology}

Constrained Policy Optimization (CPO) is one method that solves the CMDP problem. It has the advantage of maintaining constraint satisfaction throughout training, whereas other methods such as Primal-Dual Optimization \citep{PrimalDual2015} only achieve constraint satisfaction after policy converges. As one of the trust region methods, CPO aims to maximize the next updated policy's performance improvement from the old policy of the current iteration: $J(\pi^{k+1}) - J(\pi^k)$, while keeping the new policy's costs within the tolerances, $J(\pi^{k+1}; \mu, P, C_m) \leq d_m$ for all cost functions $C_m$ and all tolerances $d_m$. To avoid the problem of off-policy evaluation for $\pi^{k+1}$, in practice, only a lower bound for the performance difference and an upper bound of the cost of the new policy that is dependent on $d^\pi$ are used in the optimization. 

The proposed CPO method is as follows:
\begin{equation}
\begin{aligned}
    \pi^{k+1} &= \argmax_{\pi_\theta \in \Pi_\theta}\underset{\substack{s \sim d^{\pi_k} \\ a \sim \pi_\theta}}{\mathrm{E}}[A^{\pi^k}(s, a; \mu, P, r)] \\
    \text{ s.t. } & J(\pi^k; \mu, P, C_m) + \frac{1}{1 - \gamma}\underset{\substack{s \sim d^{\pi^k} \\ a \sim \pi_\theta}}{\mathrm{E}}[A^{\pi^k}(s, a; \mu, P, C_m)] \leq d_m \quad \forall m \\
    & \underset{\substack{s \sim \pi^k}}{\mathrm{E}}\left[ D_{KL}\left(\pi_\theta(\cdot|s) || \pi^k(\cdot |s)\right)\right] \leq \delta. \label{eq: CPO math problem}
\end{aligned}
\end{equation}

The original CPO algorithm relies on the second-order Taylor approximation and inverting a high-dimensional Fisher information matrix. A first-order method, FOCOPS, is proposed by \cite{zhang2020focops} for the CPO problem. To solve the group fairness problem, FOCOPS is required to handle more than one constraint. We extended the FOCOPS algorithm for multiple constraints in Algorithm \ref{algo:focops}, and in Algorithm \ref{algo:drl_mtgf}, we propose a multi-objective group fairness reinforcement learning algorithm. Both algorithms are included in Appendix \hyperref[sec:appendix1]{B}.






\section{Experimental Results}

\textbf{Baseline.} In the experiments, we compare our Multi-task Group Fairness algorithm (MTGF) to the Infinite-horizon Group Fairness algorithm (IHGF) proposed by \citet{satija2023group}. The original IHGF algorithm imposes a fairness constraint on only one task, as it was designed for single-task settings. Applying this algorithm to multiple tasks leaves other tasks unconstrained, leading to violations of the fairness threshold. To establish a fairer comparison, we alternate the single-task constraint across the two tasks during training, making it a much stronger baseline than the original algorithm. 

\textbf{Environments.} We followed the customized environment from~\cite {satija2023group} alongside the standard Ant, Hopper, and Humanoid environments. Specifically, we modify the default Half-Cheetah-v3 from OpenAI Gym~\citep{brockman2016openaigym} to create three subgroups with distinct dynamics: a BigFoot Half-Cheetah with feet $2\times$ larger than the default, a LargeFriction Half-Cheetah with $30\times$ the friction of the default setting, and a HugeGravity Half-Cheetah with $1.5\times$ the default gravity.

\textbf{Tasks.} We consider two distinct tasks: in the forward running task, the agent is rewarded for maximizing its velocity in the forward direction, while in the backward running task, it is incentivized to move in the backward direction. To further emulate a realistic training scenario, we impose a task imbalance by sampling forward and backward-running episodes at a 1:3 ratio, respectively. Experiments were conducted between two social groups on the two tasks as detailed in Table \ref{tab:experiment_summary}.

\begin{table}[htbp]  
\centering
\small 
\setlength{\tabcolsep}{4pt} 


\begin{tabular}{@{}lccc@{}}
\toprule
\textbf{Experiment} & \textbf{Social Group A}      & \textbf{Social Group B}        & \textbf{Tasks}                   \\ \midrule
1                   & Ant        & Humanoid        & Backward, Forward Running         \\ \midrule
2                   &  Hopper        & Humanoid            & Backward, Forward Running         \\ \midrule
3                   & Hopper        & HugeGravity HalfCheetah        & Backward, Forward Running         \\ \midrule
4                   &  Original HalfCheetah        & HugeGravity HalfCheetah            & Backward, Forward Running         \\ \midrule
5                   &  Original HalfCheetah        & BigFoot HalfCheetah            & Backward, Forward Running         \\ \midrule
6                   &  Hopper        & LargeFricion HalfCheetah            & Backward, Forward Running         \\ \bottomrule
\end{tabular}
\caption{Summary of Experiments with HalfCheetah Variants Across Social Groups and Tasks}
\label{tab:experiment_summary}
\end{table}


\begin{table}[htbp] 
    \centering
    \begin{tabular}{lcccc}
        \toprule
        Groups & IHGF (MFV) & IHGF (SE) & \textbf{Ours} (MFV) & \textbf{Ours} (SE) \\
        \midrule
        Ant - Humanoid          & 406.34 & $\pm$16.11 & \textbf{336.09} & $\pm$13.23 \\
        Hopper - Humanoid       & 469.22 & $\pm$33.31 & \textbf{238.35} & $\pm$29.76 \\
        Hopper - HugeGravity    & 856.43 & $\pm$114.08 & \textbf{589.04} & $\pm$81.44 \\
        HalfCheetah - HugeGravity & 310.01 & $\pm$72.13 & \textbf{219.10} & $\pm$48.27 \\
        HalfCheetah - BigFoot   & 392.32 & $\pm$44.43 & \textbf{165.83} & $\pm$29.81 \\
        Hopper - LargeFric          & 805.44 & $\pm$105.26 & \textbf{488.09} & $\pm$78.12 \\
        \bottomrule
    \end{tabular}
    \caption{A comparison of Maximum Fairness Violation (MFV) over the fairness violations of the two tasks between the IHGF baseline and our Multi-Task Group Fairness (Ours) algorithm. Standard error of the fairness violations are also reported for each method.}
    \label{tab:mfv}
\end{table}


\textbf{Results.} As shown in Table~\hyperref[tab:mfv]{2}, our Multi-task Group Fairness (MTGF) algorithm consistently achieves a smaller maximum fairness gap across the two tasks compared to the single-task Group Fairness RL (IHGF) baseline. Notably, while GFRL may exhibit reasonable fairness on one task, it often violates fairness constraints substantially on the other task. By contrast, our approach effectively enforces fairness simultaneously on both tasks without significantly compromising mean returns. Additional plots and performance metrics provided in Appendix~\ref{sec:appendix2}.

\section{Conclusion}
In conclusion, this paper presents a comprehensive framework for achieving group fairness in multi-task reinforcement learning by formulating novel constrained optimization problems in both finite-horizon and infinite-horizon settings. Our approach rigorously extends single-task fairness concepts to multi-task environments, providing theoretical guarantees that ensure zero fairness violations with high probability and sublinear regret bounds, and practical algorithms for both the tabular setting and the deep reinforcement learning setting. Experiments on modified RiverSwim and continuous control environments further validate that our approach consistently achieves smaller fairness gaps in multiple tasks, without significantly compromising the overall performance, paving the way for more robust and socially responsible RL applications in real-world scenarios.

\subsubsection*{Acknowledgments}
This work was supported in part by the US National Science Foundation under grants III-2128019 and SLES-2331904. 
This work was also supported in part by the Coastal Virginia Center for Cyber Innovation (COVA CCI) and the Commonwealth Cyber Initiative (CCI), 
an investment in the advancement of cyber research $\text{\&}$ development, innovation, and workforce development. For more information about CCI, visit 
\url{www.covacci.org} and \url{www.cyberinitiative.org}.



\bibliography{main}
\bibliographystyle{rlj}


\beginSupplementaryMaterials

\appendix
\section{Algorithm for Finite-horizon MDP Problem}
\label{sec:appendix_tabular}
This appendix provides a detailed description of the LP-Based Algorithm for Multiple Tasks (Algorithm \ref{alg:lp_multitask}), designed to solve finite-horizon Markov Decision Process (MDP) problems across multiple tasks. The algorithm leverages empirical model updates and reward estimations to iteratively refine policy selection.

At each iteration, the algorithm updates the empirical estimates of the transition model and rewards, then computes optimistic and pessimistic reward estimates to guide decision-making. A policy is selected based on a comparison of performance across different tasks. If no predefined policy satisfies the performance criteria, an optimal policy is chosen to maximize cumulative rewards across tasks. The selected policy is then executed in the true environment, and the collected data is used to update future estimates.
\begin{widealgorithm}
    \caption{LP Based Algorithm for Multiple Tasks}
    \label{alg:lp_multitask}
    \KwIn{$\pi^0, \epsilon^0, \epsilon, K, \delta, M$ (Number of tasks)}
    
    \textbf{Initialize:} $N_h^m(s,a) = 0, \forall (s,a,h) \in S \times \mathcal{A} \times [H], \forall m \in \{1, \dots, M\}$.
    
    \For{$k = 1, \dots, K$}{
        Update the empirical estimates for the model of MDP $\hat{P}^{k}$; \\
        \For{$m = 1, \dots, M$}{
            Update the empirical estimates $\hat{r}^{k}_m, {\hat{r}^{\mathrm{opt}, k}}_m$\;
            Compute the optimistic and pessimistic reward estimates ${r}^{\mathrm{opt}, k}_m, \bar{r}^{k}_m, \underbar{r}^{k}_m$\;
        }
        Set $\pi^{k} \gets \text{Null}$\;

        \For{$m = 1, \dots, M$}{
            \For{$i \geq j; (i, j) \in \mathcal{Z}^2$}{
                \If{$J(\pi^0_i; \mu_i, \hat{P}_i^{k}, \bar{r}^{k}_m) - J(\pi^0_j; \mu_j, \hat{P}_j^{k}, \underbar{r}^{k}_m) > (\epsilon + \epsilon^0)/2$ \textbf{or} \\
                     $J(\pi^0_j; \mu_j, \hat{P}_j^{k}, \bar{r}^{k}_m) - J(\pi^0_i; \mu_i, \hat{P}_i^{k}, \underbar{r}^k_m) > (\epsilon + \epsilon^0)/2$}{
                    Set $\pi^{k} \gets \pi^0$\;
                }
            }
        } 
        \If{$\pi^{k} == \text{Null}$}{
            Set $\pi^{k} \gets \arg \max_{\pi \in \Pi^k} \sum_m^M J(\pi; \mu_i, \hat{P}^{k}_i, {r}^{\mathrm{opt}, k}_m)$\;
        }

        Execute $\pi^{k}$ in the true environment and collect a trajectory\;
        \[
        ({S}_h^{k}, A_h^{k}, r_{m,h}^{k}({S}_h^{k}, A_h^{k})), \forall h \in [H].
        \]
        Update counters $N_h({S}_h^{k}, A_h^{k}), \forall h \in [H]$\;
        
    }
\end{widealgorithm}

\section{First Order Constraint Policy Optimization Algorithm Extended for Multiple Constraints}
\label{sec:appendix1}

\paragraph{Algorithm \ref{algo:focops} (FOCOPS for \(\boldsymbol{M}\) Constraints)}
Algorithm \ref{algo:focops} extends First-Order Constrained Optimization in Policy Space to handle multiple constraints. It collects trajectories, estimates cost returns, updates the Lagrange multipliers for constraint satisfaction, and then updates the value functions and policy parameters. By enforcing a trust region (KL divergence \(\le \delta\)), it prevents large, destabilizing policy steps while ensuring all \(M\) constraints are satisfied.

\setcounter{algocf}{1}
\begin{widealgorithm}
    \caption{First-Order Constrained Optimization in Policy Space (FOCOPS) for $M$ Constraints}
    \label{algo:focops}
    \KwIn{Initial policy parameters $\theta^0$, initial value function parameters $\phi^0$, initial cost value function parameters $\{\psi^0_m\}_{m=1}^M$, Cost functions $\{C_m\}_{m=1}^M$, Cost tolerances $\{b_m\}_{m=1}^M$}
    \KwOut{Final policy parameters $\theta^\text{final}$, Final value function parameters $\phi^\text{final}$, Final cost value function parameters $\{\psi^\text{final}_m\}_{m=1}^M$}
    \textbf{Hyperparameters:} Discount rate $\gamma$, GAE parameter $\beta$, Learning rates $\alpha_\nu, \alpha_V, \alpha_\pi$, Temperature $\lambda$, Initial cost constraint parameter $\nu$, Cost constraint parameter bound $\nu_{\max}$, Trust region bound $\delta$

    \While{Stopping criteria not met}{
        Generate batch data of $H$ episodes of length $T$ of $(s_{i,t}, a_{i,t}, r_{i,t}, s_{i,t+1}, \{c_{m,i,t}\}_{m=1}^M)$ from $\pi_\theta$, 
        where $i=1,\ldots,H$, $t=1,\ldots,T$
        
        \For{$m = 1,\ldots,M$}{
            For cost function $m$, estimate cost-return by averaging over $C$-return for all episodes:
            \[\hat{J}_{C_m}=\frac{1}{M} \sum_{i=1}^M \sum_{t=0}^{T-1} \gamma^t c_{m,i,t}\]
        }
        Store old policy $\theta^{\prime} \leftarrow \theta$\\
        Estimate advantage functions $\hat{A}_{i,t}$ and $\{\hat{A}_{i,t}^{C_m}\}_{m=1}^M$, $i=1,\ldots,H$, $t=1,\ldots,T$ using GAE\\
        Get $V_{i,t}^{\text{target}}=\hat{A}_{i,t}+V_\phi(s_{i,t})$ and $V_{i,t}^{C_m,\text{target}}=\hat{A}_{i,t}^{C_m}+V_{\psi_m}^{C_m}(s_{i,t})$, for $m = 1,\ldots,M$
        
        \For{$m = 1,\ldots,M$}{
            Update $\nu_m$ by:
            $\nu_m \leftarrow \text{proj}_{\nu_m}[\nu_m-\alpha_{\nu_m}(b-\hat{J}_{C_m})]$
        }

        \For{$K$ epochs}{
            \For{each minibatch $\left\{s_j, a_j, A_j, \{A_j^{C_m}\}_{m=1}^M, V_j^{\text{target}}, \{V_j^{C_m,\text{target}}\}_{m=1}^M\right\}$ of size $B$}{
                Update value loss functions:
                $\mathcal{L}_V(\phi) =\frac{1}{2N} \sum_{j=1}^B(V_\phi(s_j)-V_j^{\text{target}})^2$

                \For{$m = 1,\ldots,M$}{
                    \[\mathcal{L}_{V^C_m}(\psi_m) =\frac{1}{2N} \sum_{j=1}^B(V_{\psi_m}^{C_m}(s_j)-V_j^{C_m,\text{target}})^2\]
                }
                Update value networks:
                $\phi \leftarrow \phi-\alpha_V \nabla_\phi \mathcal{L}_V(\phi)$

                \For{$m = 1,\ldots,M$}{
                    \[\psi_m \leftarrow \psi_m-\alpha_V \nabla_{\psi_m} \mathcal{L}_{V^{C_m}}(\psi_m)\]
                }
                Update policy:
                $\theta \leftarrow \theta-\alpha_\pi \hat{\nabla}_\theta \mathcal{L}_\pi(\theta)$,
                where
                \begin{align*}
                &\hat{\nabla}_\theta \mathcal{L}_\pi(\theta) \approx \frac{1}{B} \sum_{j=1}^B\Big[\nabla_\theta D_{\text{KL}}(\pi_\theta \| \pi_{\theta^{\prime}})[s_j]
                -\frac{1}{\lambda} \frac{\nabla_\theta \pi_\theta(a_j \mid s_j)}{\pi_{\theta^{\prime}}(a_j \mid s_j)}(\hat{A}_j-\sum_{m=1}^M\nu_m \hat{A}_j^{C_m})\Big] \\&\quad \quad \quad \quad \quad \cdot 
                \mathbf{1}_{D_{\text{KL}}(\pi_\theta \| \pi_{\theta^{\prime}})[s_j] \leq \delta}
                \end{align*}
                
                \If{$\frac{1}{HT} \sum_{i=1}^H \sum_{t=0}^{T-1} D_{\text{KL}}(\pi_\theta \| \pi_{\theta^{\prime}})[s_{i,t}] > \delta$}{
                    Break
                }
            }
        }
    }
\end{widealgorithm}

\paragraph{Algorithm \ref{algo:drl_mtgf} (Multi-Task Fairness RL)}
Algorithm \ref{algo:drl_mtgf} applies the multi-constraint FOCOPS procedure to achieve group fairness across multiple tasks. For each group \(z\), it samples trajectories, computes performance under several reward functions, and formulates fairness constraints (limiting inter-group differences by \(\epsilon\)). It then invokes FOCOPS to update that group's policy and cost functions. Repeating this for all groups produces a list of policies that maintain multi-task group fairness.

\setcounter{algocf}{2}
\begin{widealgorithm}
    \KwIn{Initial policy parameters $\theta_z^0, \forall z \in |\mathcal{Z}| $, initial value function parameters $\phi_z^0$, initial cost value function parameters $\psi_{m,z}^0, \forall z \in |\mathcal{Z}|, m \in 1, 2, ..., M$, where $M = (|Z| - 1)2N$.}
    \KwOut{Final policy parameters $\theta^{\text{final}}_z$, final value function $\phi_z^{\text{final}}$, and final cost function parameters $\{\psi_{m,z}^{\text{final}}\}_{m=1}^M$,} for each group $z$
    
    \textbf{Initialize:} Group fairness threshold $\epsilon$, M constraint funcitons \textbf{C}, M constraint thresholds $\textbf{b}$, $m = 1$.
    \For{$k = 0, 1, 2, \dots$}{
        \tcp{Calculate performance estimates of policies for all groups.}
         \For{$z \in |\mathcal{Z}|$}{
            \For{$z_1 \in |\mathcal{Z}|$}{
                \For{$i \in 1, 2, \dots, H$}{
                    \text{Sample the $i$th trajectory of length $T$ for group $z_1$: $(s_{i, t}, a_{i, t}, \{r_{n, i, t}\}_{n=1}^N, s_{i, t+1})$,}
                    \text{for $t = 1, \dots, T$.}
                    
                    \For {$n \in 1, 2, ..., N$}{
                        \text{Use the Monte Carlo Method to estimate the return $\bar{J}_z(r_n)$ of policy $\pi_z$ at}
                        \text{reward function $r_n$:}
                        
                         \[\bar{J}_z(r_n) = \frac{1}{H}\sum_{i=1}^H\sum_{t=0}^{T-1}\gamma^t r_{n, i, t}\]

                        \uIf{$z \neq z_1$}{
                        \text{Set the cost functions as the reward function and the negative reward}
                        \text{function:}
                            \[\textbf{C}[m] = r_n\]
                            \[\textbf{C}[m+1] = -r_n\] 
                        \text{Calculate the thresholds for M constraints:}
                            \[\textbf{b}[m] = \epsilon + \bar{J}_z(r_n)\]
                            \[\textbf{b}[m+1] = \epsilon - \bar{J}_z(r_n)\]
                            \[m=m+1\]
                        }

                }

            }

            }

            \text{Update the parameters for policy, value function, and cost functions of group $z$ by }
            \begin{center}
                    $\theta^{k+1}_z$, $\phi_z^{k+1}$, $\{\psi_{m,z}^{k+1}\}_{m=1}^M$ $    = \textbf{FOCOPS}(\theta^{k}_z$, $\phi_z^{k}$, $\{\psi_{m,z}^{k}\}_{m=1}^M, \textbf{C}, \textbf{b})$.
                \end{center}
        }

    }
    \caption{Outline of the Multi-Task Fairness RL Algorithm}
    \label{algo:drl_mtgf}
\end{widealgorithm}

\newpage
\section{Proofs for the Finite-Horizon Setting}
The theorems and lemmas presented in the paper are provided with full details in this appendix.

First, lets define some notations. Let $\{ \mathcal{F}_k\}_{k\leq 0}$ denotes the filtration with $\mathcal{F}_k = \sigma \left( (S^{k'}_{z, h}, A^{k'}_{z,h}, R^{k'}_{m,z, h})_{z \in \mathcal{Z}, h \in [H], m \in [M], k' \in [k]} \right) \forall k \in[K]$, and $\mathcal{F}_0$ denotes the trivial sigma algebra. 
The sequence of deployed policy $\{\pi^k\}_{k \in [K]}$ is predictable with respect to the filtration $\{\mathcal{F}_k\}_{k\leq 0}$. 

$N_{z, h}^k(s, a)$ denotes the number of times the state-action tuple $(s, a)$ for group $z$ was observed at time step $h$ in the episodes [1, \ldots, k-1]. The expectation operator $\mathbbm{E}_{\mu_z, P_z, \pi}[\cdot]$ is the expectation with respect to the stochastic trajectory $(S_h, A_h)_{h\in[H]}$ generated according to the markov chain induced by $(\mu_z, P_z, \pi)$. 

Additionally, we use $J^{\pi}_z(P_z, r)$ as a the notation short hand for the return $J(\pi_z; \mu_z, P_z, r)$.

\subsection{High Probability Good Event}
\label{high_prob_good_event}

Our subsequent analysis on performance guarantees depends on establishing a high probability "good" event $\boldsymbol{\mathcal{E}}$.

For each $(z, s, a, h) \in \mathcal{Z} \times \mathcal{S} \times \mathcal{A} \times [H]$, the empirical estimates of the transition is defined as:

\begin{align}
    \hat{P}_{z, h}^k (s' | s, a) \coloneqq \frac{\sum_{k' = 1}^{k-1}\mathbbm{1}(S_{z, h}^{k'} = s, A_h^{k'} = a, S^{k'}_{z, h+1} = s')}{\max(N_{z,h}^k(s, a), 1)}
\end{align}

We define the event $\mathcal{E}_\mathcal{G}$ for the event sequence $\mathcal{G}_k \in \mathcal{F}_{k-1}, \forall k \in [K]$:

\begin{equation}
\begin{aligned}
    \mathcal{E}_\mathcal{G}(\delta) \dot{=} &\left\{ \forall K' \in [K]. \right.\\
    &\sum^{K'}_{k=1} \sum^H_{h=1} \sum_{z, s, a} \frac{\mathbbm{1}(\mathcal{G}_k) d^{\pi^k}_{z,h}(s, a)}{\max(N^k_{z,h}(s, a), 1)} \leq 4H|\mathcal{Z}||S||A| + 2H|\mathcal{Z}||S||A|\ln K'_\mathcal{G} + 4\ln \frac{2HK}{\delta}, \\
    &\sum^{K'}_{k=1} \sum^H_{h=1} \sum_{z, s, a} \frac{\mathbbm{1}(\mathcal{G}_k) d^{\pi^k}_{z,h}(s, a)}{\sqrt{\max(N^k_{z,h}(s, a), 1)}} \leq  6H|\mathcal{Z}||S||A| + 2H\sqrt{|\mathcal{Z}||S||A|\ln K'_\mathcal{G}}  \\
    & \quad \quad \quad \quad \quad \quad \quad \quad \quad \quad \quad \quad \quad \quad \quad \enspace  + 2H|\mathcal{Z}||S||A|\ln K'_\mathcal{G} +   5\ln \frac{2HK}{\delta}, \left. \right\},
\end{aligned}
\end{equation}

where $K'_\mathcal{G} \dot{=} \sum^{K'}_{k=1} \mathbbm{1}(\mathcal{G}_k)$ and $d^{\pi^k}_z$ is the occupancy measure of policy $\pi^k$ such that $d^{\pi^k}_{z, h}(s, a) = \mathbbm{E}_{\mu_z, P_z, \pi^k}[\mathbbm{1}(S_{z, h} = s, A_h = a | \mathcal{F}_{k-1})]$.

Let $\mathcal{E}_\Omega(\delta)$ be the event with the event sequence $\mathcal{G}_k = \Omega, \forall k \in [K]$, where $\Omega$ is the sample space. let $\mathcal{E}_0(\delta)$ denote $\mathcal{E}_{\mathcal{G}'}$, for the event that we choose the strictly safe policy $\pi^0$,  with the event sequence
\begin{align}
    \mathcal{G'}_{1:K} = \left \{ J_i^{\pi^0}( \hat{P}^k_i, \bar{r}^k_m) - J_j^{\pi^0}(\hat{P}^k_j, \underbar{r}^k_m) \leq (\epsilon + \epsilon^0)/2, \forall i, j \in \mathcal{Z}^2, m \in [M] \right\}
\end{align}

Our subsequent analysis on performance guarantees depends on establishing a high probability "good" event $\boldsymbol{\mathcal{E}}$.

\textbf{Good Event} $\mathcal{E}$ is defined as:
\begin{equation}
\begin{aligned}
    \mathcal{E} \dot{=} &\Big \{ \forall k \in [K], \forall h \in [H], \forall z \in \mathcal{Z}, \forall s \in \mathcal{S}, \forall a \in \mathcal{A}, \\
    & |P^k_{z,h}(s'|s, a) - \hat{P}^k_{z,h}(s'|s, a)| \leq \beta^k_{z,h}(s,a), \forall s' \in \mathcal{S} \Big \} \cap \mathcal{E}_\Omega(\delta/4) \cap \mathcal{E}_0(\delta/4),
\end{aligned}
\end{equation}
where $\hat{\beta}^k_{z,h}(s, a) \coloneqq \sqrt{\frac{1}{\max(N^k_{z,h}(s, a), 1)}C} $ and $C \coloneqq \log(2|\mathcal{Z}|{|S|}^2|A|HK/\delta)$ 

\textbf{Lemma C.1} \textit{Fix any $\delta \in (0, 1)$, the good event $\mathcal{E}$ occurs with probability at least $1 - \delta$.}

\textit{Proof of Lemma C.1} For each $(z, s, a, h) \in \mathcal{Z} \times \mathcal{S} \times \mathcal{A} \times [H]$, we take $K$ mutually independent samples of next states from the distribution specified by the true MDP model: 
\begin{align}
    \{S^n_z(s, a, h)\}_{n=1}^K. 
\end{align}

Let $\hat{P}_{z,h}^n$ be running empirical means for the samples

\begin{align}
    \{S_z^i(s, a, h)\}^n_{i=1}.
\end{align}

We can define the failure event:

\begin{align}
    F^P_{n} \dot{=} \{ \exists z, s, a, s', h: |P_{z,h}(s'|s, a) - \hat{P}^n_{z,h}(s'|s, a)| \geq \beta(n)\},
\end{align}

We define a generated event $\mathcal{E}^{gen}$,

\begin{align}
    \mathcal{E}^{gen} \dot{=} \left(\cup^K_{n=1}(F^P_n)\right)^C \cap \mathcal{E}_\Omega(\delta/4) \cap \mathcal{E}_0(\delta/4)
\end{align}

Let $n_{z, k}(s, a, h)$ denote the quantity $N^k_{z, h}(s, a) + 1$. Then the problem in our setting can be simulated as follows: for group $z$, at an episode $k$, taking action $a$ in state $s$ at time-step $h$, we get the sample $(S_z^{n_{z, k}(s, a, h)}(s, a, h))$. Therefore, the set
\begin{align}
    \{S_z^n(s, a, h) \}_{n=1}^K
\end{align}
already contains all the samples drawn in the learning problem and the sample averages calculated by the algorithms are:
\begin{align}
    \hat{P}^k_{z,h}(s'|s, a) = P_z^{n_k(z, \tilde{s}, a, h)}(\cdot | s, a, h).
\end{align}

As a result, the $\mathcal{E}^{gen}$ implies $\mathcal{E}$, and it is sufficient to show that $\mathcal{E}^{gen}$ occurs with probability at least $1 - \delta$.

Using \hyperref[Lemma H.4]{Lemma 8} and union bound, $\mathcal{E}_\Omega(\delta/4) \cap \mathcal{E}_0(\delta/4)$ occurs with probability at least $1 - \delta/2$. To see this, let A denotes $\mathcal{E}_\Omega(\delta/4)$ and let B denotes $\mathcal{E}_0(\delta/4)$. By \hyperref[Lemma H.4]{Lemma 5}, $\Pr(A) = 1 - \delta/4$ and $\Pr(B) = 1 - \delta/4$
\begin{equation}
\begin{aligned}
    \Pr(A\cap B) &= \Pr(A) + \Pr(B) - \Pr(A \cup B) \\
    & \geq \Pr(A) + \Pr(B) - 1 \\
    &= 1 - \delta/4 + 1 - \delta/4 - 1 \\
    &= 1 - \delta/2
\end{aligned}
\end{equation}

For the failure event $F^P_n$, by Hoeffding's inequality in \hyperref[Lemma Hoeffding's inequality]{Lemma 3} and Union Bound, we have:

\begin{equation}
\begin{aligned}
      \Pr(\cup_{n=1}^K F^P_n) &\leq \sum_n^K\sum_{z \in \mathcal{Z}}\sum_{s \in \mathcal{S}} \sum_{a \in \mathcal{A}}  \sum_{h}^H \sum_{s' \in \mathcal{S}}\exp(-n(\beta(n))^2) \\ 
    &=\sum_n^K \sum_{z \in \mathcal{Z}} \sum_{s \in \mathcal{S}} \sum_{a \in \mathcal{A}}  \sum_{h}^H \sum_{s' \in \mathcal{S}} \exp\left(-n \cdot \sqrt{\frac{1}{{\max(n,1)}\log(2|\mathcal{Z}|{|S|}^2|A|HK/\delta)}}^2\right) \\
    & =K |\mathcal{Z}||\mathcal{S}|^2|\mathcal{A}|H \frac{\delta}{2|\mathcal{Z}|{|S|}^2|A|HK} \\
    & = \delta/2
\end{aligned}
\end{equation}

The event $(\cup_{n=1}^K F^P_n)^C$ occurs with probability at least $1 - \delta/2$. Combining the results we have $\Pr(\mathcal{E}^\text{gen}) = \Pr((\cup_{n=1}^K F^P_n)^C \cap \mathcal{E}_\Omega(\delta/4) \cap \mathcal{E}_0(\delta/4)) \leq 1 - \delta$, which implies $\mathcal{E}$ occurs with probability at least $1-\delta$. 

\subsection{Proof for Theorem 1.1}
Now, we are ready to present the proof for \hyperref[thm:fairness_violation]{Theorem 1.1}. Without loss of generality, let $\{i, j\}$ denote any pair of subgroups in $\mathcal{Z}^2$. $\mathbf{\Pi}^k$ consists of either the singleton set $\{\pi^0\}$ or the selected policies $\mathbf{\Pi}^k_\text{F}$ defined in Equation (\ref{eq:selecting_fair_policies}).   For $\pi^0$, we have $|J_i^{\pi^0}(r_m, P_i) - J_j^{\pi^0}(r_m, P_j)| \leq \epsilon, \forall m \in [M]$ by definition of initial fair policy (\hyperref[assumption:initial_fair_policy]{Assumption 1.1}). We will now show that our construction of $\mathbf{\Pi}_F^k$ also satisfies the zero constraint violation property for any such pair of subgroups. For $\pi \in \mathbf{\Pi}^k_\text{F}$, to show $|J_i^{\pi}(r_m, P_i) - J_j^{\pi}(r_m, P_j)| \leq \epsilon, \forall m \in [M]$ holds under the good event, we will first show $J_i^{\pi}(r_m, P_i) - J_j^{\pi}(r_m, P_j) \leq \epsilon, \forall m \in [M]$, i.e. the return of group $i$ is no more than the return of group $j$ by $\epsilon$ for all tasks $m$ in part 1, and then show $J_j^{\pi}(r_m, P_j) - J_i^{\pi}(r_m, P_i) \leq \epsilon, \forall m \in [M]$, i.e. the return of group $j$ is no more than the return of group $i$ by $\epsilon$ for all tasks $m$ in part 2. 

\textbf{Part 1:} In the first part of the proof, we will show that on the good event $\mathcal{E}$, for any $k \in [K]$ and policy $\pi \in \mathbf{\Pi}_F^k$,
\begin{align}
    J_i^{\pi}(r_m, P_i) - J_j^{\pi}(r_m, P_j) \leq \epsilon, \forall m \in [M].
\end{align}

\textit{Proof.} Using \hyperref[lemma:optimistic return]{Lemma 1}, we have:
\begin{equation}
    \label{eq:optimistic return of group i}
    J_i^{\pi}(r_m, P_i) \leq J_i^{\pi}(\bar{r}_m^k, \hat{P}_i^k), \forall m \in [M].
\end{equation}
Similarly, using \hyperref[lemma:pessimistic return]{Lemma 2}, we get
\begin{equation}
    \label{eq:pessimistic return of group j}
    -J_j^{\pi}(r_m, P_j) \leq -J_j^{\pi}(\underbar{r}_m^k, \hat{P}_j^k), \forall m \in [M].
\end{equation}
Combining Equation (\ref{eq:optimistic return of group i}) and Equation (\ref{eq:pessimistic return of group j}), we have:
\begin{align}
    J_i^{\pi}(r_m, P_i) - J_j^{\pi}(r_m, P_j) \leq J_i^{\pi}(\bar{r}_m^k, \hat{P}_i^k) - J_j^{\pi}(\underbar{r}_m^k, \hat{P}_j^k), \forall m \in [M].
\end{align}
Note that from the definition of $\Pi_F^k$ in Equation (\ref{eq:selecting_fair_policies}), we know any policy in $\pi \in \Pi_F^k$ satisfies the constraint:
\begin{align}
    J_i^{\pi}(\bar{r}_m^k, \hat{P}_i^k) - J_j^{\pi}(\underbar{r}_m^k, \hat{P}_j^k) \leq \epsilon, \forall m \in [M].
\end{align}
Therefore, we have the following relation:
\begin{align}
    J_i^{\pi}(r_m, P_i) - J_j^{\pi}(r_m, P_j) \leq J_i^{\pi}(\bar{r}_m^k, \hat{P}_i^k) - J_j^{\pi}(\underbar{r}_m^k, \hat{P}_j^k) \leq \epsilon, \forall m \in [M].
\end{align}

\textbf{Part 2:} In the first part of the proof, we will show that on the good event $\mathcal{E}$, for any $k \in [K]$ and policy $\pi \in \mathbf{\Pi}_F^k$,
\begin{align}
    J_j^{\pi}(r_m, P_j) - J_i^{\pi}(r_m, P_i) \leq \epsilon, \forall m \in [M].
\end{align}

\textit{Proof.} Using \hyperref[lemma:optimistic return]{Lemma 1}, we have:
\begin{equation}
    \label{eq:optimistic return of group j}
    J_j^{\pi}(r_m, P_j) \leq J_j^{\pi}(\bar{r}_m^k, \hat{P}_j^k), \forall m \in [M].
\end{equation}
Similarly, using \hyperref[lemma:pessimistic return]{Lemma 2}, we get
\begin{equation}
    \label{eq:pessimistic return of group i}
    -J_i^{\pi}(r_m, P_i) \leq -J_i^{\pi}(\underbar{r}_m^k, \hat{P}_i^k), \forall m \in [M].
\end{equation}
Combining Equation (\ref{eq:optimistic return of group j}) and Equation (\ref{eq:pessimistic return of group i}), we have:
\begin{align}
    J_j^{\pi}(r_m, P_j) - J_i^{\pi}(r_m, P_i) \leq J_j^{\pi}(\bar{r}_m^k, \hat{P}_j^k) - J_i^{\pi}(\underbar{r}_m^k, \hat{P}_i^k), \forall m \in [M].
\end{align}
Note that from the definition of $\Pi_F^k$ in Equation (\ref{eq:selecting_fair_policies}), we know any policy in $\pi \in \Pi_F^k$ satisfies the constraint:
\begin{align}
    J_j^{\pi}(\bar{r}_m^k, \hat{P}_j^k) - J_i^{\pi}(\underbar{r}_m^k, \hat{P}_i^k) \leq \epsilon, \forall m \in [M].
\end{align}
Therefore, we have the following relation:
\begin{align}
    J_j^{\pi}(r_m, P_j) - J_i^{\pi}(r_m, P_i) \leq J_j^{\pi}(\bar{r}_m^k, \hat{P}_j^k) - J_i^{\pi}(\underbar{r}_m^k, \hat{P}_i^k) \leq \epsilon, \forall m \in [M].
\end{align}

Combining the results of $\mathbf{\Pi}^k$ being the singleton set $\{\pi^0\}$ or $\mathbf{\Pi}^k_\text{F}$, we have for $\pi \in \mathbf{\Pi}^k$, 

\begin{align}
    |J_i^{\pi}(r_m, P_i) - J_j^{\pi}(r_m, P_j)| \leq \epsilon, \forall m \in [M], \forall k \in [K],
\end{align}
which holds for any pair of group $\{i, j\} \in \mathcal{Z}^2$. Extending to all pairs of groups: 
\begin{align}
    |J_i^{\pi}(r_m, P_i) - J_j^{\pi}(r_m, P_j)| \leq \epsilon, \forall m \in [M], \forall k \in [K], \forall i \geq j; (i, j) \in \mathcal{Z}^2.
\end{align}

\subsection{Proof for Theorem 1.2}

From the definition of the conservative set of policies in Equation (\ref{eq:conservative set of policies}), we will apply $\pi^0$ when there exist one pair of groups $(i, j) \in \mathcal{Z}^2$ and one task $m \in [M]$ such that the return difference under a optimistic MDP and a pessimistic MDP is greater than or equal to $\frac{\epsilon + \epsilon^0}{2}$. In this case, $|\Pi^k| = |\{\pi^0\}| = 1$. By the \hyperref[assumption:initial_fair_policy]{Assumption 1.1}, we have $\epsilon^0 < \epsilon$ and therefore $\frac{\epsilon+\epsilon^0}{2} < \epsilon$. When the return difference of applying $\pi^0$ for all pair of groups $(i, j) \in \mathcal{Z}^2$ and for all task $m \in [M]$ is less than or equal to $\frac{\epsilon+\epsilon^0}{2}$, which is strictly less than $\epsilon$, then there exist infinitely many policies that are close to $\pi^0$ that can result in a return difference less than $\epsilon$ and thus satisfy the constraint in Equation (\ref{eq:selecting_fair_policies}). In this case, $|\Pi^k|=|\Pi^k_F| > 1$. 

We can follow \cite{liu2021} and break down the regret according to the above two cases: $|\Pi^k|=1$ and $|\Pi^k| > 1$. For all task $m$, the regret can be broken down in into three terms. Providing upper bounds for each of the three terms by Lemma A.1, Lemma A.2 and Lemma A.3 will conclude our regret analysis.



\begin{align*}
\text{Reg}(K; r_m) &= \sum_{k=1}^{K} \mathbbm{1}(|\Pi^k| = 1) (J^{\pi^*}(r_m, P) - J^{\pi^0}(r_m, P)) \quad &(\text{I}) \\
&\quad + \sum_{k=1}^{K} \mathbbm{1}(|\Pi^k| > 1) (J^{\pi^*}(r_m, P) - J^{\pi^k}( {r^{\mathrm{opt}}}_m^k, \hat{P}^k)) \quad &(\text{II}) \\
&\quad + \sum_{k=1}^{K} \mathbbm{1}(|\Pi^k| > 1) (J^{\pi^k}({r^{\mathrm{opt}}}_m^k, \hat{P}^k) - J^{\pi^k}(r_m, P)) \quad &(\text{III})
\end{align*}

\textbf{Lemma A.1}(Similar to lemma C.6 in \cite{satija2023group}) \textit{On good event $\mathcal{E} $}, 

\begin{align}
    \sum_k^K \mathbbm{1}(|\Pi^k|=1) 
    &\leq \tilde{\mathcal{O}}\left (\frac{|\mathcal{Z}|^2 M^4H^4|\mathcal{S}|^3|\mathcal{A}|}{(\epsilon - \epsilon^0)\min\{1, (\epsilon - \epsilon^0) } \right ).
\end{align}

\textit{Proof.} For part I, we want to obtain an upper bound for $\sum_k^K (|\Pi^k|=1)$. We start by giving an upper bound for $\sum_k^K \mathbbm{1}(|\Pi^k|=1; (i, j), m)$, which denotes when two particular groups $i, j$ led to the fairness violation in task $m$.  In this case, when the fairness constraint is violated with respect to $\pi^0$, either group $i$'s return is much larger than group $j$'s return as in the following Case $A_{(i, j), m}$, or group $j$'s return is much larger than group $i$'s return as in Case $B_{(i, j), m}$.  


\begin{align}
    \intertext{Case $A_{(i, j), m}$} J^{\pi^k}_i( r_m, P_i) - J^{\pi^k}_j(r_m, P_j) \geq (\epsilon + \epsilon^0)/2 \\
    \intertext{Case $B_{(i, j), m}$} J^{\pi^k}_j(r_m, P_j) - J^{\pi^k}_i(r_m, P_i) \geq (\epsilon + \epsilon^0)/2
\end{align}

 We define $K' = \sum_k^K \mathbbm{1}(|\Pi^k|=1; (i, j), m)$.


\begin{equation}
\begin{split}
\left(\frac{\varepsilon - \varepsilon^0}{2}\right) K' &= \sum_{k=1}^{K} \mathbbm{1}(|\Pi^k| = 1; (i, j), m)\left(\frac{\varepsilon - \varepsilon^0}{2}\right)\\
&= \sum_{k=1}^{K} \mathbbm{1}(|\Pi^k| = 1; (i, j), m)\left(\frac{\varepsilon + \varepsilon^0}{2} - \varepsilon^0\right) \\
&\leq \sum_{k=1}^{K} \mathbbm{1}(|\Pi^k| = 1; A_{(i,j), m})\left(\frac{\varepsilon + \varepsilon^0}{2} - \varepsilon^0\right)\\
&\quad + \sum_{k=1}^{K} \mathbbm{1}(|\Pi^k| = 1; B_{(i,j), m})\left(\frac{\varepsilon + \varepsilon^0}{2} - \varepsilon^0\right)
\end{split}
\end{equation}

For Case $A_{(i, j), m}$:

\begin{equation}
\begin{aligned}
&\sum_{k=1}^{K} \mathbbm{1}(|\Pi^k| = 1; A_{(i, j), m})\left(\frac{(\varepsilon + \varepsilon^0)}{2} - \varepsilon^0\right)\\
&\leq \mathbbm{1}(|\Pi^k| = 1; A_{(i, j), m})\left((J^{\pi^k_i}(r^k, \hat{P}^k) - J^{\pi^k_j}(r_m^k, \hat{P}^k)) - (J^{\pi^k_i}(r_m, P) - J^{\pi^k_j}(r_m, P))\right) \\
&= \underbrace{\mathbbm{1}(|\Pi^k| = 1; A_{(i, j), m})(J^{\pi^k_i}(r^k_m, \hat{P}^k) - J^{\pi^k_i}(r_m, P))}_{(A.1)}\\
& \quad + \underbrace{\mathbbm{1}(|\Pi^k| = 1; A_{(i, j), m})(J^{\pi^k_j}(r_m, P) - J^{\pi^k_j}(r_m^k, \hat{P}^k))}_{(A.2)},
\end{aligned}
\end{equation}

For the first term, we use \hyperref[lemma:H3]{Lemma 5} with the designed optimistic reward function from Equation (\ref{eq:optimistic reward})
\begin{equation}
\begin{aligned}
    |\bar{r}^k_{m,h} - r_{m,h}| &= |\alpha \beta_{m,h}^k| \\
    &\leq (|\mathcal{S}|H)\beta_{m,h}^k,
\end{aligned}
\end{equation}

Plugging in $\alpha=|\mathcal{S}H|$ in \hyperref[lemma:H3]{Lemma 5}, the first term A.1 is bounded by 

\begin{align}
    A.1 =\tilde{\mathcal{O}}(H^4|\mathcal{S}|^3|\mathcal{A}| + H^2\sqrt{|\mathcal{S}|^3|\mathcal{A}|K'})
\end{align}

For the second term A.2, we use the following relation from the designed pessimistic reward function from Equation (\ref{eq:pessimistic reward})
\begin{align}
    |r_{m,h} - \underbar{r}_{m, h}^k| &= |-(-\alpha \beta_{m,h}^k)| \\
    &\leq (|\mathcal{S}|H)\beta_{m,h}^k
\end{align}
Applying \hyperref[lemma:H3]{Lemma 5},
\begin{align}
    A.2 =\tilde{\mathcal{O}}(H^4|\mathcal{S}|^3|\mathcal{A}| + H^2\sqrt{|\mathcal{S}|^3|\mathcal{A}|K'})
\end{align}

Therefore, 
\begin{align}
    \sum_{k=1}^{K} \mathbbm{1}(|\Pi^k| = 1; A_{(i,j), m})\left(\frac{(\varepsilon + \varepsilon^0)}{2} - \varepsilon^0\right) = \tilde{\mathcal{O}}(H^4|\mathcal{S}|^3|\mathcal{A}| + H^2\sqrt{|\mathcal{S}|^3|\mathcal{A}|K'})
\end{align}

Since case A and case B are symmetric with respect to the two groups $i$ and $j$, we can follow the above steps and obtain the same big O notation for case B. 

\begin{align}
    \sum_{k=1}^{K} \mathbbm{1}(|\Pi^k| = 1; B_{(i,j),m})\left(\frac{(\varepsilon + \varepsilon^0)}{2} - \varepsilon^0\right) = \tilde{\mathcal{O}}(H^4|\mathcal{S}|^3|\mathcal{A}| + H^2\sqrt{|\mathcal{S}|^3|\mathcal{A}|K'})
\end{align}

Combining results for Case A and Case B, we will have the same big O notation for $K'$.

\begin{align}
    \frac{(\epsilon + \epsilon^0)}{2}K' = 
    \tilde{\mathcal{O}}(H^4|\mathcal{S}|^3|\mathcal{A}| + H^2\sqrt{|\mathcal{S}|^3|\mathcal{A}|K'})
\end{align}

By \hyperref[lemma:H5]{Lemma 7} (Lemma D.6 in \cite{liu2021}),
\begin{align}
    K' = \sum_k^K \mathbbm{1}(|\Pi^k|=1; (i, j), m) \leq \tilde{\mathcal{O}}\left (\frac{H^4|\mathcal{S}|^3|\mathcal{A}|}{(\epsilon - \epsilon^0)\min\{1, (\epsilon - \epsilon^0) } \right )
\end{align}

Now, to obtain the upper bound for fairness violation by any possible pairs of groups and for all tasks $\sum_k^K \mathbbm{1}(|\Pi^k|=1)$, by the union bound we have

\begin{align}
    \sum_k^K \mathbbm{1}(|\Pi^k|=1) &\leq \sum_{i, j \in \mathcal{Z}^2}\sum_m^M \sum_k^K \mathbbm{1}(|\Pi^k|=1; (i, j), m) \\
    &\leq |\mathcal{Z}|^2 M K' \\
    &\leq \tilde{\mathcal{O}}\left (\frac{|\mathcal{Z}|^2 MH^4|\mathcal{S}|^3|\mathcal{A}|}{(\epsilon - \epsilon^0)\min\{1, (\epsilon - \epsilon^0) } \right ).
\end{align}

\textbf{Lemma A.2} \textit{For $\alpha_l = |S|H + \frac{8M^2|S|H^2}{\epsilon-\epsilon^0}$, on good event $\mathcal{E}$,} 
\begin{align}
    \sum_{k=1}^{K} \mathbbm{1}(|\Pi^k| > 1) (J^{\pi^*}(r_m, P) - J^{\pi^k}(\bar{r}_m, \hat{P}^k)) \leq 0
\end{align}

\textit{Proof.} When $\pi^* \in \Pi^k$, the inequality holds because of the reward bonus and the fact that $\pi^k$ maximizes the optimistic CMDP from $\eqref{eq:finite-horizon practical problem}$. 

When $\pi^* \not\in \Pi^k$, 
we first show the difference in cost is less or equal to 0 for any pair of groups $i, j$, then it holds for all groups.

Let $B_{\gamma_k}$ denote an independent Bernoulli distributed random variable with mean $\gamma_k$. We can define a probability mixed policy $\tilde{\pi}^k$ as:
\begin{align}
    \tilde{\pi} = B_{\gamma_k} \pi^* + (1-B_{\gamma_k}) \pi^0
\end{align}

Let $\gamma_k \in [0, 1]$ be the largest coefficient that ensures the constraint is not violated by the mixed policy $\tilde{\pi}^k$,

\begin{align}
    J^{\tilde{\pi}}_i(\bar{r}_m^k, \hat{P}_i^k) - J^{\tilde{\pi}}_j(\bar{r}_m^k, \hat{P}_j^k) \leq \epsilon
     \label{ineq: mixed policy constraint}
\end{align}

If $J^{\pi^*}_i(\bar{r}_m^k, \hat{P}_i^k) - J^{\pi^*}_j(\underbar{r}_m, \hat{P}^k_j) < \epsilon$, then $\gamma_k=1$. Else, we will obtain a $\gamma_k$ that make the equality hold for (\ref{ineq: mixed policy constraint}).

Denote the pessimistic cost of the difference in value between the two groups as:

\begin{align}
    \tilde{J}^\pi_{i,j} := J^\pi_i(  \bar{r}_m^k, \hat{P}_i^k) - J^\pi_j(  \underbar{r}^k_m, \hat{P}^k_j), 
\end{align}
where $\pi$ could be $\pi^*$ or $\pi^0$, and denote the difference in value in the true MDP as
\begin{align}
    J^\pi_{i,j} := J^\pi_i(r_m, P_i) - J^\pi_j( r_m, P_j)
\end{align}
When the equality holds, we have
$$
\begin{aligned}
&\epsilon= \gamma_k \tilde{J}^{\pi^*}_{i,j} + (1-\gamma_k) \tilde{J}^{\pi^0}_{i, j}\\
&\leq \gamma_k \tilde{J}^{\pi^*}_{i,j} + (1-\gamma_k)\frac{\epsilon + \epsilon^0}{2} \\
&= \gamma_k ( \tilde{J}^{\pi^*}_{i,j} - J^{\pi^*}_{i,j}) + \gamma_k J^{\pi^*}_{i,j} + (1-\gamma_k)\frac{\epsilon + \epsilon^0}{2} \\
&\leq \gamma_k ( \tilde{J}^{\pi^*}_{i,j} - J^{\pi^*}_{i,j}) + \gamma_k \epsilon + \frac{\epsilon + \epsilon^0}{2} -\gamma_k\frac{\epsilon + \epsilon^0}{2} \\
&\leq \gamma_k ( \tilde{J}^{\pi^*}_{i,j} - J^{\pi^*}_{i,j} + \frac{\epsilon - \epsilon^0}{2}) +\frac{\epsilon + \epsilon^0}{2}
\end{aligned}
$$

Using \hyperref[lemma:optimistic return]{Lemma 1} and \hyperref[lemma:pessimistic return]{Lemma 2}, we have
\begin{align}
    J^{\pi}_i(r_m, P_i) \leq J^{\pi}_i(  \bar{r}^k_m, \hat{P}_i^k), \label{ineq: A2 part 1}
\end{align}
and 
\begin{align}
    -J^{\pi^*}_j(r^k_m, \hat{P}^k_j) \leq -J^{\pi^*}_j(\underbar{r}_m, P_j). \label{ineq: A2 part 2}
\end{align}

Adding (\ref{ineq: A2 part 1}) and (\ref{ineq: A2 part 2}), 
\begin{align}
    \tilde{J}^{\pi^*}_{i, j} - J^{\pi^*}_{i, j} \geq 0.
\end{align}

Since $\epsilon > \epsilon^0$, $\tilde{J}^{\pi^*}_{i, j} - J^{\pi^*}_{i, j} + \frac{\epsilon - \epsilon^0}{2} \geq 0$. Therefore, 

\begin{align}
    \gamma_k \geq \frac{\epsilon - \epsilon^0}{\epsilon - \epsilon^0 + 2(\tilde{J}^{\pi^*}_{i, j} - J^{\pi^*}_{i, j})}
\end{align}

Using \hyperref[lemma:optimistic return diff]{Lemma 3} and \hyperref[lemma:pessimistic return diff]{Lemma 4} , we have
\begin{align}
    J_i^{\pi}(\bar{r}_m, \hat{P}_i^k) - J_i^{\pi}(r_m, P_i) \leq 2(|\mathcal{S}|H) J^{\pi}_i(\beta_{m,h}^k(s,a), \hat{P}_i^k), \label{ineq: A2 part 3}
\end{align}
and 
\begin{align}
     J^{\pi}_j(r_m, P_j) - J^{\pi}_j(\underbar{r}_m, \hat{P}^k_j) \leq 2(|\mathcal{S}|H) J^\pi_j(\beta_{m,h}^k(s,a), \hat{P}^k_j). \label{ineq: A2 part 4}
\end{align}

Adding (\ref{ineq: A2 part 3}) and (\ref{ineq: A2 part 4}), 

\begin{align}
    \tilde{J}^\pi_{i, j} - J^\pi_{i, j} \leq 2(|\mathcal{S}|H) \left(J^\pi_i( \beta_{m,h}^k(s,a), \hat{P}^k) + J^\pi_j( \beta_{m,h}^k(s,a), \hat{P}^k_j)\right).
\end{align}

Because $\pi^k$ is the optimal policy in the optimistic CMDP, we have:
\begin{equation}
\begin{aligned}
    J^{\pi^k}_i({r^{\mathrm{opt}}}_m, \hat{P}^k_i) + J^{\pi^k}_j({r^{\mathrm{opt}}}_m, \hat{P}^k_j) 
    &\geq J^{\hat{\pi}^k}_i({r^{\mathrm{opt}}}_m, \hat{P}^k_i) + J^{\hat{\pi}^k}_j({r^{\mathrm{opt}}}_m, \hat{P}^k_j) \\
    &= J^{\tilde{\pi}^k}_i({r^{\mathrm{opt}}}_m, \hat{P}^k_i) + J^{\tilde{\pi}^k}_j({r^{\mathrm{opt}}}_m, \hat{P}^k_j) \\
    &=\gamma_k (J^{{\pi^*}^k}_i({r^{\mathrm{opt}}}_m, \hat{P}^k_i) + J^{{\pi^*}^k}_j(\ddot{r}_m, \hat{P}^k_j)) \\ 
    &\quad + \underbrace{(1-\gamma_k)(J^{{\pi^0}^k}_i({r^{\mathrm{opt}}}_m, \hat{P}^k_i) + J^{{\pi^0}^k}_j({r^{\mathrm{opt}}}_m, \hat{P}^k_j))}_{\geq 0} \\
    &\geq \gamma_k (J^{{\pi^*}^k}_i({r^{\mathrm{opt}}}_m, \hat{P}^k_i) + J^{{\pi^*}^k}_j({r^{\mathrm{opt}}}_m, \hat{P}^k_j)) \\
    &\geq \frac{\epsilon - \epsilon^0}{\epsilon - \epsilon^0 + 2(\tilde{J}^{\pi^*}_{i, j} - J^{\pi^*}_{i, j})} \\ & \quad \cdot (J^{{\pi^*}^k}_i({r^{\mathrm{opt}}}_m, \hat{P}^k_i) + J^{{\pi^*}^k}_j({r^{\mathrm{opt}}}_m, \hat{P}^k_j)) \\
    &\geq \frac{\epsilon - \epsilon^0}{\epsilon - \epsilon^0 + 4|\mathcal{S}|H \left(J^\pi_i(\beta_h^k(s,a), \hat{P}_i^k) + J^\pi_j(\beta_h^k(s,a), \hat{P}^k_j)\right)} \\ & \quad \cdot (J^{{\pi^*}^k}_i({r^{\mathrm{opt}}}_m, \hat{P}^k_i) + J^{{\pi^*}^k}_j({r^{\mathrm{opt}}}_m, \hat{P}^k_j))
\end{aligned}
\end{equation}

To make $J^{\pi^k}_i({r^{\mathrm{opt}}}_m, \hat{P}^k_i) + J^{\pi^k}_j({r^{\mathrm{opt}}}_m, \hat{P}^k_j) \leq J^{\pi^*}_i(r_m, P_i) + J^{\pi^*}_j(r_m, P_j)$, it is sufficient to show

\begin{equation}
\begin{aligned}
    &\frac{\epsilon - \epsilon^0}{\epsilon - \epsilon^0 + 4|\mathcal{S}|H \left(J^\pi_i(\beta_h^k(s,a), \hat{P}_i^k) + J^\pi_j(\beta_h^k(s,a), \hat{P}^k_j)\right)} \left(J^{\pi^k}_i({r^{\mathrm{opt}}}_m, \hat{P}^k_i) + J^{\pi^k}_j({r^{\mathrm{opt}}}_m, \hat{P}^k_j)\right) \\
    &\geq J^{\pi^*}_i(r_m, P_i) + J^{\pi^*}_j(r_m, P_j),
\end{aligned}
\end{equation}
which is equivalent to 

\begin{equation}
\begin{aligned}
    \label{A2 ineq to show}
    (\epsilon - \epsilon^0)\left(\left(J^\pi_i({r^{\mathrm{opt}}}_{m,h}(s,a), \hat{P}_i^k) + J^\pi_j({r^{\mathrm{opt}}}_{m,h}(s,a), \hat{P}^k_j)\right) - \left(  J^{\pi^*}_i(r_m, P_i) + J^{\pi^*}_j(r_m, P_j)\right)\right) \\
    \geq 4|\mathcal{S}|H \left(J^\pi_i(\beta_{m,h}^k(s,a), \hat{P}_i^k) + J^\pi_j(\beta_{m,h}^k(s,a), \hat{P}^k_j)\right)\left(  J^{\pi^*}_i(r_m, P_i) + J^{\pi^*}_j(r_m, P_j)\right) 
\end{aligned}
\end{equation}

From the value difference lemma (\hyperref[lemma:value difference lemma]{Lemma 6}), for any group $z \in \mathcal{Z}$, 
\begin{equation}
\begin{aligned}
    &J^{\pi^*}_z({r^{\mathrm{opt}}}_m, \hat{P}^k_z) - J^{\pi^*}_z(r_m, P_z) \\
    &=\mathbbm{E}\Biggl[\sum_{h=1}^H \Bigl({r^{\mathrm{opt}}}_m(s_h, a_h) - r_m(s_h, a_h) \\ &\quad \quad + \sum_{s'}\bigl(\hat{P}^k_{z,h} - P_{z,h}\bigr)(s'|s_h,a_h) V^{\pi^*_z}_{h+1}\bigl(s';\sum_m r_m, P_{z,h}\bigr)\Bigr) \Biggr| \mathcal{F}_{k-1}\Biggr] \\
    &\geq \mathbbm{E} \left[\sum_{h=1}^H(\alpha_l-|\mathcal{S}|H)\beta_{m,h}^k(s_h, a_h) \Big| \mathcal{F}_{k-1}\right] \\
    & =(\alpha_l-|\mathcal{S}|H)J^{\pi^*}_z(\beta_m^k, \hat{P}^k_z).
\end{aligned}
\end{equation}

Using the above result for group $i$ and $j$ seperately, we have
\begin{align}
    J^{\pi^*}_i({r^{\mathrm{opt}}}_m, \hat{P}^k_i) - J^{\pi^*}_i(r_m, P_i) \geq (\alpha_l-|\mathcal{S}|H)J^{\pi^*}_i(\beta_m^k, \hat{P}^k_i).
\end{align}
\begin{align}
    J^{\pi^*}_j({r^{\mathrm{opt}}}_m, \hat{P}^k_j) - J^{\pi^*}_j(r_m, P_j) \geq (\alpha_l-|\mathcal{S}|H)J^{\pi^*}_j(\beta_m^k, \hat{P}^k_j).
\end{align}

Adding the above two inequalities, 

\begin{equation}
\begin{aligned}
    \left(J^\pi_i({r^{\mathrm{opt}}}_{m,h}(s,a), \hat{P}_i^k) + J^\pi_j({r^{\mathrm{opt}}}_{m,h}(s,a), \hat{P}^k_j)\right) - \left(  J^{\pi^*}_i(r_m, P_i) + J^{\pi^*}_j(r_m, P_j)\right) \geq \\ (\alpha_l-|\mathcal{S}|H)(J^{\pi^*}_i(\beta_m^k, \hat{P}^k_i) + J^{\pi^*}_j(\beta_m^k, \hat{P}^k_j))
\end{aligned}
\end{equation}

Letting $\alpha_l = |\mathcal{S}|H + \frac{4|\mathcal{S}|H}{\epsilon - \epsilon^0}2H$, 

\begin{equation}
\begin{aligned}
    \left(J^\pi_i({r^{\mathrm{opt}}}_{m,h}(s,a), \hat{P}_i^k) + J^\pi_j({r^{\mathrm{opt}}}_{m,h}(s,a), \hat{P}^k_j)\right) - \left(  J^{\pi^*}_i(r_m, P_i) + J^{\pi^*}_j(r_m, P_j)\right) \geq \\ \frac{4|\mathcal{S}|H}{\epsilon - \epsilon^0}(J^{\pi^*}_i(\beta_m^k, \hat{P}^k_i) + J^{\pi^*}_j(\beta_m^k, \hat{P}^k_j))2H.
\end{aligned}
\end{equation}

Since $ J^{\pi^*}_i(r_m, P_i) + J^{\pi^*}_j(r_m, P_j) \leq 2H $, the inequality \eqref{A2 ineq to show} is satisfied. Now we've shown the difference in cost is less or equal to 0 for any pair of groups i, j, which is $J^{\pi^k}_i({r^{\mathrm{opt}}}_m, \hat{P}^k_i) + J^{\pi^k}_j({r^{\mathrm{opt}}}_m, \hat{P}^k_j) \leq J^{\pi^*}_i(r_m, P_i) + J^{\pi^*}_j(r_m, P_j)$ 

Using the above result for consecutive pairs of subgroups \(\{(1,2), (2,3), \ldots, (|\mathcal{Z}|-1, |\mathcal{Z}|), (|\mathcal{Z}|,1)\}\), and adding them together we get

\begin{align}
2 \sum_{z=1}^{|\mathcal{Z}|} J_z^{\pi^k} ({r^{\mathrm{opt}}}_m^k, \hat{P_z}^k) &\geq 2 \sum_{z=1}^{|\mathcal{Z}|} J_z^{\pi^*} (r_m, P_z),
\end{align}
which is
\begin{align}
\sum_{z \in \mathcal{Z}} \left( J_z^{\pi^*} (r_m, P_z) - J_z^{\pi^k} ({r^{\mathrm{opt}}}_m^k, \hat{P}_z^k) \right) &\leq 0
\end{align}

In our setting, we iterate through every group \(z\) from \(\mathcal{Z}\), therefore we have:
\begin{equation}
\begin{aligned}
    \sum_{k=1}^{K} \mathbbm{1}(|\Pi^k| > 1)\bigl(J^{\pi^*}(r_m, P) &- J^{\pi^k}({r^{\mathrm{opt}}}^k, \hat{P}^k)\bigr) \\
    &= \sum_{k=1}^{K} \mathbbm{1}(|\Pi^k| > 1)\sum_{z \in \mathcal{Z}}
        \bigl(J_z^{\pi^*}(r_m, P_z) - J_z^{\pi^k}({r^{\mathrm{opt}}}_m^k, \hat{P}_z^k)\bigr) \\
    &= \sum_{k=1}^{K} \mathbbm{1}(|\Pi^k| > 1)\sum_{z \in \mathcal{Z}}
        \bigl(J_z^{\pi^*}(r_m, P_z) - J_z^{\pi^k}({r^{\mathrm{opt}}}_m^k, \hat{P}_z^k)\bigr) \\
    &\le 0.
\end{aligned}
\end{equation}

\textbf{Lemma A.3} \textit{On good event $\mathcal{E}$,} 
\begin{equation}
\begin{aligned}
    \sum_{k=1}^{K} \mathbbm{1}(|\Pi^k| > 1) (J(\pi^k, \mu, \sum_m {r^{\mathrm{opt}}}_m^k, &\hat{P}^k) - J(\pi^k, \mu, \sum_m r_m, P)) \\ &= \tilde{\mathcal{O}}\left( \frac{|\mathcal{Z}|H^3}{(\epsilon - \epsilon^0)}\sqrt{|\mathcal{S}|^3|\mathcal{A}|K} + \frac{|\mathcal{Z}|H^5|\mathcal{S}|^3|\mathcal{A}|}{(\epsilon - \epsilon^0)}\right)
\end{aligned}
\end{equation}
\textit{Proof.} Since we build the optimistic reward with bonus, we have $|{r^{\mathrm{opt}}}_{m,h} - r_{m,h}| \leq \alpha_l \beta_h^k$. By applying Lemma B.1, 

\begin{equation}
\begin{aligned}
    &\sum_{k=1}^{K} \mathbbm{1}(|\Pi^k| > 1) (J^{\pi^k}({r^{\mathrm{opt}}}_m, \hat{P}^k) - J^{\pi^k}( r_m, P)) \\
    &\leq \sum_{k=1}^{K} \sum_{z \in \mathcal{Z}} J_z^{\pi^k}( {r^{\mathrm{opt}}}_m^k, \hat{P}_z^k) - J_z^{\pi^k}(r_m, P_z) \\
    &= \tilde{\mathcal{O}}\left(|\mathcal{Z}|(\alpha_l + \sqrt{2|\mathcal{S}|}H)(H\sqrt{|\mathcal{S}||\mathcal{A}|K}) + \alpha_l |\mathcal{Z}|H^3|\mathcal{S}|^2|\mathcal{A}|\right) \\
    &= \tilde{\mathcal{O}}\left( \frac{|\mathcal{Z}|H^3}{(\epsilon - \epsilon^0)}\sqrt{|\mathcal{S}|^3|\mathcal{A}|K} + \frac{|\mathcal{Z}|H^5|\mathcal{S}|^3|\mathcal{A}|}{(\epsilon - \epsilon^0)}\right)
\end{aligned}
\end{equation}

Combining the results for term (I), term(II) and term(III), we have

\begin{align}
    \text{Reg}(K; r_m) &= \sum_k [J(\pi_i^*, \mu, P,  r_m) - J(\pi^k, \mu, P,  r_m)] \\ &=\tilde{\mathcal{O}}\left(\frac{|\mathcal{Z}|H^3}{(\epsilon - \epsilon^0)}\sqrt{|\mathcal{S}|^3|\mathcal{A}|K} + \frac{|\mathcal{Z}|^2MH^5|\mathcal{S}|^3|\mathcal{A}|}{\min\{(\epsilon - \epsilon^0),(\epsilon - \epsilon^0)^2\}}\right)
\end{align}

\newpage

\section{Lemmas for Pessimistic and Optimistic MDP Estimates}
\begin{lemma}
\label{lemma:optimistic return}
\text{(Lemma C.2 of \cite{satija2023group})}\textit{On good event $\mathcal{E}$, for any policy $\pi$ and group $z \in \mathcal{Z}$, using the optimistic reward leads to a higher return compared to the true return.}

\begin{align}
    J_z^\pi(r_m, P_z) \leq J_z^\pi(\bar{r}_m, \hat{P}_z^k), \forall m \in [M].
\end{align}

\textit{Proof.} For any $k, h, s, a$, by the definition of optimistic reward from Equation (\ref{eq:optimistic reward}), we have
\begin{align}
    \bar{r}_{m,h}(s, a) - r_{m,h}(s, a) \geq |\mathcal{S}|H \beta_{m,h}^k(s,a)
\end{align}

Additionally, by Holder's inequality
\begin{align}
    \sum_{s'} (\hat{P}^k_{z,h} - P_{z,h})(s'|s, a)V^{\pi_z}_{h+1}(s'; r_m, P_z) \geq -H \sum_{s'}\beta_{m,h}^k(s, a) = - H|\mathcal{S}|\beta_{m,h}^k(s,a),
\end{align}

Using the value difference lemma (\hyperref[lemma:value difference lemma]{Lemma 2}), for any policy $\pi$ and any task $m$, we have:
\begin{equation}
\begin{aligned}
    &J_z^\pi(\bar{r}^k_m, \hat{P}^k_z) - J_z^\pi(r_m, P_z) \\
    &= \mathbbm{E}\left[\sum^H_{h=1}( \bar{r}_{m,h}(s_h, a_h) -  r_{m,h}(s_h, a_h)) + \sum_{s'} (\hat{P}^k_{z,h} - P_{z,h})(s'|s, a)V^{\pi_z}_{h+1}(s'; r_m, P_z) \Big| \mathcal{F}_{k-1}\right] \\
    &\geq \mathbbm{E}\left[\sum_{h=1}^H |\mathcal{S}|H\beta^k_{m,h}(s, a) - H|\mathcal{S}|\beta^k_{m,h}(s,a)\Big| \mathcal{F}_{k-1}\right] \\
    &\geq 0.
\end{aligned}
\end{equation}
Therefore, we have
\begin{align}
    J_z^\pi(r_m, P_z) \leq J_z^\pi(\bar{r}_m, \hat{P}_z^k), \forall m \in [M].
\end{align}
\end{lemma}

\begin{lemma}
\label{lemma:pessimistic return}
\text{(Lemma C.3 of \cite{satija2023group})}\textit{On good event $\mathcal{E}$, for any policy $\pi$ and group $z \in \mathcal{Z}$, using the optimistic reward leads to a higher return compared to the true return.}

\begin{align}
    J_z^\pi(\underbar{r}_m, P_z) \leq J_z^\pi(r_m, \hat{P}_z^k), \forall m \in [M].
\end{align}

\textit{Proof.} For any $k, h, s, a$, by the definition of optimistic reward from Equation (\ref{eq:optimistic reward}), we have
\begin{align}
    \underbar{r}_{m,h}(s, a) - r_{m,h}(s, a) \leq - |\mathcal{S}|H \beta_{m,h}^k(s,a)
\end{align}

Additionally, by Holder's inequality
\begin{align}
    \sum_{s'} (\hat{P}^k_{z,h} - P_{z,h})(s'|s, a)V^{\pi_z}_{h+1}(s'; r_m, P_z) \leq H \sum_{s'}\beta_h^k(s, a) =  H|\mathcal{S}|\beta_{m,h}^k(s,a),
\end{align}

Using the value difference lemma (\hyperref[lemma:value difference lemma]{Lemma 2}), for any policy $\pi$ and any task $m$, we have:
\begin{equation}
\begin{aligned}
    &J_z^\pi(\underbar{r}_m, P_z) - J_z^\pi(r_m, \hat{P}_z^k) \\
    &= \mathbbm{E}\left[\sum^H_{h=1}( \underbar{r}_{m,h}(s_h, a_h) -  r_{m,h}(s_h, a_h)) + \sum_{s'} (\hat{P}^k_{z,h} - P_{z,h})(s'|s, a)V^{\pi_z}_{h+1}(s'; r_m, P_z) \Big| \mathcal{F}_{k-1}\right] \\
    &\leq \mathbbm{E}\left[\sum_{h=1}^H -|S|H\beta^k_{m,h}(s, a) + H|S|\beta^k_{m,h}(s,a)\Big| \mathcal{F}_{k-1}\right] \\
    &\leq 0.
\end{aligned}
\end{equation}
Therefore, we have
\begin{align}
    J_z^\pi(\underbar{r}_m, P_z) \leq J_z^\pi(r_m, \hat{P}_z^k), \forall m \in [M].
\end{align}
\end{lemma}

\begin{lemma}
\label{lemma:optimistic return diff}
\text{(Lemma C.4 of \cite{satija2023group})}\textit{On good event $\mathcal{E}$, for any policy $\pi$ and group $z \in \mathcal{Z}$, the difference in return using the optimistic reward function and the true reward function can be bounded in terms of $\beta^k$:}

\begin{align}
    J_z^\pi(\bar{r}_m, \hat{P}_z^k) - J_z^\pi(r_m, P_z) \leq (|\mathcal{S}|H)J_z^\pi(\beta_m^k, \hat{P}^k), \forall m \in [M].
\end{align}

\textit{Proof.} For any $k, h, s, a$, by the definition of optimistic reward from Equation(\ref{eq:optimistic reward}), we have
\begin{align}
    \bar{r}_{m,h}(s, a) - r_{m,h}(s, a) = |\mathcal{S}|H \beta_{m,h}^k(s,a)
\end{align}

Additionally, by Holder's inequality
\begin{align}
    \sum_{s'} (\hat{P}_{z,h} - P_{z,h})(s'|s, a)V^{\pi_z}_{h+1}(s'; r_m, P_z) \leq H \sum_{s'}\beta_{m,h}^k(s, a) =  H|\mathcal{S}|\beta_{m,h}^k(s,a),
\end{align}

Using the value difference lemma (\hyperref[lemma:value difference lemma]{Lemma 2}), for any policy $\pi$ and any task $m$, we have:
\begin{equation}
\begin{aligned}
    &J_z^\pi(\bar{r}_m,\hat{P}_z) - J_z^\pi(r_m, P_z) \\
    &= \mathbbm{E}\left[\sum^H_{h=1}( \bar{r}_{m,h}(s_h, a_h) -  r_{m,h}(s_h, a_h)) + \sum_{s'} (\hat{P}^k_{z,h} - P_{z,h})(s'|s, a)V^{\pi_z}_{h+1}(s'; r_m, P_z) \Big| \mathcal{F}_{k-1}\right] \\
    &\leq \mathbbm{E}\left[\sum_{h=1}^H |\mathcal{S}|H\beta^k_{m,h}(s, a) + H|\mathcal{S}|\beta^k_{m,h}(s,a)\Big| \mathcal{F}_{k-1}\right] \\
    &\leq 2(|\mathcal{S}|H)J_z^\pi(\beta_m^k, \hat{P}^k).
\end{aligned}
\end{equation}
Therefore, we have
\begin{align}
    J_z^\pi(\bar{r}_m, \hat{P}_z^k) - J_z^\pi(r_m, P_z) \leq  2(|\mathcal{S}|H)J_z^\pi(\beta_m^k, \hat{P}^k), \forall m \in [M].
\end{align}
\end{lemma}

\begin{lemma}
\label{lemma:pessimistic return diff}
\text{(Lemma C.5 of \cite{satija2023group})}\textit{On good event $\mathcal{E}$, for any policy $\pi$ and group $z \in \mathcal{Z}$, the difference in return using the true reward function and the pessimistic reward function can be bounded in terms of $\beta_m^k$:}

\begin{align}
    J_z^\pi(r_m, \hat{P}_z^k) - J_z^\pi(\underbar{r}_m, P_z) \leq (|\mathcal{S}|H)J_z^\pi(\beta_m^k, \hat{P}^k), \forall m \in [M].
\end{align}

\textit{Proof.} For any $k, h, s, a$, by the definition of optimistic reward from Equation(\ref{eq:optimistic reward}), we have
\begin{align}
    r_{m,h}(s, a) - \underbar{r}_{m,h}(s, a) = |\mathcal{S}|H \beta_{m,h}^k(s,a)
\end{align}

Additionally, by Holder's inequality
\begin{align}
    \sum_{s'} (\hat{P}^k_{z,h} - P_{z,h})(s'|s, a)V^{\pi_z}_{h+1}(s'; r_m, P_z) \leq H \sum_{s'}\beta_{m,h}^k(s, a) =  H|\mathcal{S}|\beta_{m,h}^k(s,a),
\end{align}

Using the value difference lemma (\hyperref[lemma:value difference lemma]{Lemma 2}), for any policy $\pi$ and any task $m$, we have:
\begin{equation}
\begin{aligned}
    &J_z^\pi(r_m, P_z) - J_z^\pi(\underbar{r}_m, \hat{P}^k_z) \\
    &= \mathbbm{E}\left[\sum^H_{h=1}( r_{m,h}(s_h, a_h) -  \underbar{r}_{m,h}(s_h, a_h)) + \sum_{s'} (\hat{P}^k_{z,h} - P_{z,h})(s'|s, a)V^{\pi_z}_{h+1}(s'; r_m, P_z) \Big| \mathcal{F}_{k-1}\right] \\
    &\leq \mathbbm{E}\left[\sum_{h=1}^H |\mathcal{S}|H\beta^k_{m,h}(s, a) + H|\mathcal{S}|\beta^k_{m,h}(s,a)\Big| \mathcal{F}_{k-1}\right] \\
    &\leq 2(|\mathcal{S}|H)J_z^\pi(\beta_m^k, \hat{P}_z^k).
\end{aligned}
\end{equation}
Therefore, we have
\begin{align}
    J_z^\pi(r_m, P_z) - J_z^\pi(\underbar{r}_m, \hat{P}_z^k) \leq  2(|\mathcal{S}|H)J_z^\pi(\beta_m^k, \hat{P}_z^k), \forall m \in [M].
\end{align}
\end{lemma}

\section{Supporting Lemmas}

\begin{lemma}
\label{Lemma Hoeffding's inequality}
(Hoeffding's inequality). For independent zero-mean 1/2-sub-gaussian random variables $X_1, X_2, ..., X_n$,

\begin{align}
    Pr(\frac{1}{n}\sum^N_{n=1}X_n \geq \epsilon) \leq \exp(-n\epsilon^2).
\end{align}
\end{lemma}

\begin{lemma}
    \label{lemma:value difference lemma}
    (Value difference lemma, \cite{NIPS2017_PAC&Regret}, Lemma E.15).
    $$
    \begin{aligned}
    & V_1^\pi\left(\mu ;  r'_m, P^{\prime}\right)-V_1^\pi(\mu ; r_m, P) \\
    = & \mathbb{E}_{\mu, P, \pi}\left[\sum_{h=1}^H\left( r'_m\left(S_h, A_h\right)- r_m \left(S_h, A_h\right)+\sum_{s^{\prime}}\left(P_h^{\prime}-P_h\right)\left(s^{\prime} \mid S_h, A_h\right) V_{h+1}^\pi\left(s^{\prime} ;  r'_m, P'\right)\right) \mid \mathcal{F}_{k-1}\right] \\
    = & \mathbb{E}_{\mu, P', \pi}\left[\sum_{h=1}^H\left( r'_m\left(S_h, A_h\right)- r_m \left(S_h, A_h\right)+\sum_{s^{\prime}}\left(P_h'-P_h\right)\left(s' \mid S_h, A_h\right) V_{h+1}^\pi\left(s' ;  r_m, P\right)\right) \mid \mathcal{F}_{k-1}\right],
    \end{aligned}
    $$
    where $r_m$ denotes the reward function of task $m$.
\end{lemma}

\begin{lemma}
\label{lemma:H3}
\textit{(Lemma H.3 of {\cite{satija2023group}}, Lemma D.4 of \cite{liu2021})}. Let $\mathcal{G}_{1:K}$ be a sequence of events such that $\mathcal{G}_k \in \mathcal{F}_{k-1}$ for each $k \in [K]$. Suppose $|\tilde{g}^k - g| \leq \alpha \beta^k$, $\alpha \geq 1$. On good event $\mathcal{E}$, for any $K' \leq K$,
\begin{align}
    \sum_{k=1}^{K'} \mathbbm{1}(\mathcal{G}_k) \left| J_z^{\pi^k}(\tilde{g}^k, \hat{P}_z^k) - J_z^{\pi^k}(g, P_z) \right| \leq (3\alpha + 3\sqrt{2H} \sqrt{|\tilde{\mathcal{S}}|}) H \sqrt{|\tilde{\mathcal{S}}||\mathcal{A}| K'_\mathcal{G}} + \tilde{O}(\alpha H^3 |\tilde{\mathcal{S}}|^2 |\mathcal{A}|),
\end{align}
where $K'_\mathcal{G} = \sum_{k=1}^{K'} \mathbbm{1}(\mathcal{G}_k)$.
\end{lemma}

\begin{lemma}
\label{Lemma H.4}
\textit{(Lemma H.4 of {\cite{satija2023group}}, Lemma D.5 of \cite{liu2021})}. Given a sequence of events $\mathcal{G}_{1:K}$ that $\mathcal{G}_k \in \{\mathcal{F}\}_{k-1}$ for each $k \in [K]$. With probability at least $1 - \delta$, for any $K' \leq K$,

\begin{align}
    &\sum^{K'}_{k=1} \sum^H_{h=1} \sum_{z, s, a} \frac{\mathbbm{1}(\mathcal{G}_k) d^{\pi^k}_{z,h}(s, a)}{\max(N^k_{z,h}(s, a), 1)} \leq 4H|\mathcal{Z}||S||A| + 2H|\mathcal{Z}||S||A|\ln K'_\mathcal{G} + 4\ln \frac{2HK}{\delta}, 
\end{align}
\begin{equation}
\begin{aligned}
    &\sum^{K'}_{k=1} \sum^H_{h=1} \sum_{z, s, a} \frac{\mathbbm{1}(\mathcal{G}_k) d^{\pi^k}_{z,h}(s, a)}{\sqrt{\max(N^k_{z,h}(s, a), 1)}} \leq  6H|\mathcal{Z}||S||A| + 2H\sqrt{|\mathcal{Z}||S||A|\ln K'_\mathcal{G}} \\
    & \quad \quad \quad \quad \quad \quad \quad \quad \quad \quad \quad \quad \quad \quad \quad + 2H|\mathcal{Z}||S||A|\ln K'_\mathcal{G} +   5\ln \frac{2HK}{\delta}, \left. \right\},
\end{aligned}
\end{equation}

where $N^k_{z, h}(s, a)$ denotes the number of times the state-action tuple $(s, a)$ was observed at time step $h$ so far in episodes $[1, \ldots, k-1]$, $K'_g \dot{=} \sum_{k'=1}^{K'} \mathbbm{1}(\mathcal{G}_k)$, and $d_h^{\pi^k}(s, a)$ is the occupancy measure of policy $\pi^k$ such that $d^{\pi^k}_{z, h}(s, a) = \mathbbm{E}_{\mu_z, P_z, \pi^k}[\mathbbm{1}(S_{z, h} = s, A_h = a | \mathcal{F}_{k-1})]$.
\end{lemma}

\begin{lemma}
\label{lemma:H5}
\textit{(Lemma H.5 of {\cite{satija2023group}}, Lemma D.6 of \cite{liu2021})}. Suppose $0 \leq x \leq a + b\sqrt{x}$, for some $a, b > 0$,
\begin{align}
    x \leq \frac{3}{2}a + \frac{3}{2}b^2.
\end{align}
\end{lemma}

\newpage
\section{Plots for Infinite Horizon Experiments}
\label{sec:appendix2}

In this section, we present additional experimental results and visualizations for the infinite-horizon setting. These plots provide a more direct comparison of the Multi-Task Group Fairness (MTGF) algorithm against the single-task Group Fairness RL (GFRL) baseline. As the figures illustrate, MTGF consistently exhibits reduced maximum fairness violations compared to the GFRL baseline.

\begin{figure}[htbp]
    \centering
    \begin{subfigure}[b]{0.45\textwidth}
        \includegraphics[width=\textwidth]{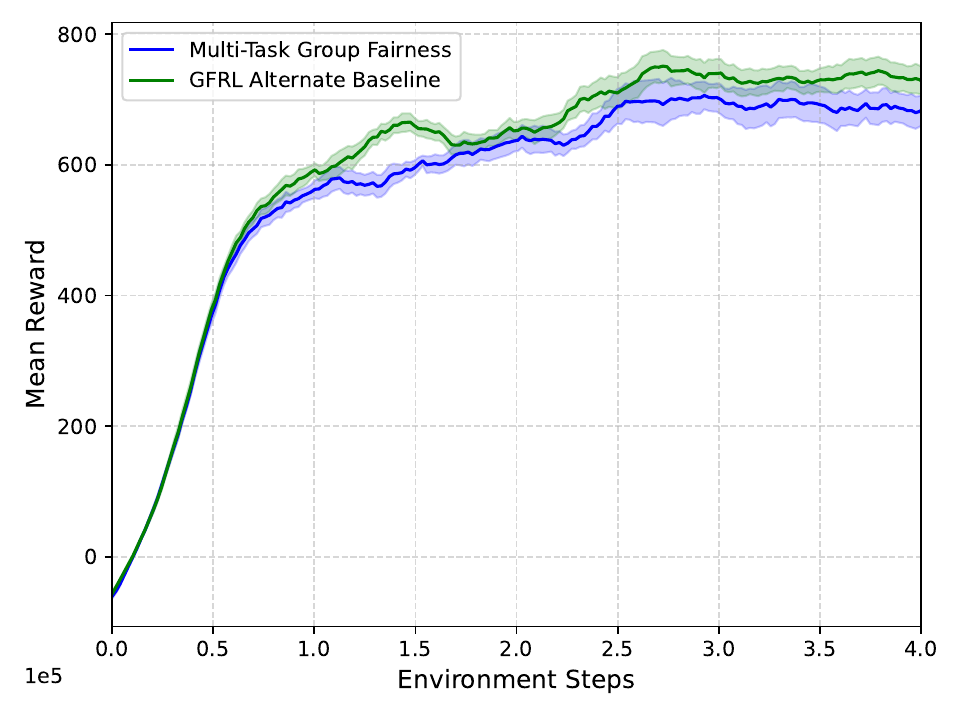}
        \caption{Ant - Forward Running}
        \label{fig:HG_orig_forward}
    \end{subfigure}
    \begin{subfigure}[b]{0.45\textwidth}
        \includegraphics[width=\textwidth]{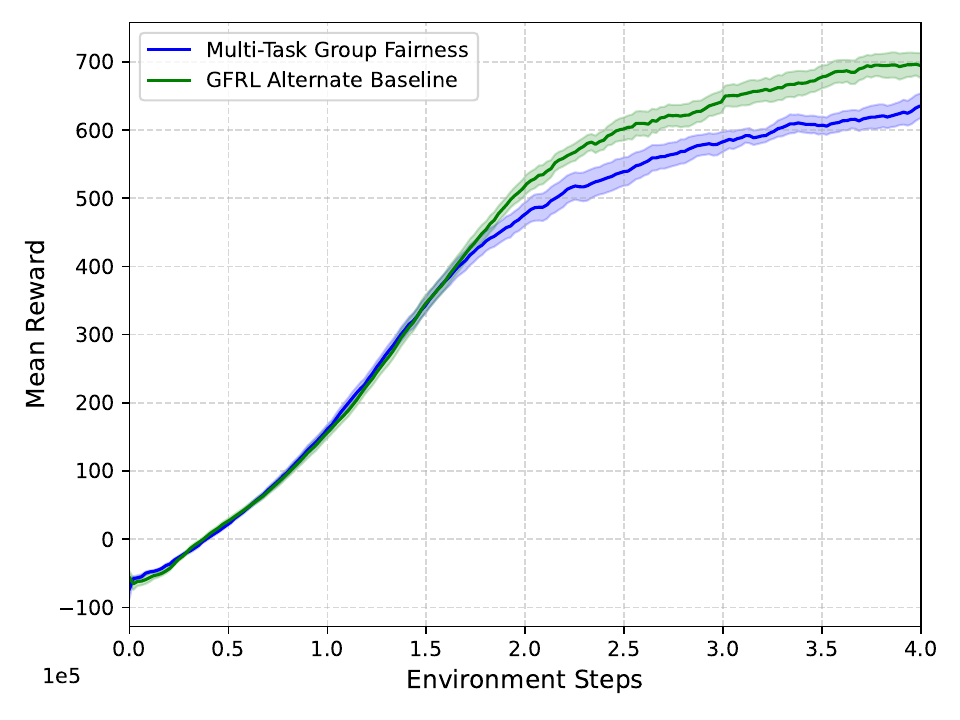}
        \caption{Ant - Backward Running}
        \label{fig:HG_orig_backward}
    \end{subfigure}
    
    \begin{subfigure}[b]{0.45\textwidth}
        \includegraphics[width=\textwidth]{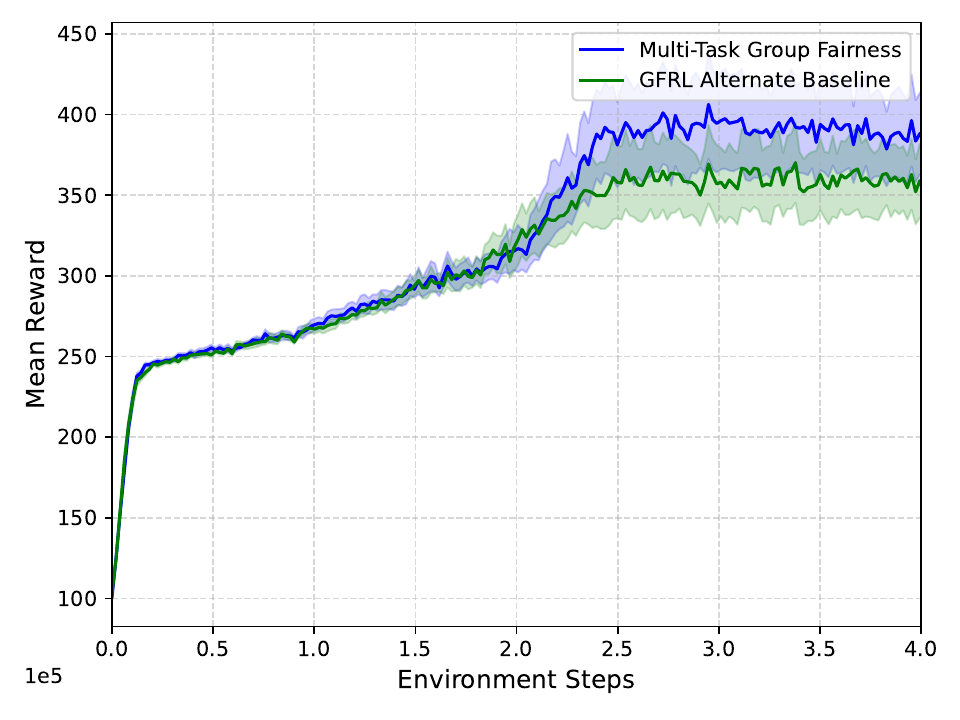}
        \caption{Humanoid - Forward Running}
        \label{fig:HG_huge_forward}
    \end{subfigure}
    \begin{subfigure}[b]{0.45\textwidth}
        \includegraphics[width=\textwidth]{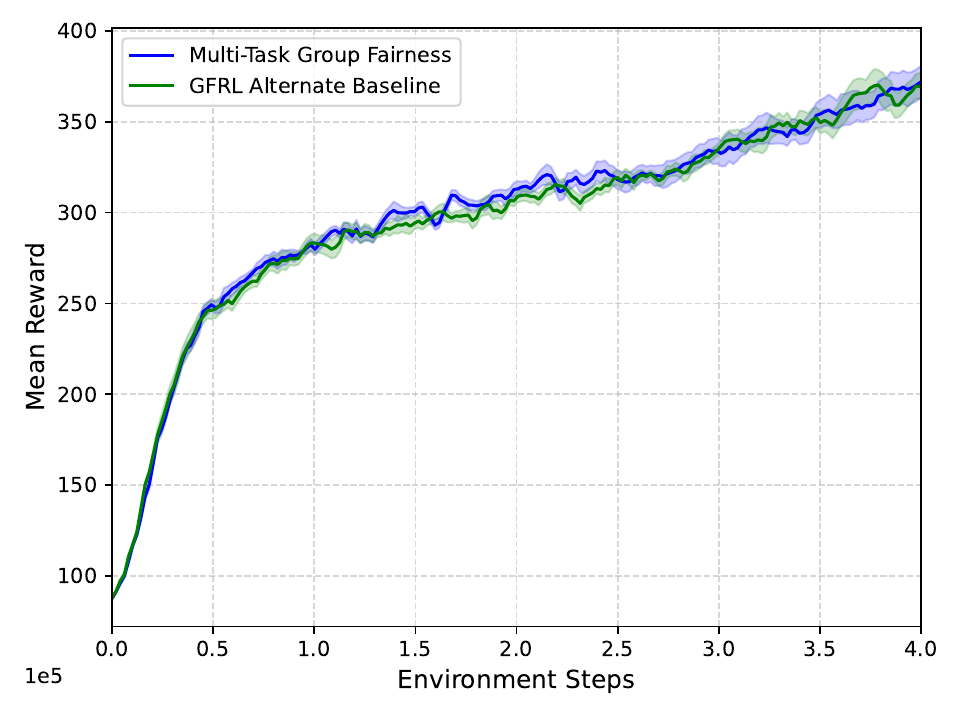}
        \caption{Humanoid - Backward Running}
        \label{fig:HG_huge_backward}
    \end{subfigure}
    
    \begin{subfigure}[b]{0.45\textwidth}
        \includegraphics[width=\textwidth]{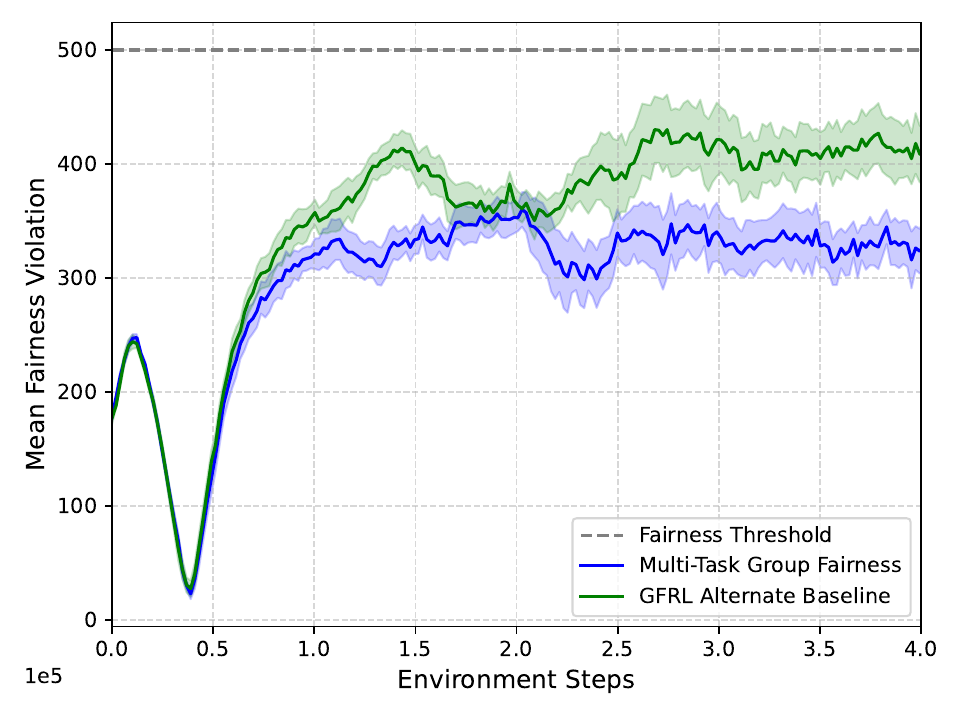}
        \caption{Forward Running Fairness Gap}
        \label{fig:HG_gap_forward}
    \end{subfigure}
    \begin{subfigure}[b]{0.45\textwidth}
        \includegraphics[width=\textwidth]{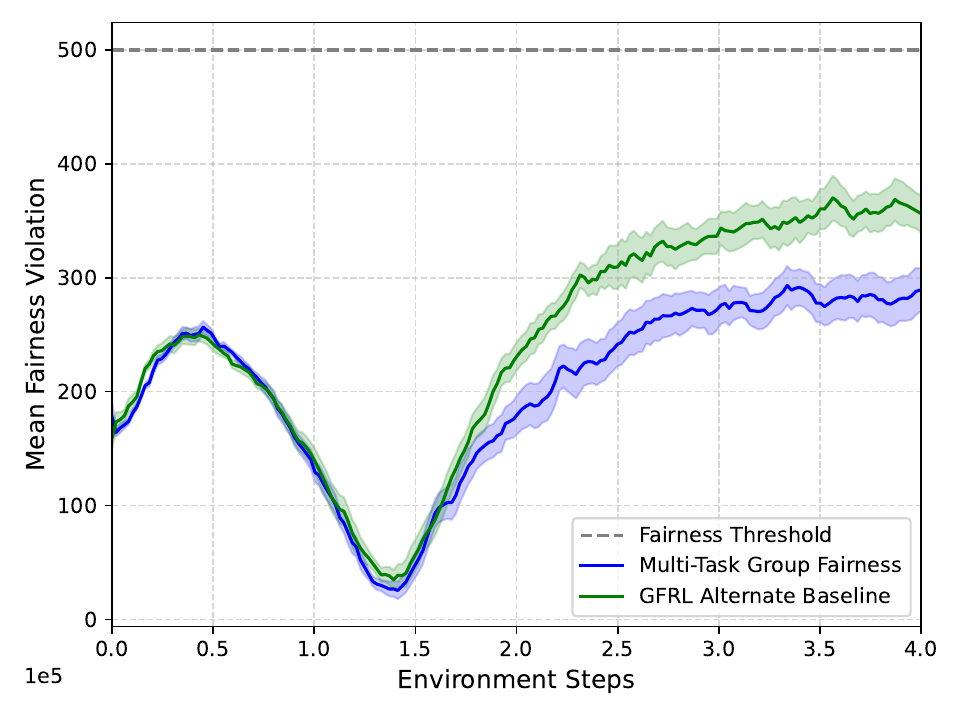}
        \caption{Backward Running Fairness Gap}
        \label{fig:HG_gap_backward}
    \end{subfigure}
    \caption{Comparison between Ant and Humanoid: Performance and Fairness Gaps}
    \label{fig:hugegravity_comparison}
\end{figure}

\begin{figure}[htbp]
    \centering
    \begin{subfigure}[b]{0.45\textwidth}
        \includegraphics[width=\textwidth]{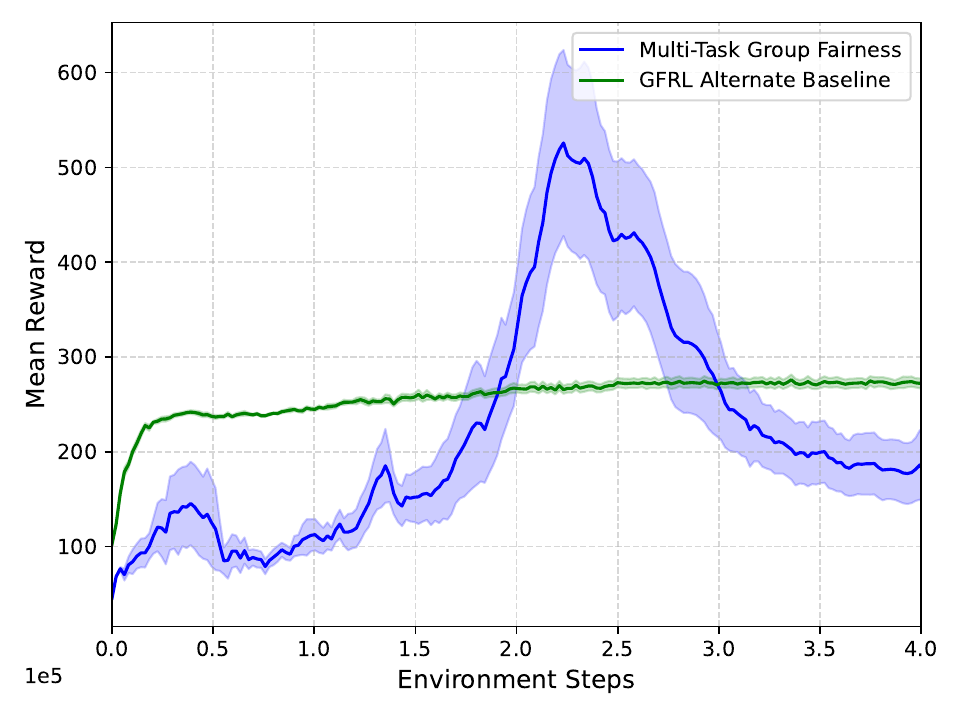}
        \caption{Hopper - Forward Running}
        \label{fig:HG_orig_forward}
    \end{subfigure}
    \begin{subfigure}[b]{0.45\textwidth}
        \includegraphics[width=\textwidth]{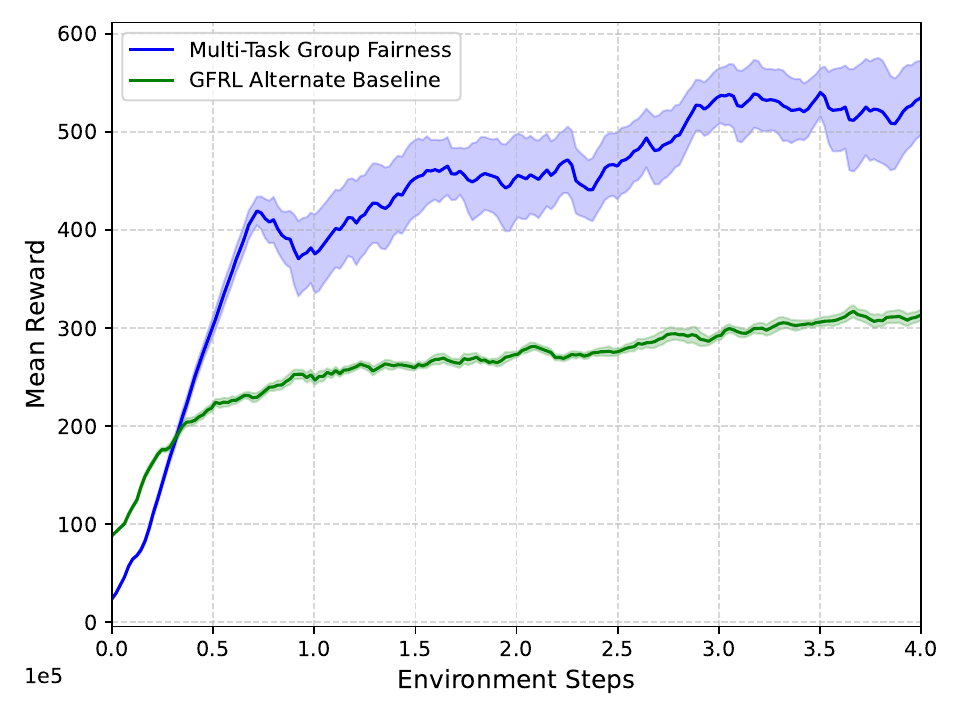}
        \caption{Hopper - Backward Running}
        \label{fig:HG_orig_backward}
    \end{subfigure}
    
    \begin{subfigure}[b]{0.45\textwidth}
        \includegraphics[width=\textwidth]{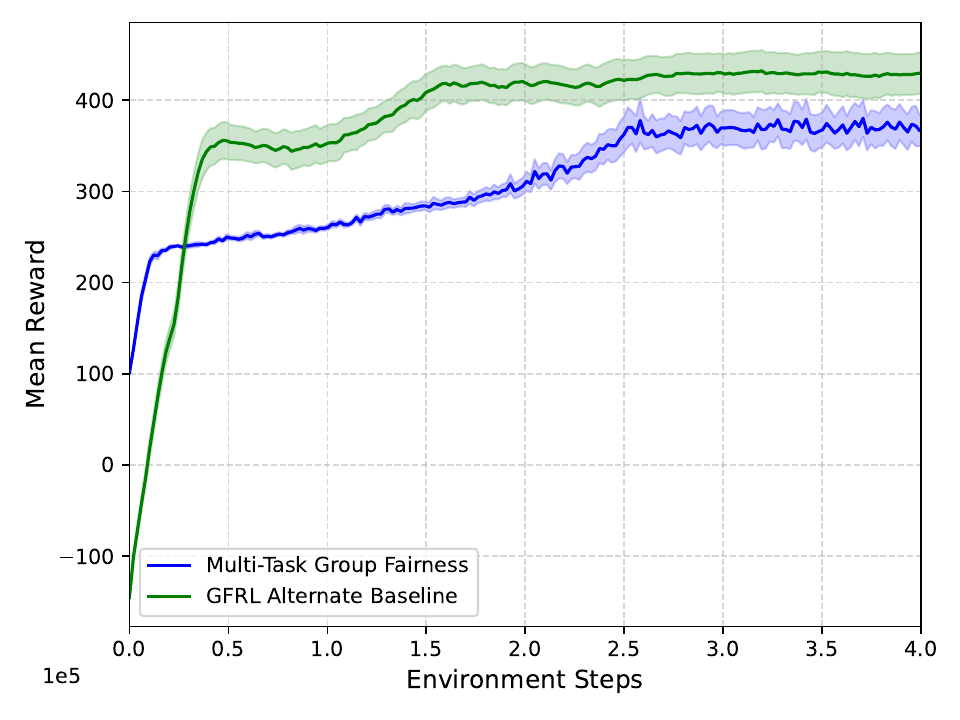}
        \caption{Humanoid - Forward Running}
        \label{fig:HG_huge_forward}
    \end{subfigure}
    \begin{subfigure}[b]{0.45\textwidth}
        \includegraphics[width=\textwidth]{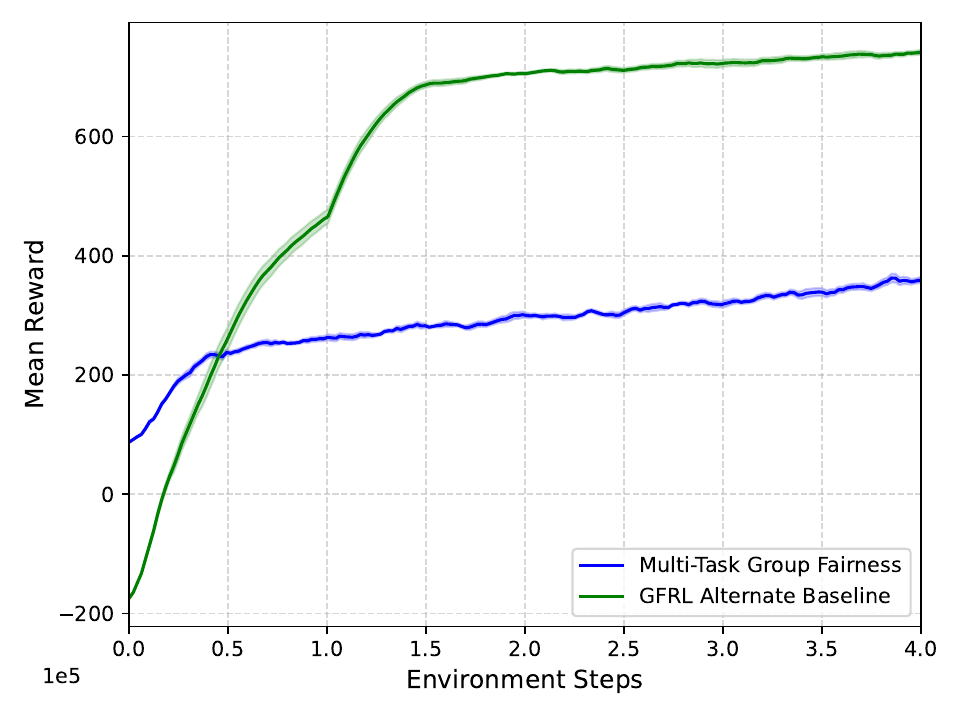}
        \caption{Humanoid - Backward Running}
        \label{fig:HG_huge_backward}
    \end{subfigure}
    
    \begin{subfigure}[b]{0.45\textwidth}
        \includegraphics[width=\textwidth]{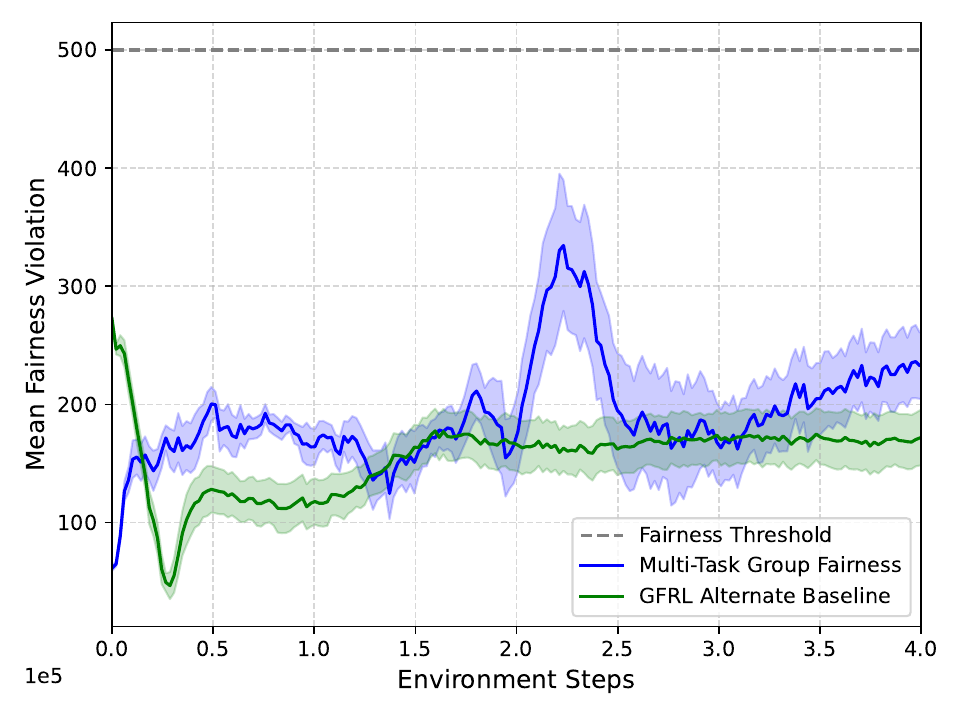}
        \caption{Forward Running Fairness Gap}
        \label{fig:HG_gap_forward}
    \end{subfigure}
    \begin{subfigure}[b]{0.45\textwidth}
        \includegraphics[width=\textwidth]{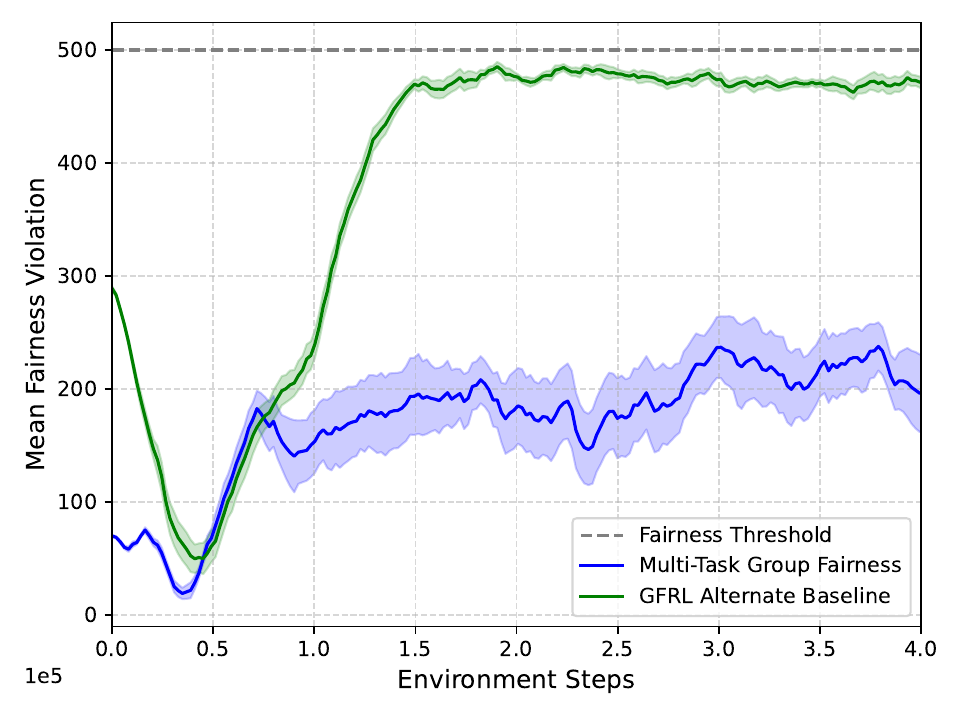}
        \caption{Backward Running Fairness Gap}
        \label{fig:HG_gap_backward}
    \end{subfigure}
    \caption{Comparison between Hopper and Humanoid: Performance and Fairness Gaps}
    \label{fig:hugegravity_comparison}
\end{figure}

\begin{figure}[htbp]
    \centering
    \begin{subfigure}[b]{0.45\textwidth}
        \includegraphics[width=\textwidth]{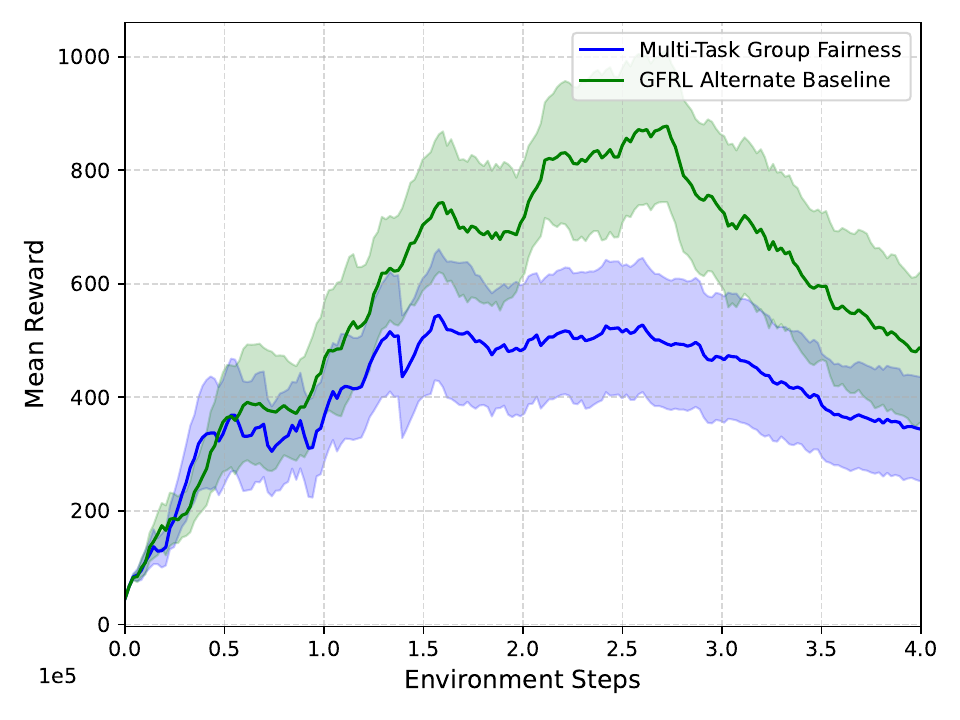}
        \caption{Hopper - Forward Running}
        \label{fig:HG_orig_forward}
    \end{subfigure}
    \begin{subfigure}[b]{0.45\textwidth}
        \includegraphics[width=\textwidth]{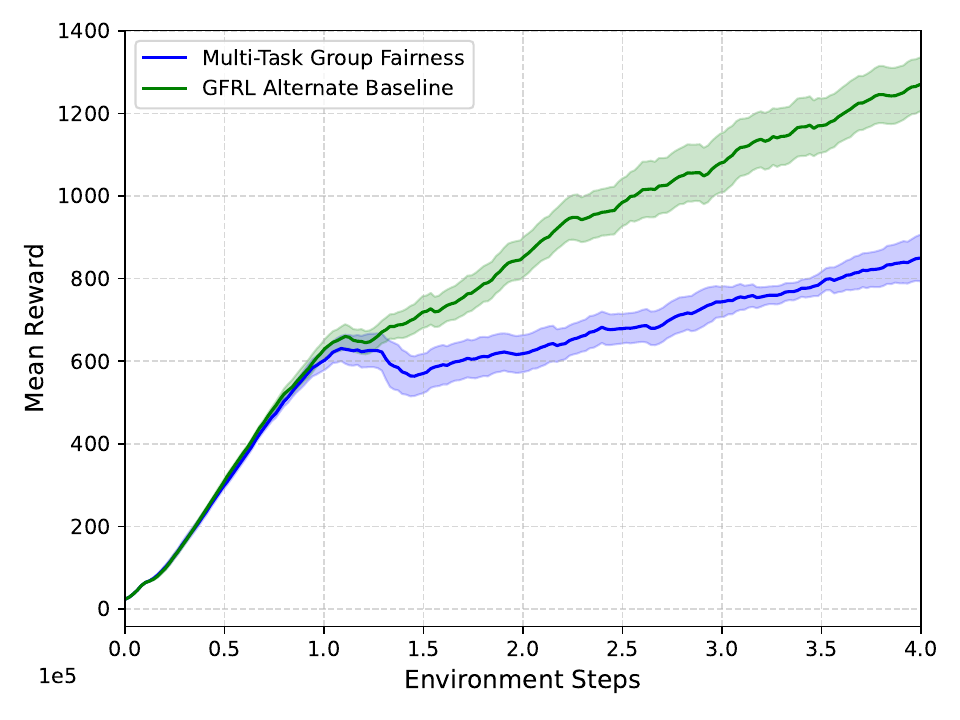}
        \caption{Hopper - Backward Running}
        \label{fig:HG_orig_backward}
    \end{subfigure}
    
    \begin{subfigure}[b]{0.45\textwidth}
        \includegraphics[width=\textwidth]{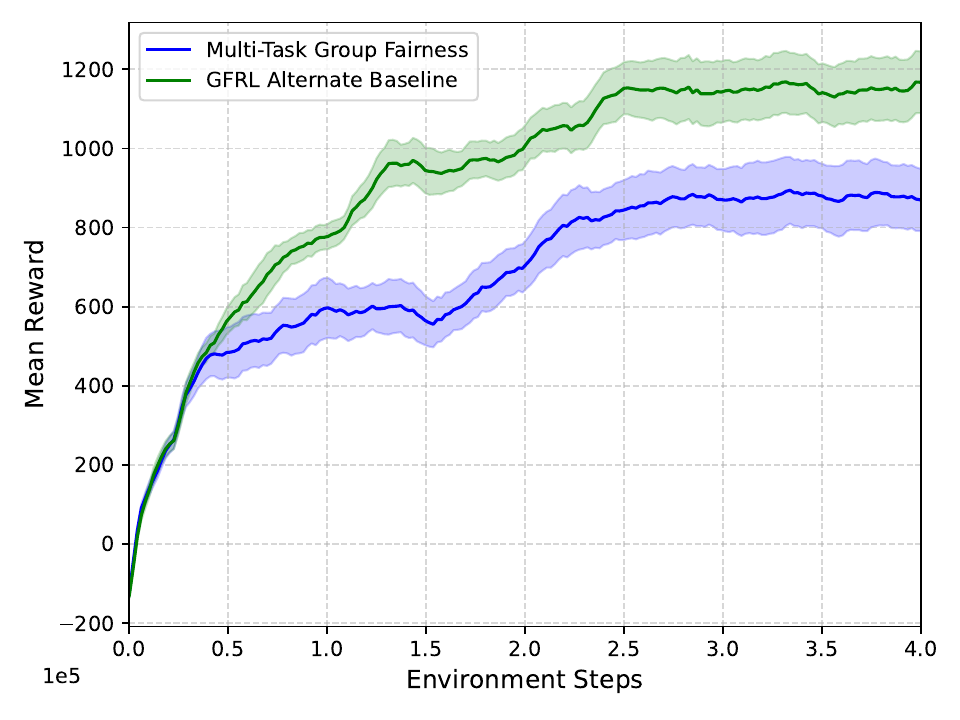}
        \caption{HugeGravity HalfCheetah - Forward Running}
        \label{fig:HG_huge_forward}
    \end{subfigure}
    \begin{subfigure}[b]{0.45\textwidth}
        \includegraphics[width=\textwidth]{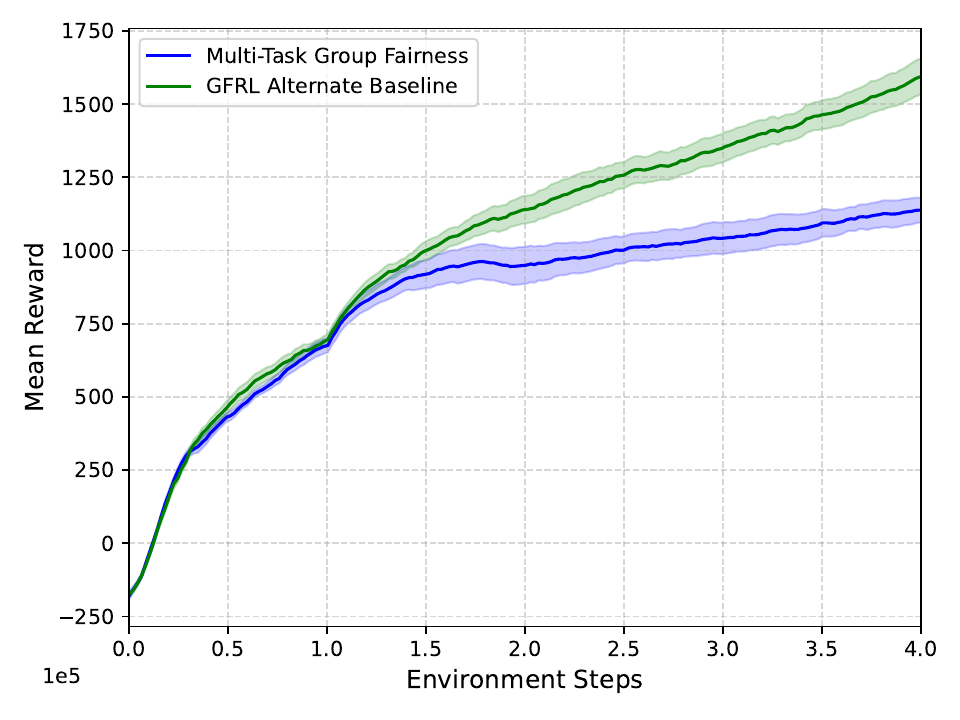}
        \caption{HugeGravity HalfCheetah-Backward Running}
        \label{fig:HG_huge_backward}
    \end{subfigure}
    
    \begin{subfigure}[b]{0.45\textwidth}
        \includegraphics[width=\textwidth]{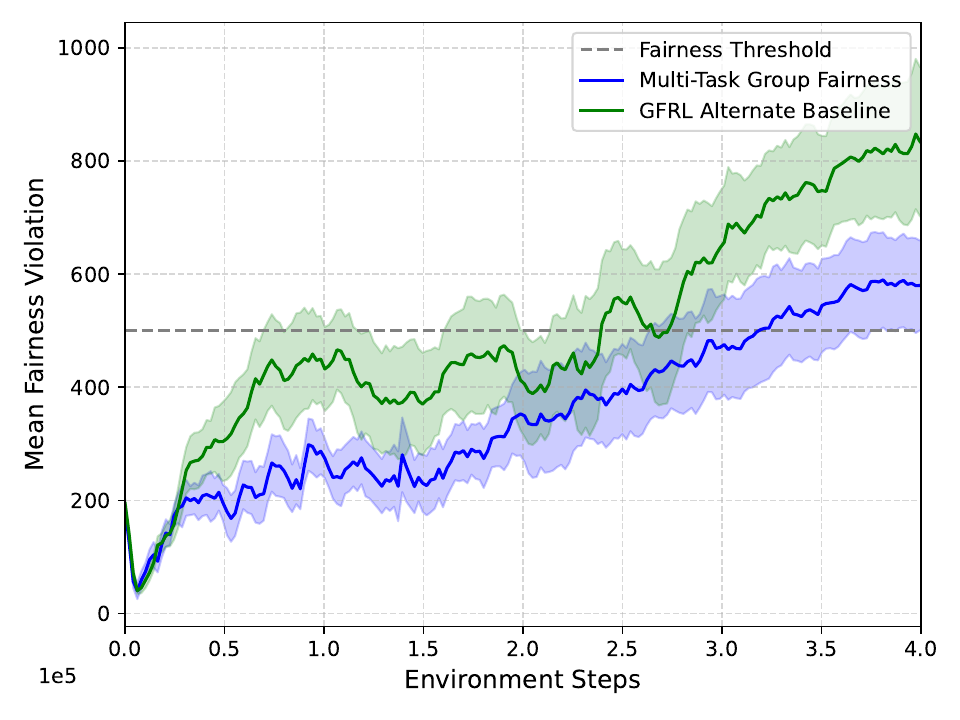}
        \caption{Forward Running Fairness Gap}
        \label{fig:HG_gap_forward}
    \end{subfigure}
    \begin{subfigure}[b]{0.45\textwidth}
        \includegraphics[width=\textwidth]{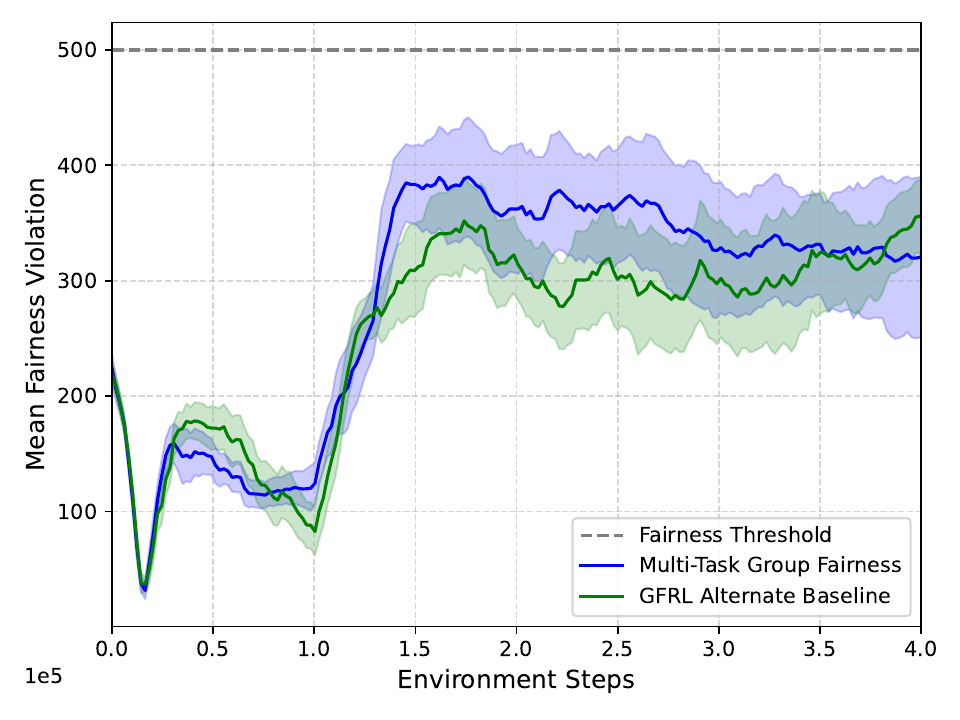}
        \caption{Backward Running Fairness Gap}
        \label{fig:HG_gap_backward}
    \end{subfigure}
    \caption{Comparison between Hopper and HugeGravity HalfCheetah: Performance and Fairness Gaps}
    \label{fig:hugegravity_comparison}
\end{figure}

\begin{figure}[htbp]
    \centering
    \begin{subfigure}[b]{0.45\textwidth}
        \includegraphics[width=\textwidth]{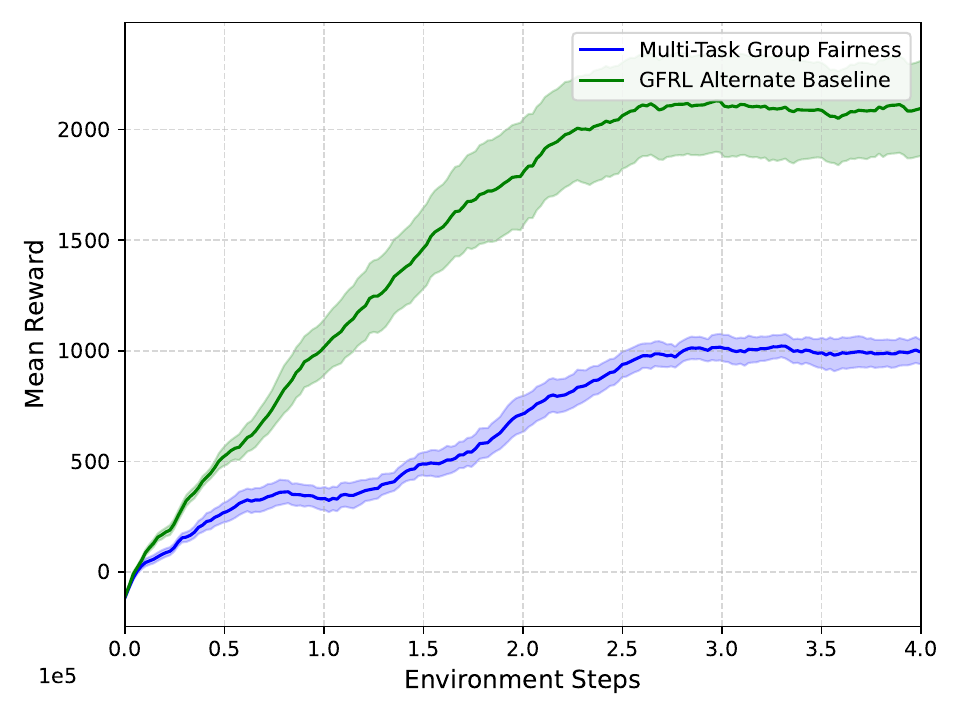}
        \caption{Original HalfCheetah - Forward Running}
        \label{fig:HG_orig_forward}
    \end{subfigure}
    \begin{subfigure}[b]{0.45\textwidth}
        \includegraphics[width=\textwidth]{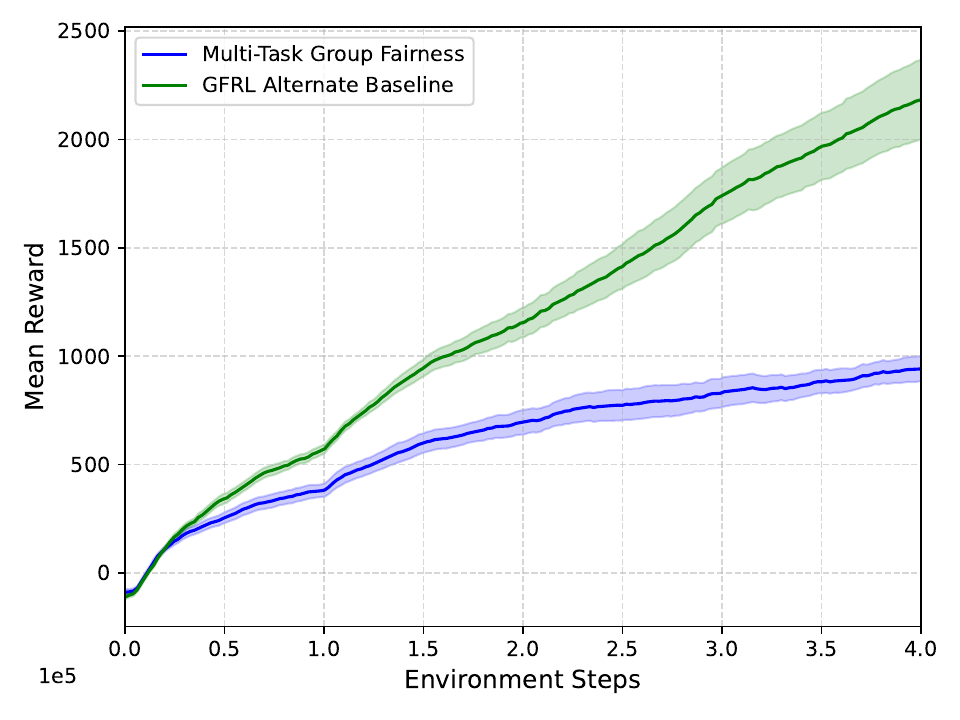}
        \caption{Original HalfCheetah - Backward Running}
        \label{fig:HG_orig_backward}
    \end{subfigure}
    
    \begin{subfigure}[b]{0.45\textwidth}
        \includegraphics[width=\textwidth]{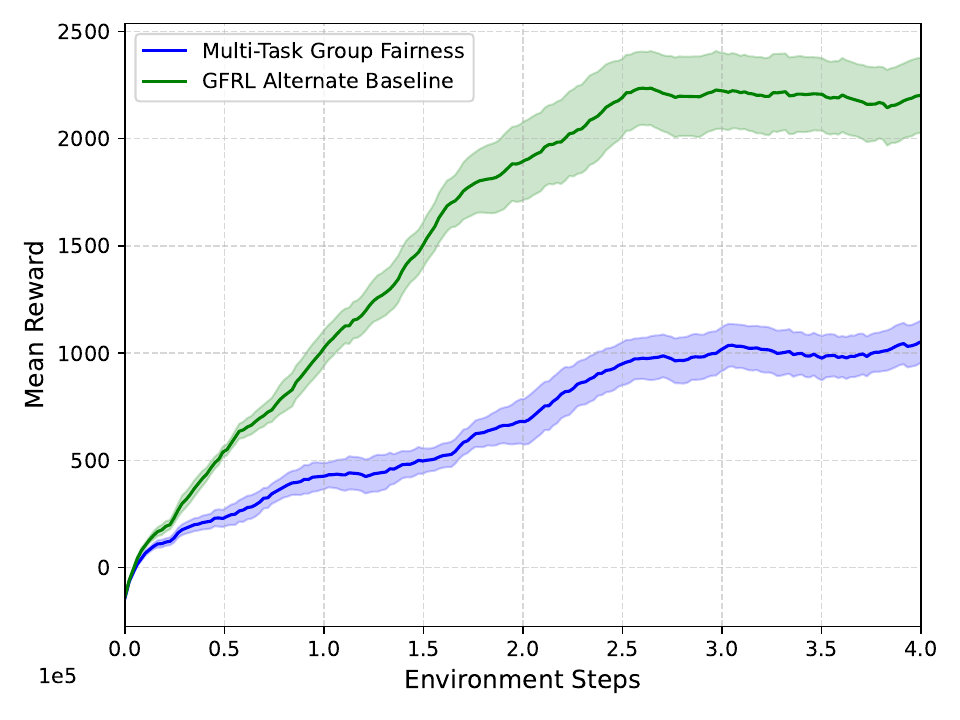}
        \caption{HugeGravity HalfCheetah - Forward Running}
        \label{fig:HG_huge_forward}
    \end{subfigure}
    \begin{subfigure}[b]{0.45\textwidth}
        \includegraphics[width=\textwidth]{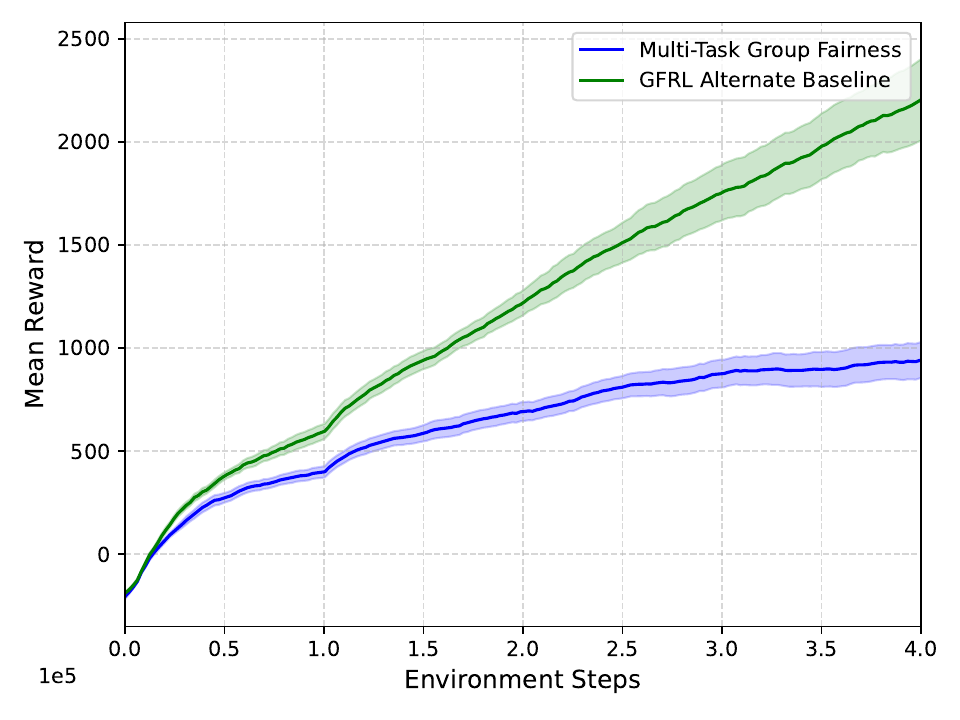}
        \caption{HugeGravity HalfCheetah-Backward Running}
        \label{fig:HG_huge_backward}
    \end{subfigure}
    
    \begin{subfigure}[b]{0.45\textwidth}
        \includegraphics[width=\textwidth]{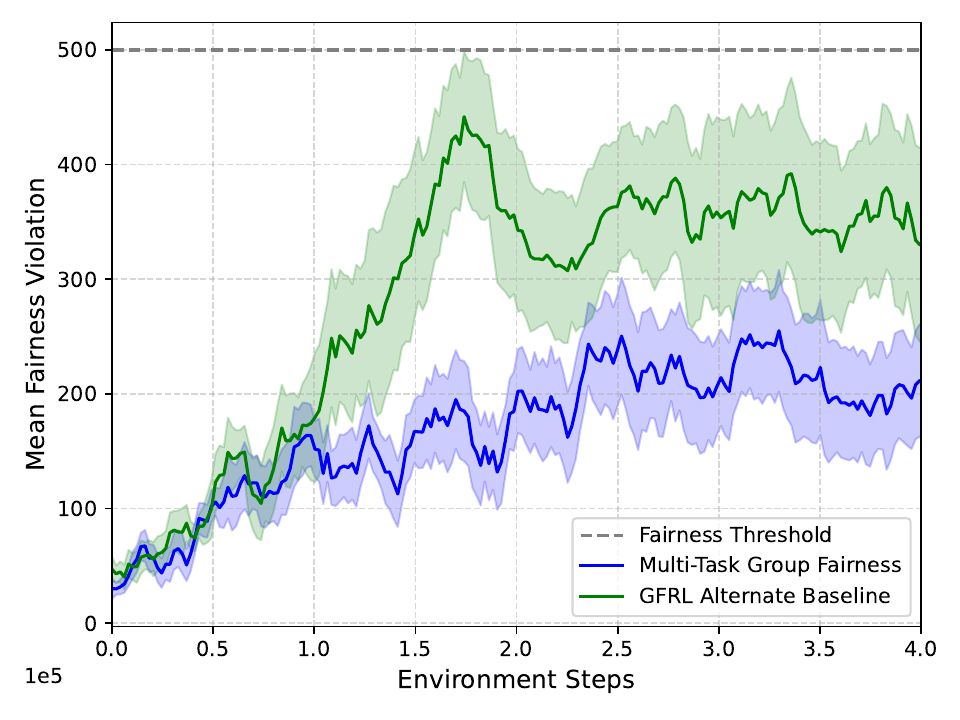}
        \caption{Forward Running Fairness Gap}
        \label{fig:HG_gap_forward}
    \end{subfigure}
    \begin{subfigure}[b]{0.45\textwidth}
        \includegraphics[width=\textwidth]{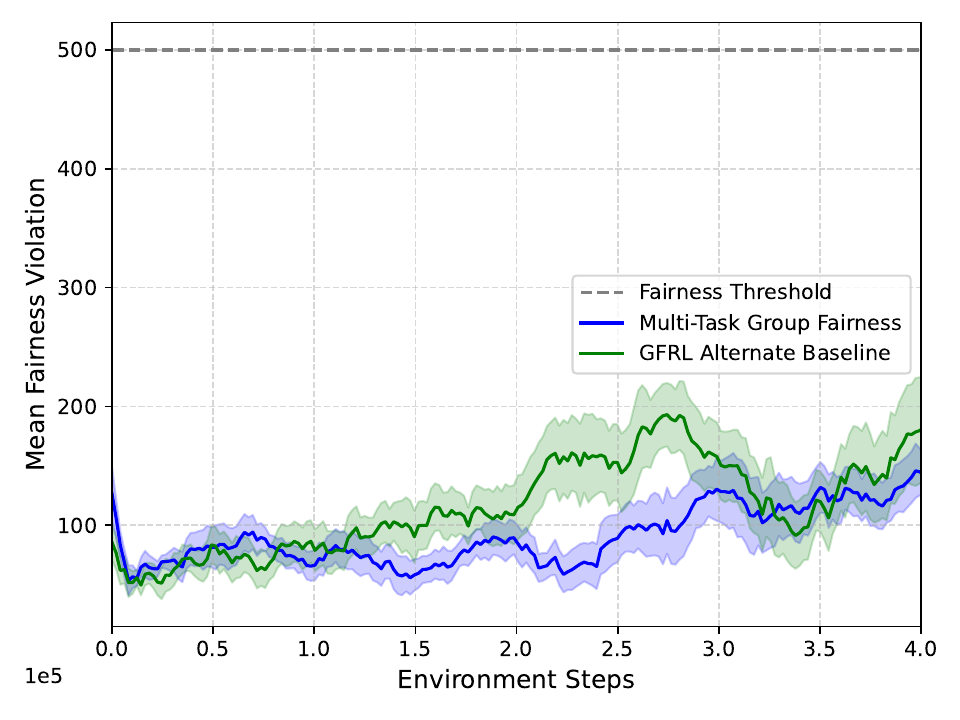}
        \caption{Backward Running Fairness Gap}
        \label{fig:HG_gap_backward}
    \end{subfigure}
    \caption{Comparison between Original HalfCheetah and HugeGravity HalfCheetah: Performance and Fairness Gaps}
    \label{fig:hugegravity_comparison}
\end{figure}

\begin{figure}[htbp]
    \centering
    \begin{subfigure}[b]{0.45\textwidth}
        \includegraphics[width=\textwidth]{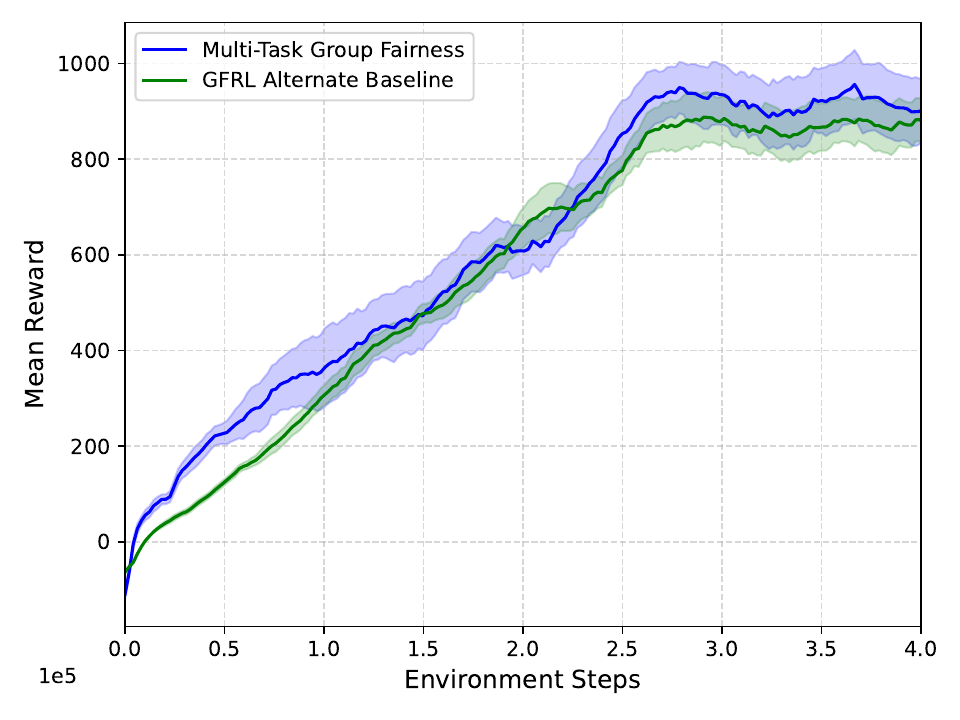}
        \caption{Original HalfCheetah - Forward Running}
        \label{fig:HG_orig_forward}
    \end{subfigure}
    \begin{subfigure}[b]{0.45\textwidth}
        \includegraphics[width=\textwidth]{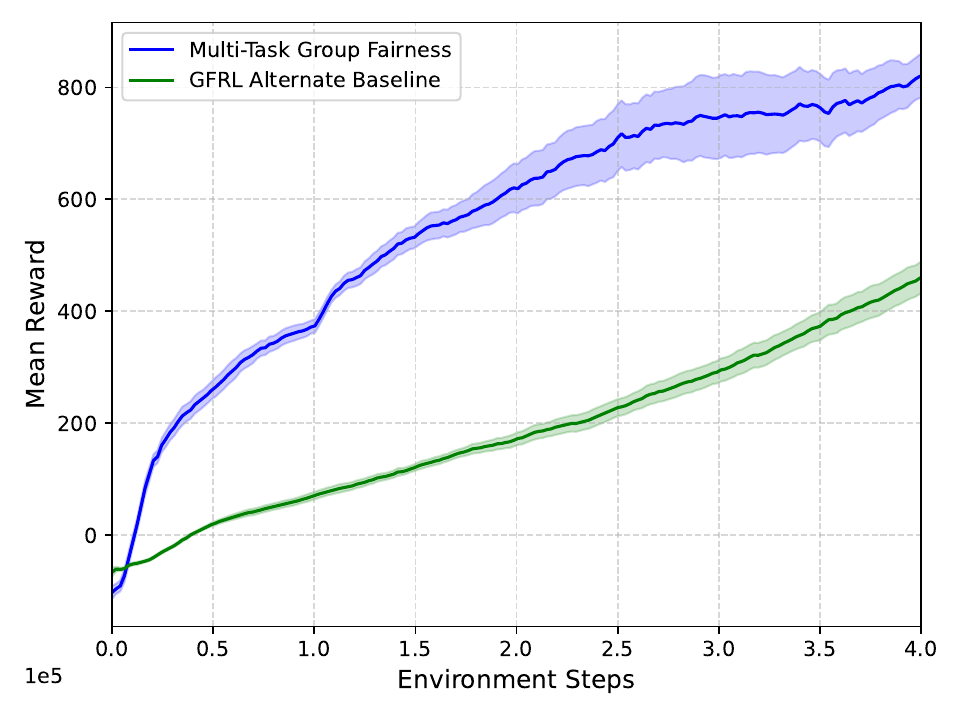}
        \caption{Original HalfCheetah - Backward Running}
        \label{fig:HG_orig_backward}
    \end{subfigure}
    
    \begin{subfigure}[b]{0.45\textwidth}
        \includegraphics[width=\textwidth]{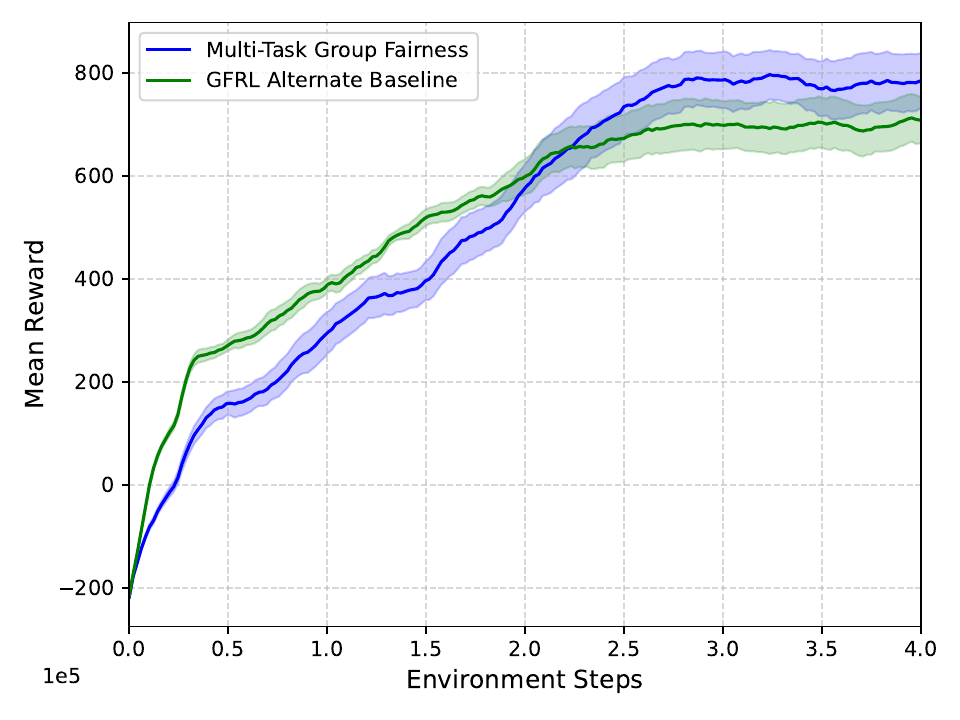}
        \caption{BigFoot HalfCheetah - Forward Running}
        \label{fig:HG_huge_forward}
    \end{subfigure}
    \begin{subfigure}[b]{0.45\textwidth}
        \includegraphics[width=\textwidth]{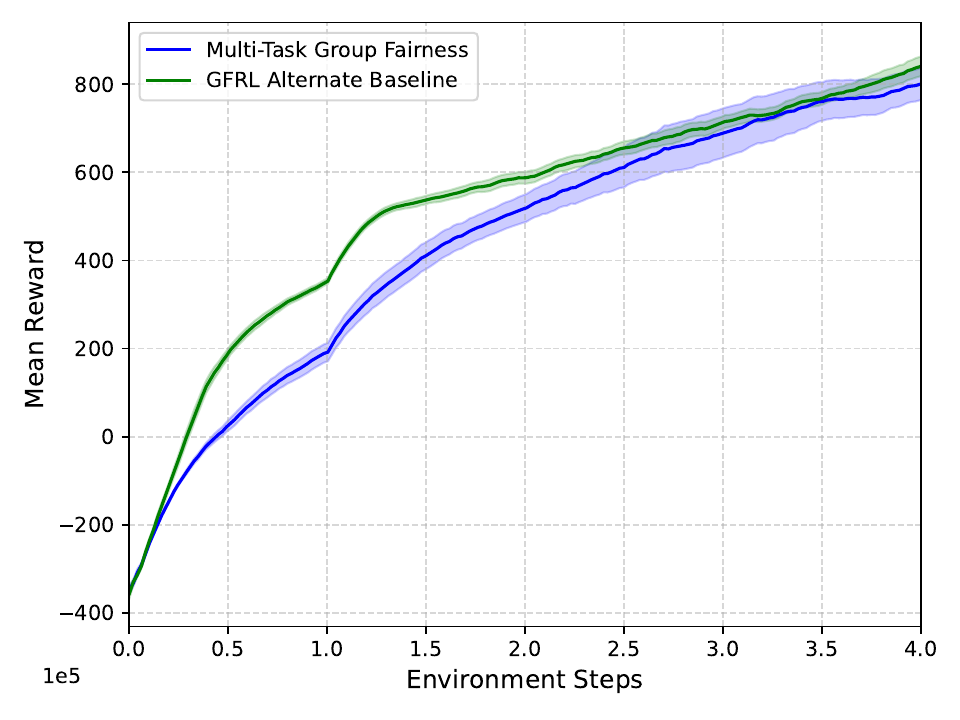}
        \caption{BigFoot HalfCheetah-Backward Running}
        \label{fig:HG_huge_backward}
    \end{subfigure}
    
    \begin{subfigure}[b]{0.45\textwidth}
        \includegraphics[width=\textwidth]{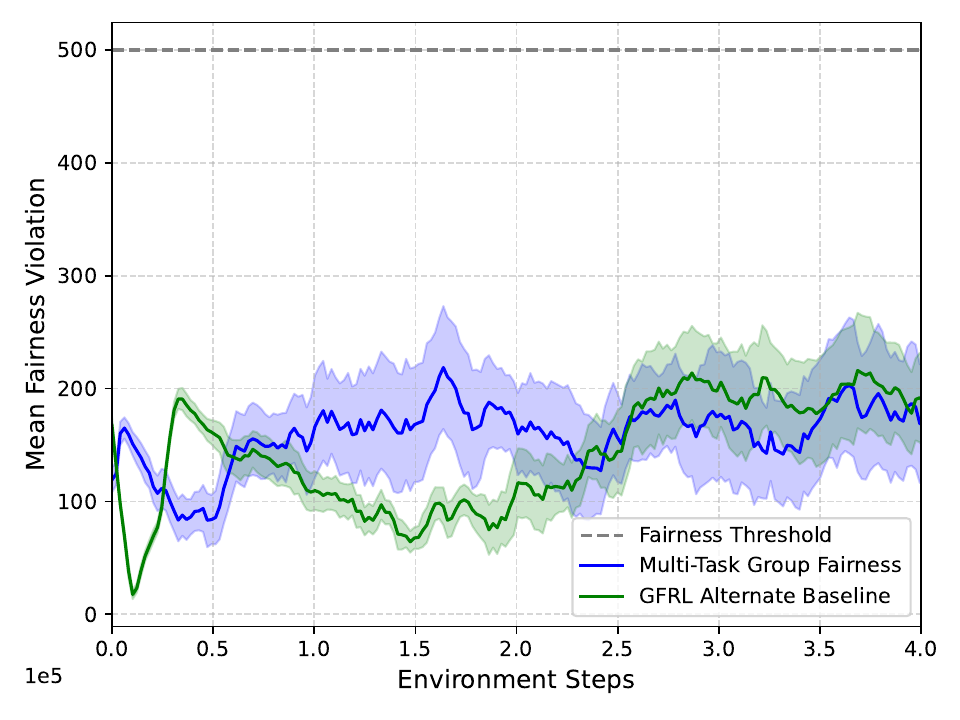}
        \caption{Forward Running Fairness Gap}
        \label{fig:HG_gap_forward}
    \end{subfigure}
    \begin{subfigure}[b]{0.45\textwidth}
        \includegraphics[width=\textwidth]{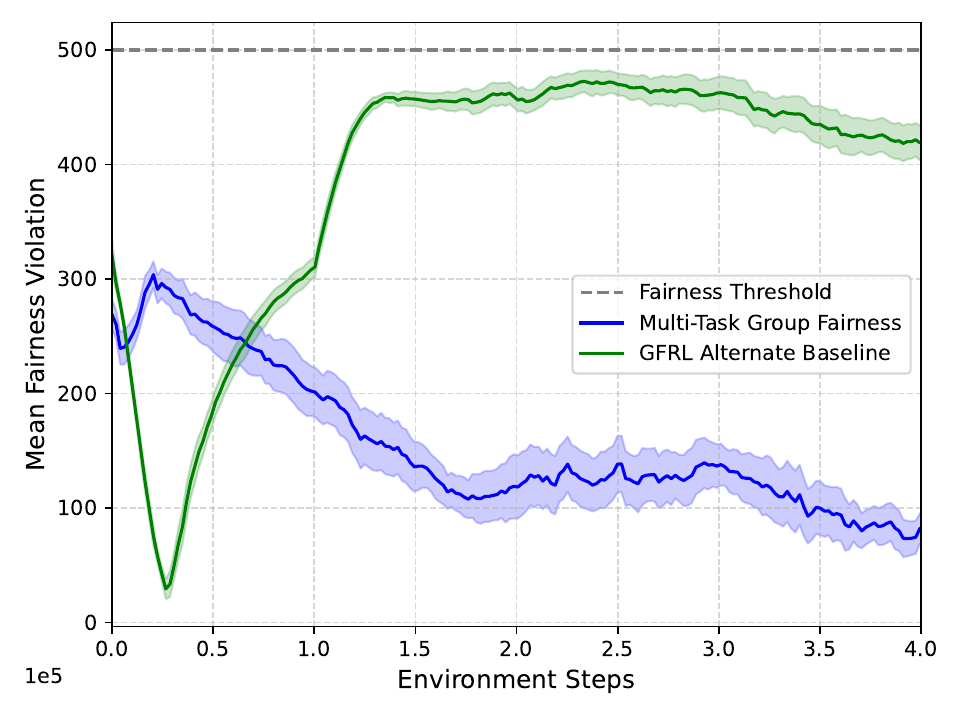}
        \caption{Backward Running Fairness Gap}
        \label{fig:HG_gap_backward}
    \end{subfigure}
    \caption{Comparison between Original HalfCheetah and BigFoot HalfCheetah: Performance and Fairness Gaps}
    \label{fig:hugegravity_comparison}
\end{figure}

\begin{figure}[htbp]
    \centering
    \begin{subfigure}[b]{0.45\textwidth}
        \includegraphics[width=\textwidth]{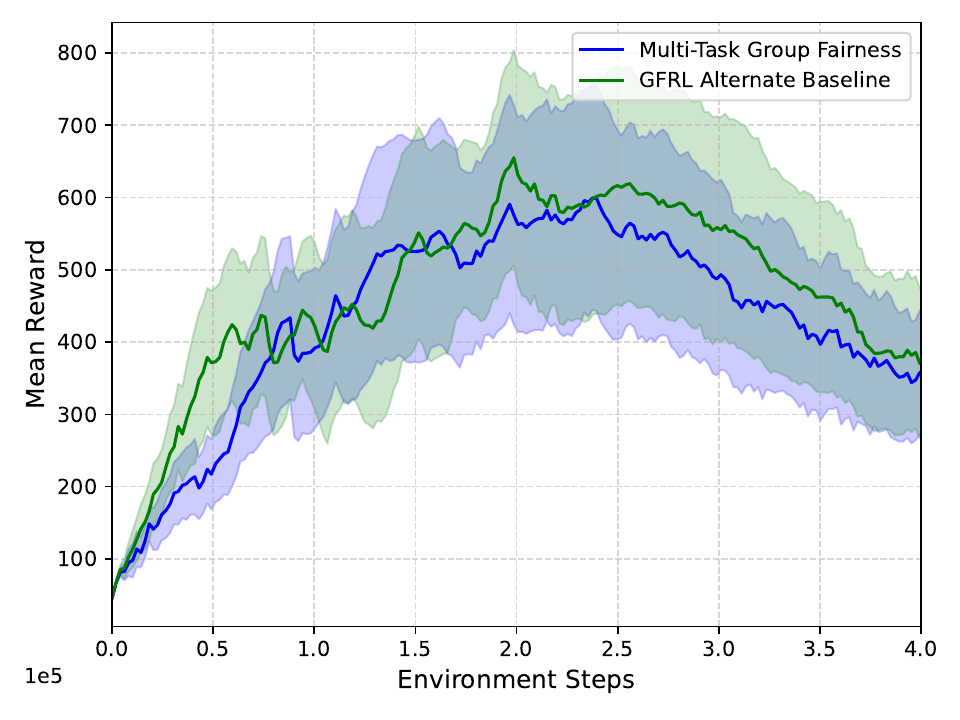}
        \caption{Hopper - Forward Running}
        \label{fig:HG_orig_forward}
    \end{subfigure}
    \begin{subfigure}[b]{0.45\textwidth}
        \includegraphics[width=\textwidth]{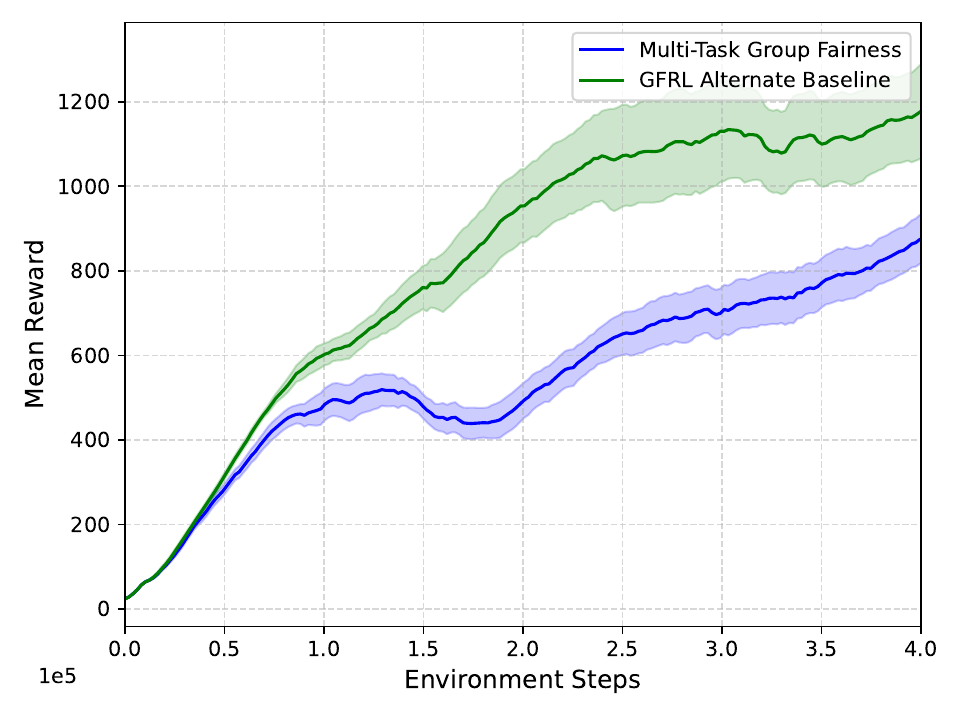}
        \caption{Hopper - Backward Running}
        \label{fig:HG_orig_backward}
    \end{subfigure}
    
    \begin{subfigure}[b]{0.45\textwidth}
        \includegraphics[width=\textwidth]{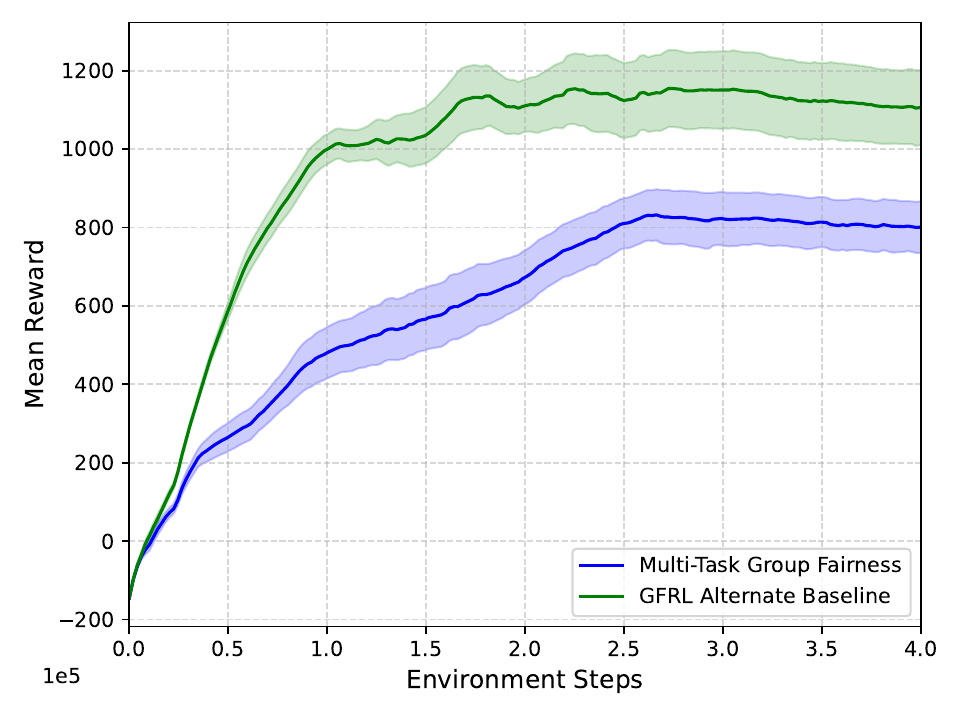}
        \caption{LargeFric HalfCheetah - Forward Running}
        \label{fig:HG_huge_forward}
    \end{subfigure}
    \begin{subfigure}[b]{0.45\textwidth}
        \includegraphics[width=\textwidth]{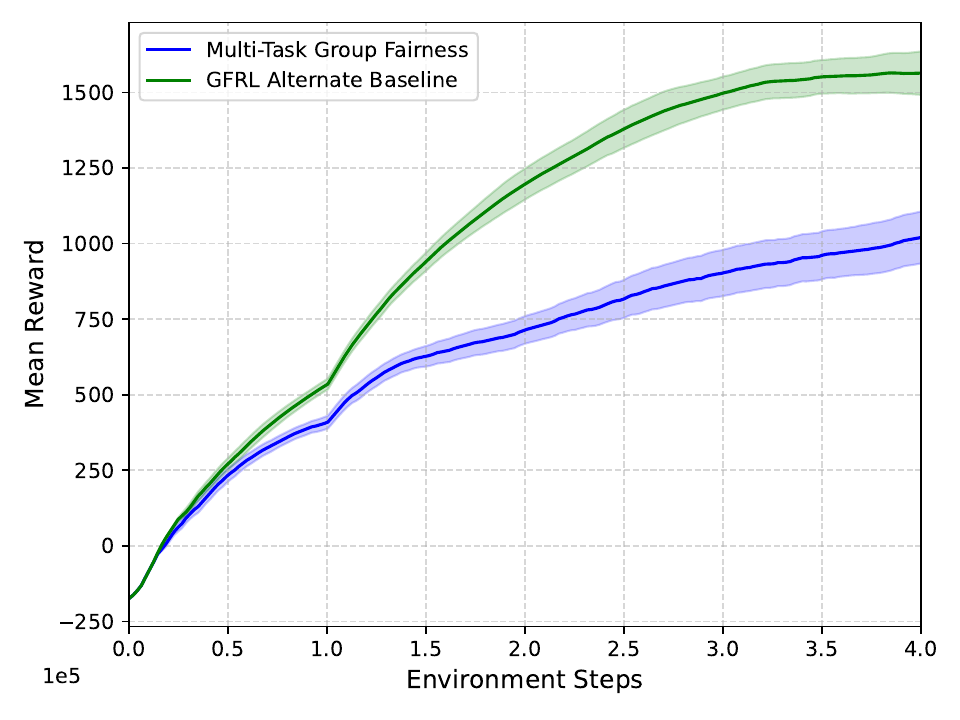}
        \caption{LargeFric HalfCheetah-Backward Running}
        \label{fig:HG_huge_backward}
    \end{subfigure}
    
    \begin{subfigure}[b]{0.45\textwidth}
        \includegraphics[width=\textwidth]{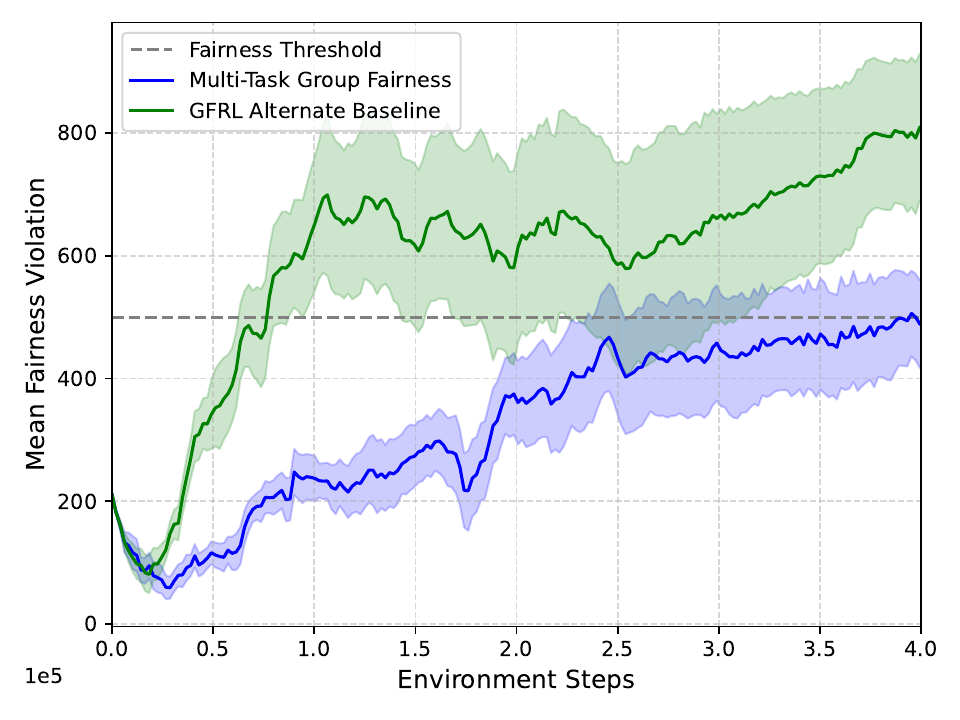}
        \caption{Forward Running Fairness Gap}
        \label{fig:HG_gap_forward}
    \end{subfigure}
    \begin{subfigure}[b]{0.45\textwidth}
        \includegraphics[width=\textwidth]{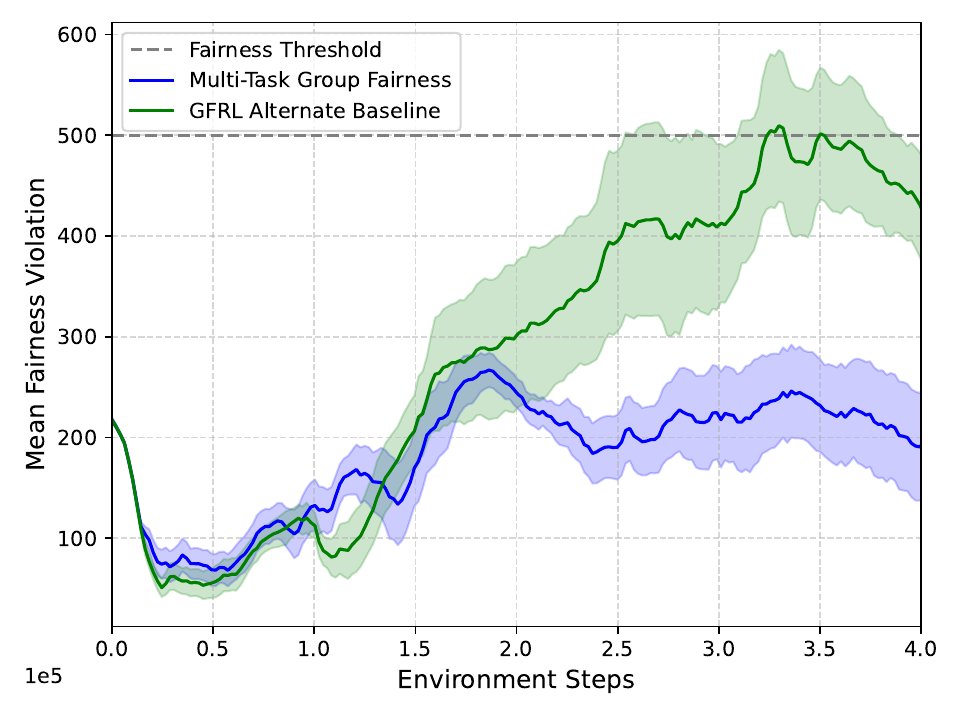}
        \caption{Backward Running Fairness Gap}
        \label{fig:HG_gap_backward}
    \end{subfigure}
    \caption{Comparison between Hopper and LargeFriction HalfCheetah: Performance and Fairness Gaps}
    \label{fig:hugegravity_comparison}
\end{figure}

\end{document}